# UNIFORM INFERENCE FOR HIGH-DIMENSIONAL QUANTILE REGRESSION: LINEAR FUNCTIONALS AND REGRESSION RANK SCORES

Jelena Bradic[†] and Mladen Kolar[‡]

Department of Mathematics, University of California, San Diego [†]
Booth School of Business, The University of Chicago [‡]

**Abstract**

Hypothesis tests in models whose dimension far exceeds the sample size can be formulated much like the classical studentized tests only after the initial bias of estimation is removed successfully. The theory of debiased estimators can be developed in the context of quantile regression models for a fixed quantile value. However, it is frequently desirable to formulate tests based on the quantile regression process, as this leads to more robust tests and more stable confidence sets. Additionally, inference in quantile regression requires estimation of the so called sparsity function, which depends on the unknown density of the error. In this paper we consider a debiasing approach for the uniform testing problem. We develop high-dimensional regression rank scores and show how to use them to estimate the sparsity function, as well as how to adapt them for inference involving the quantile regression process. Furthermore, we develop a Kolmogorov-Smirnov test in a location-shift high-dimensional models and confidence sets that are uniformly valid for many quantile values. The main technical result are the development of a Bahadur representation of the debiasing estimator that is uniform over a range of quantiles and uniform convergence of the quantile process to the Brownian bridge process, which are of independent interest. Simulation studies illustrate finite sample properties of our procedure.

**Keyword:** Inference post-model selection; linear testing; quantile process inference; p-values; robust confidence sets.

## 1 Introduction

High dimensional quantile inference plays a critical role in contemporary robust statistics and machine learning. Development of high-dimensional methods have been driven by the need to study data sets arising in diverse fields, ranging from biology and astronomy to economy and social science, where the technological improvements in ways data are collected have resulted in data sets that could not be studied using the traditional tools. Furthermore, these data sets are often heterogeneous and require the use of robust methods like quantile regression. In this paper, we consider the following linear model

$$Y_i = \beta_0^* + x_{i1}\beta_1^* + \ldots + x_{ip}\beta_p^* + u_i, \qquad i = 1, \ldots, n, \tag{1.1}$$

where $\boldsymbol{\beta}^* = (\beta_0^*, \beta_1^*, \ldots, \beta_p^*)^T \in \mathbb{R}^{p+1}$ is the unknown vector of parameters, $\mathbf{X}_i = (x_{i0}, x_{i1}, \ldots, x_{ip})^T \in \mathbb{R}^{p+1}$ are vectors of (random) input variables with $x_{i0} = 1$, and $u_i$ are random errors. We assume that $u_1, \ldots, u_n$ are i.i.d. with $P[u_i \leq 0] = \frac{1}{2}$. The focus of the paper is construction of robust, uniformly valid confidence intervals for large and growing number of subsets of model parameters in (1.1). In particular, we are interested in constructing confidence bands that are robust to a number of classes of error distributions. A confidence band $C_n = C_n(Y_1, X_1, \ldots, Y_n, X_n)$ is a family of random $\mathbb{R}^p$ intervals $C_n := \{C_n(x) = [cL(y,x), cU(y,x)]\}$ that contain the true parameter $\boldsymbol{\beta}^*$ with a guaranteed probability uniformly over a range of true parameter values $\boldsymbol{\beta}^*$.

Let $F(\cdot)$, $F^{-1}(\cdot)$, and $f(\cdot)$ denote the distribution, quantile and density function of $u_i$. The conditional quantile function of $Y$ given $\mathbf{X}_i$ is

$$Q_\tau[Y \mid \mathbf{X}_i] = \beta_0^* + x_{i1}\beta_1^* + \ldots + x_{ip}\beta_p^* + F^{-1}(\tau) = \mathbf{X}_i^T \boldsymbol{\beta}^*(\tau),$$

where $\boldsymbol{\beta}^*(\tau) = \boldsymbol{\beta}^* + \mathbf{e}_1 F^{-1}(\tau)$. The seminal work of Koenker and Bassett (1978) introduced the quantile regression as a way to estimate the conditional quantile function by finding a minimizer of the following convex program $n^{-1} \sum_{i=1}^n \rho_\tau(Y_i - \mathbf{X}_i^T \boldsymbol{\beta})$, where $\rho_\tau(z) = z(\tau - \mathbb{I}\{z < 0\})$ is the quantile loss function, also known as the check function. During the past four decades, various approaches were proposed for constructing confidence intervals and confidence bands for quantiles with $p \ll n$. They can be grouped into three general approaches: studentization, the bootstrap approach and the direct distribution-free approach. For a fixed $\tau$ and $\alpha$, an approximate studentized $(1-2\alpha)100\%$ confidence interval for $\boldsymbol{\beta}^*(\tau)$, has the standard form $\widehat{\boldsymbol{\beta}}(\tau) \pm \widehat{\varsigma}(\tau) z_\alpha \sqrt{n^{-1} \mathbf{X}^T \mathbf{X} \tau(1-\tau)}$, where $\mathbf{X} \in \mathbb{R}^{n \times (p+1)}$ and $\widehat{\varsigma}(\tau)$ is an $n^{-1/4}$-consistent estimator of the sparsity function $1/f(F^{-1}(\tau))$ and $z_\alpha$ denotes the $(1-\alpha)$th standard normal percentile point (Zhou and Portnoy, 1996). An alternative approach that circumvents the problem of estimating the sparsity function is the bootstrap. However, this approach is considered unsatisfying even in the low dimensional settings (Hall, 1992). Finally, the direct method (Csörgő and Révész, 1984; Zhou and Portnoy, 1996) constructs a confidence interval without estimating the sparsity function as $\widehat{\boldsymbol{\beta}}\left(\tau \pm z_\alpha \sqrt{n^{-1} \mathbf{X}^T \mathbf{X} \tau(1-\tau)}\right)$. Here the confidence interval for $\boldsymbol{\beta}^*(\tau)$ is obtained by two estimates for the nearby quantiles.

In a high dimensional setting, under an assumption that the vector $\boldsymbol{\beta}$ is sparse, an effective way to perform estimation is using penalized methods

$$\widehat{\boldsymbol{\beta}}(\tau) = \arg \min_{\boldsymbol{\beta} \in \mathbb{R}^{p+1}} \frac{1}{n} \sum_{i=1}^n \rho_\tau(Y_i - \mathbf{X}_i^T \boldsymbol{\beta}) + \sum_{j=1}^{p+1} \lambda_j(\tau) \|\beta_j\|. \tag{1.2}$$

Achieving asymptotic coverage when $p \gg n$ is nontrivial due to the presence of bias, which does not vanish even asymptotically irrespective of the penalty function considered (Fan and Lv, 2011). Approaches with non convex penalties (Fan and Li, 2001; Zhang, 2010; Fan and Lv, 2011; Fan et al., 2014; Wang et al., 2013) provide viable alternatives, however, these procedures rely on oracle properties and, as such, do not lead to uniform inference. With an assumption of signal separation, also known as the "beta-min" condition (Zhao and Yu, 2006; Wainwright, 2009), one can establish asymptotic normality for the selected variables, but is left with no inference results for the non-selected ones. Results in Gautier and Tsybakov (2011), Juditsky et al. (2012), and Gautier and Tsybakov (2013) can be used to establish valid conservative bounds on the true parameter based on the $\ell_\infty$ norm convergence rate. These confidence intervals have width of the order $\mathcal{O}(\sqrt{n^{-1} \log(p)})$ under suitable conditions, while we concentrate on construction of confidence intervals with parametric rate. Zhang and Zhang (2013) introduced and van de Geer et al. (2014) further developed, the de-biasing technique, which includes a correction term in the candidate estimator $\widehat{\boldsymbol{\beta}}$ that is able to remove the aforementioned bias. In high-dimensional linear and generalized linear models Belloni et al. (2013a), Belloni et al. (2013d), Javanmard and Montanari (2014) and Farrell (2013), develop a set of complementary de-biasing techniques with one-step or two-step corrections.

In the context of quantile regression, a lot of the early work focused on estimation, prediction and variable selection consistency in low (Li and Zhu, 2008; Zou and Yuan, 2008; Wu and Liu, 2009) and high-dimensional setting (Belloni and Chernozhukov, 2011; Wang et al., 2012; Zheng et al., 2013). Comparatively, little work has been done on construction of valid hypothesis tests, confidence intervals and quantification of uncertainty. Unfortunately, unless perfect model selection is achieved (consistent variables selection), the above three general procedures designed for the confidence intervals with $p \ll n$ do not simply carry over to the high-dimensional setting. If "beta-min" condition is violated, distribution free confidence interval may not be optimal in width, as some components will be shrunk to zero. Moreover, in the context of linear regression with $p \gg n$ Bradic (2013), Chatterjee and Lahiri (2013), Liu and Yu (2013), Chernozhukov et al. (2013) and Lopes (2014) study properties of various bootstrap methods. General conclusion is that, under "no-separation condition" bootstrap approaches may suffer from severe bias. This in turn, may lead to the poor convergence properties of the bootstrap confidence intervals. Therefore, we concentrate on the studentized method.

Belloni et al. (2013c) and Belloni et al. (2013b) propose a three-stage refitting procedure for construction of studentized confidence intervals. However, the point-wise confidence intervals for regression quantiles are valid for a fixed quantile value and a single parameter at a time. There are subtle and yet important issues with the practical use of uniform inference for the quantile process. There is no clear scientific support in choosing one $\tau$ versus another value that is near by. Hence, confidence intervals that can provide coverage for a range of quantile values, say $\tau \in \mathcal{T}$, are very important. Typically the set $\mathcal{T}$ is selected as an interval of quantile levels that well captures a part of the conditional distribution. This is where our paper makes progress on. We establish a uniform Bahadur representation with exact rates of the reminder term for a one-step debasing method, similar in nature to Zhang and Zhang (2013) and Zhao et al. (2014). Different from the previous work, we provide theoretical guarantees for the range of values of $\tau \in \mathcal{T}$ while simultaneously allowing both the sparsity or number of non-zero coordinates of $\beta^*$, $s$, and the ambient dimension, $p$, to depend on the sample size $n$. We are unaware of prior work that successfully shows uniform guarantees. These guarantees are non trivial to establish as traditional arguments do not extend, due to the non-differentiable nature of the quantile process and high-dimensionality of the parameter space.

In addition to the uniform results above, we propose and develop a high-dimensional rank score process, closely related to the quantile process $\boldsymbol{\beta}(\tau)$. We show that this process is exceptionally useful in high-dimensional regimes as it does not depend on the dimension $p$. Instead, it scales linearly with the sample size $n$. Moreover, we illustrate the usefulness of the rank process by providing a novel procedure for the estimation of the sparsity function. Unlike linear models, studentized method for quantiles requires knowledge of the sparsity function. The sparsity function estimator is constructed using the rank scale statistics, which is a linear functional of the proposed rank process, and hence is distribution-free. It is well known (Jurečková et al., 2012) that estimators based on the rank scale statistics are less sensitive to estimation of nuisance parameters, which is useful for approximately sparse models. We establish a number of results illustrating good properties of the proposed estimator. Moreover, we establish the Bahadur representation of the proposed scale rank statistics and establish results that are valid for all quantile levels $\tau \in \mathcal{T}$.

Lastly, we provide a new set of results pertaining to the exact and uniform coverage rates of the studentized confidence intervals and confidence bands. We show that the rate of the coverage is of the order of $\mathcal{O}\left(\left(n^{-1}s^3 \log^5 (p \vee n)\right)^{1/4}\right)$ and that the estimation of the sparsity function does not affect the coverage probability. When $s$ and $p$ are fixed, that is, do not depend on $n$, the rate matches that of Zhou and Portnoy (1996), while in the high-dimensional setting for a fixed $\tau$ the scaling matches that of Belloni et al. (2013b). Moreover, we consider a wide range of hypothesis testing problems including those based on linear functionals of growing dimensions of the process $\boldsymbol{\beta}(\tau)$. We propose a new Kolmogorov-Smirnov test statistics and show its point-wise and uniform convergence under the null hypothesis. The proofs develop arguments to approximate the quantile process by a corresponding Brownian bridge process and allow the dimensionality of the linear tests to depend on $n$. To the best of our knowledge, such results are not available in the existing literature.

## 1.1 Organization of the Paper

This paper investigates the effects of debiasing technique for uniform confidence inference in high-dimensional spaces. In particular, how to incorporate uniform guarantees to improve the continuos inference for the high-dimensional quantile process. The first result, introduced in Section 3, extends previous work on debiasing to incorporate uniform statements for the quantile process. The second result, discussed in Section 4, develops theory for the introduced high-dimensional regression rank score process. In particular, Section 4.1 develops new theory linking the introduced process to the ranks of the residuals of the model (1.1); Section 4.2 proposes a new scaled rank statistics which offers an alternative way of estimating the sparsity function over the existing Koenker's method; Section 4.3 incorporates this newly proposed estimator into the debiased estimator and establishes uniform Bahadur result. Section 5 establishes exact coverage rates of the uniform confidence intervals (Section 5.1) and bands (Section 5.2)– with known and unknown density of the error. Moreover, the subsection 5.3 proposes sup Wald test statistics and shows its usefulness in a range of continuous testing problems. Lastly, Section 6 contains numerical experiments, which show that the proposed method works extremely well.

## 1.2 Notation

Let $[n]$ denote the set $\{1, \ldots, n\}$ and let $\mathbb{I}\{\cdot\}$ denote the indicator function. We use $\mathbf{1}_p$ to denote a vector in $\mathbb{R}^p$ with all components equal to 1. For a vector $a \in \mathbb{R}^d$, we let $\mathrm{supp}(a) = \{j \,:\, a_j \neq 0\}$ be the support set (with an analogous definition for matrices $A \in \mathbb{R}^{n_1 \times n_2}$), $||a||_q$, $q \in [1, \infty)$, the $\ell_q$-norm defined as $||a||_q = (\sum_{i \in [n]} |a_i|^q)^{1/q}$ with the usual extensions for $q \in \{0, \infty\}$, that is, $||\mathbf{a}||_0 = |\mathrm{supp}(\mathbf{a})|$ and $||\mathbf{a}||_\infty = \max_{i \in [n]} |a_i|$. We denote the vector $\ell_\infty$ , $l_1$ and $l_2$-norm of a matrix $A$ as $\|A\|_{\max} = \max_{i \in [n_1], j \in [n_2]} |a_{i,j}|$, $\|A\|_1 = \max_{j \in [n_2]} \sum_{i \in [n_1]} |a_{i,j}|$ and $\|A\|_2 = \sup\{\|Ax\|_2 : \|x\|_2 = 1\}$, respectively. For two sequences of numbers $\{a_n\}_{n=1}^\infty$ and $\{\boldsymbol{\beta}_n\}_{n=1}^\infty$, we use $a_n = \mathcal{O}(\boldsymbol{\beta}_n)$ to denote that $a_n \leq C\boldsymbol{\beta}_n$ for some finite positive constant $C$, and for all $n$ large enough. If $a_n = \mathcal{O}(\boldsymbol{\beta}_n)$ and $\boldsymbol{\beta}_n = \mathcal{O}(a_n)$, we use the notation $a_n \asymp \boldsymbol{\beta}_n$. The notation $a_n = o(\boldsymbol{\beta}_n)$ is used to denote that $a_n \beta_n^{-1} \xrightarrow{n \to \infty} 0$. Throughout the paper, we let $c, C$ be two generic absolute constants, whose values will vary at different locations. For a random variable $Z$, let $||Z||_{\Psi_2} = \sup_{k \geq 1} k^{-1/2} \left( \mathbb{E}\left[|Z|^k\right] \right)^{1/k}$ be the sub-gaussian norm of $Z$. For a random vector $\mathbf{Z}$, the sub-gaussian norm is defined as $||\mathbf{Z}||_{\Psi_2} = \sup \left\{ ||\mathbf{Z}^T \mathbf{a}||_{\Psi_1} \mid ||\mathbf{a}||_2 = 1 \right\}$.

## 2 Preliminaries: de-biased estimator

Let $(X_i, Y_i)$ be $n$ independent observations from the model in (1.1). We estimate the conditional quantile function of $Y$ given $\mathbf{X}$ by minimizing the penalized objective in (1.2) with the penalty parameter set as

$$\lambda_j(\tau) = \lambda_0 \sqrt{\tau(1-\tau)} \widehat{\sigma}_j \tag{2.1}$$

where $\widehat{\sigma}_j^2 = n^{-1} \sum_{i \in [n]} x_{ij}^2$ as suggested in Belloni and Chernozhukov (2011). Let $\widehat{\boldsymbol{\beta}}$ be a minimizer of (1.2) with the penalty set as in (2.1). Given $\widehat{\boldsymbol{\beta}}$, a debiased estimator $\check{\boldsymbol{\beta}}$ can be constructed as

$$\check{\boldsymbol{\beta}}(\tau) = \widehat{\boldsymbol{\beta}}(\tau) + n^{-1} \widehat{\mathbf{D}}_\tau \sum_{i \in [n]} \mathbf{X}_i \Psi_\tau \left( Y_i - \mathbf{X}_i^T \widehat{\boldsymbol{\beta}}(\tau) \right) \tag{2.2}$$

where $\psi_\tau(z) = \tau - \mathbb{I}\{z < 0\}$ and $\widehat{\mathbf{D}}_\tau = \widehat{\varsigma}(\tau) \widehat{\mathbf{D}}$ with $\widehat{\varsigma}(\tau)$ being an estimator of the sparsity function

$$\varsigma(\tau) = 1/f(F^{-1}(\tau)) \tag{2.3}$$

and $\widehat{\mathbf{D}}$ being en estimator of the inverse of the covariance matrix $\boldsymbol{\Sigma} = \mathbb{E}[\widehat{\boldsymbol{\Sigma}}]$, with $\widehat{\boldsymbol{\Sigma}} = n^{-1} \sum_{i \in [n]} \mathbf{X}_i \mathbf{X}_i^T$. Let $\mathbf{D}^p = [\mathbf{d}_1^p, \ldots, \mathbf{d}_{p+1}^p]$ where the column $\mathbf{d}_j^p$ is obtained as a solution to the following optimization program

$$\begin{aligned} \widehat{\mathbf{d}}_j^p = \ & \arg\min_{\mathbf{d} \in \mathbb{R}^p} \ \|\mathbf{d}\|_1 \\ & \text{s.t.} \ \left\| \widehat{\boldsymbol{\Sigma}} \mathbf{d} - \mathbf{e}_j \right\|_\infty \leq \gamma_n \\ & \quad \ \left| \mathbf{X}_i^T \mathbf{d} \right| \leq L_n, \quad i = 1, \ldots, n, \end{aligned} \tag{2.4}$$

where $\gamma_n$ and $L_n$ are a-priori defined tuning parameters. As the matrix $\mathbf{D}^p$ is not symmetric in general, we symmetrize it to obtain the final estimator $\widehat{\mathbf{D}} = (\widehat{d}_{ij})_{i,j \in [p+1]}$ as follows

$$\widehat{d}_{ij} = \widehat{d}_{ji} = d_{ij}^p \, \mathbb{I}\{|d_{ij}^p| \leq |d_{ji}^p|\} + d_{ji}^p \, \mathbb{I}\{|d_{ij}^p| > |d_{ji}^p|\}.$$

We leave the details of our novel construction of $\widehat{\varsigma}(\tau)$ for Section 4. The debiased estimator above is of a similar form to that studied in Zhang and Zhang (2013) and van de Geer et al. (2014), and is different from those in Javanmard and Montanari (2014) and Zhao et al. (2014). In particular, it produces a consistent estimator of $\boldsymbol{\Sigma}^{-1}$ which is needed for construction of linear tests and uniform confidence bands as they depend on consistent estimation of possibly all correlations, whereas point-wise confidence intervals do not. Notice that in (2.4) the CLIME estimator (Cai et al., 2011) is augmented with an additional constraint in order to control the residual terms in the Bahadur representation. This allows us to provide uniform inference results for a quantile process as a function of $\tau$. To the best of our knowledge, such uniform results that allow both $s$ and $p$ to depend on $n$ are new.

Using Knight's identity $|z-z'|-|z| = -v\left[\mathbb{I}\{z>0\} - \mathbb{I}\{z<0\}\right]+2\int_o^{z'}\left[\mathbb{I}\{z\leq s\} - \mathbb{I}\{z\leq 0\}\right]ds$, for $z \neq 0$, and the fact that $\rho_\tau(z) = |z|/2 + (\tau - 1/2)z$ it is easy to see that

$$\rho_\tau\left(Y_i - \mathbf{X}_i^T\boldsymbol{\beta}\right) = \rho_\tau\left(u_i - F^{-1}(\tau)\right) - \mathbf{X}_i^T\left(\boldsymbol{\beta} - \boldsymbol{\beta}^*(\tau)\right)\psi_\tau\left(u_i - F^{-1}(\tau)\right)$$
$$+ \int_0^{\mathbf{X}_i^T(\boldsymbol{\beta}-\boldsymbol{\beta}^*)}\left[\mathbb{I}\{u_i - F^{-1}(\tau) \leq s\} - \mathbb{I}\{u_i - F^{-1}(\tau) \leq 0\}\right]ds. \tag{2.5}$$

We will use this identity to show that $\widetilde{\mathbf{D}}_\tau = \varsigma(\tau)\widehat{\mathbf{D}}$ is a uniformly good approximation of the population Hessian over a large range of values of $\tau$ and that the optimization program (2.4) is a good proxy for the one-step Newton update for the quantile process.

Before we state the characterization of the proposed estimator $\check{\boldsymbol{\beta}}(\tau)$, (2.2), we introduce some additional notation to simplify the exposition. Let $G_n(\tau, \boldsymbol{\delta})$ denote the weighted empirical process

$$G_n(\tau, \boldsymbol{\delta}) = \nu_n(\tau, \boldsymbol{\delta}) - \mathbb{E}[\nu_n(\tau, \boldsymbol{\delta})], \tag{2.6}$$

where for $\boldsymbol{\delta} \in \mathbb{R}^p$

$$\nu_n(\tau, \boldsymbol{\delta}) = n^{-1}\sum_{i\in[n]}\widetilde{\mathbf{D}}_\tau\mathbf{X}_i\left(\mathbb{I}\left\{u_i \leq \mathbf{X}_i^T\boldsymbol{\delta} + F^{-1}(\tau)\right\} - \mathbb{I}\left\{u_i \leq F^{-1}(\tau)\right\}\right). \tag{2.7}$$

**Proposition 2.1.** *Suppose that the optimization program in (2.4) is feasible. The debiased estimator (2.2) has the following representation uniformly in $\tau$*

$$\sqrt{n}(\check{\boldsymbol{\beta}}(\tau) - \boldsymbol{\beta}^*(\tau)) = n^{-1/2}\widehat{\mathbf{D}}_\tau\sum_{i\in[n]}\mathbf{X}_i\psi_\tau\left(u_i - F^{-1}(\tau)\right) - \sqrt{n}\left(\boldsymbol{\Delta}_1(\tau) + \boldsymbol{\Delta}_2(\tau) + \boldsymbol{\Delta}_3(\tau)\right), \tag{2.8}$$

*where*

$$\boldsymbol{\Delta}_1(\tau) = \frac{\widehat{\varsigma}(\tau)}{\varsigma(\tau)}G_n\left(\tau, \widehat{\boldsymbol{\beta}}(\tau) - \boldsymbol{\beta}^*(\tau)\right), \qquad \boldsymbol{\Delta}_2(\tau) = \left(\frac{\widehat{\varsigma}(\tau)}{\varsigma(\tau)}\widehat{\mathbf{D}}\widehat{\boldsymbol{\Sigma}} - \mathbf{I}\right)\left(\widehat{\boldsymbol{\beta}}(\tau) - \boldsymbol{\beta}^*(\tau)\right),$$
$$\boldsymbol{\Delta}_3(\tau) = n^{-1}\frac{\widehat{\varsigma}(\tau)}{\varsigma(\tau)}\sum_{i\in[n]}f(\bar{w}_i)\widetilde{\mathbf{D}}_\tau\mathbf{X}_i\left(\mathbf{X}_i^T\left(\widehat{\boldsymbol{\beta}}(\tau) - \boldsymbol{\beta}^*(\tau)\right)\right)^2$$

*with $\bar{w}_i$ between $\mathbf{X}_i^T\left(\widehat{\boldsymbol{\beta}}(\tau) - \boldsymbol{\beta}^*(\tau)\right) + F^{-1}(\tau)$ and $F^{-1}(\tau)$.*

We will use Proposition 2.1 to establish novel Bahadur representation in Section 3 that holds over a range of values of $\tau$. Notice that the statement of Proposition 2.1 is deterministic. Under the conditions given in the following section, we will show that the term $\sqrt{n}\left(\boldsymbol{\Delta}_1(\tau) + \boldsymbol{\Delta}_2(\tau) + \boldsymbol{\Delta}_3(\tau)\right)$ is small uniformly in $\tau$ with high probability, while the first term in (2.8) converges to a Brownian bridge process.

## 3 Uniform Bahadur Representation

Different from the classical setting, our representation provides a convergence rate of the remainder term while allowing both $s$ and $p$ to grow with $n$. Bahadur (1966) representations are useful for construction of confidence bands, especially when the loss function is not smooth, such as in M-estimation and quantile regression. They are also significantly more challenging to establish for non-smooth losses. To simplify presentation, we first establish Bahadur representation for the case when the density function $f$ is known. In Section 4 we detail a novel procedure for estimating the sparsity function based on the regression rank scores, prove that the sparsity function can be estimated uniformly over $\tau$, and give the rate of convergence.

We require the following regularity conditions.

- **(D)** The density of the error, $f$, is uniformly bounded from above by $f_{\max} < \infty$ and from below by $f_{\min} > 0$. Furthermore, $f$ is continuously differentiable with the derivative, $f'$, bounded by $f'_{\max}$.

- **(X)** The vector of covariates $\mathbf{x}_i = (1, \widetilde{\mathbf{x}}_i^T)^T \in \mathbb{R}^{p+1}$ satisfies $||\boldsymbol{\Sigma}^{-1/2}\widetilde{\mathbf{x}}_i||_{\Psi_2} = \sigma_X$ where $\boldsymbol{\Sigma} = \text{Cov}(\widetilde{\mathbf{x}}_i)$. Furthermore, the covariance matrix satisfies $0 < \Lambda_{\min} \leq \Lambda_{\min}(\boldsymbol{\Sigma}) \leq \Lambda_{\max}(\boldsymbol{\Sigma}) \leq \Lambda_{\max} < \infty$ and the precision matrix $\boldsymbol{\Sigma}^{-1}$ satisfies $\max_{j \in [p]} ||\{\boldsymbol{\Sigma}^{-1}\}_j||_0 \leq \bar{s}$ and $||\boldsymbol{\Sigma}^{-1}||_1 \leq M$ for some $\Lambda_{\min}, \Lambda_{\max}, \bar{s}$ and $M$.

- **(S)** The vector of coefficients $\boldsymbol{\beta}^* \in \mathbb{R}^{p+1}$ is sparse, that is, $s = |S| = |\{\beta_j \neq 0 \mid j \in [p+1]\}| \ll n$.

The assumptions above are standard in the literature of high-dimensional inference and are just one set of sufficient conditions. For example, assumption **(X)** was considered in the literature on inverse covariance matrix estimation (Cai et al., 2016) and in a slightly weaker form for inference of univariate parameters (van de Geer et al., 2014). However, in this paper we are interested in linear testing of multivariate parameters which requires slightly stronger conditions. Assumption **(D)** on the density of the error $f$ is common in the literature on quantile regression (Koenker, 2005). Finally, the hard sparsity assumption on $\boldsymbol{\beta}^*$, **(S)**, can be relaxed to allow a number of coefficients with small elements (Belloni et al., 2013b). We chose the above conditions to simplify the presentation and use existing results on the first step estimator $\widehat{\boldsymbol{\beta}}(\tau)$. In particular, under our regularity conditions **(D)**, **(X)**, and **(S)**, the conditions D.4 and D.5 of Belloni and Chernozhukov (2011) are satisfied (see Lemma 1 in their paper). Hence, Theorem 2 and Theorem 3 of Belloni and Chernozhukov (2011) apply and we have

$$\sup_{\tau \in [\epsilon, 1-\epsilon]} ||\widehat{\boldsymbol{\beta}}(\tau) - \boldsymbol{\beta}^*(\tau)||_2 = \mathcal{O}_P\left(\sqrt{\frac{s \log(p \vee n)}{n}}\right) \quad (3.1)$$

and $\sup_{\tau \in [\epsilon, 1-\epsilon]} ||\widehat{\boldsymbol{\beta}}(\tau)||_0 = \mathcal{O}_P(s)$. The latter can be treated as a condition on the initial step estimator – see Belloni et al. (2013) for more details where the authors introduce an additional step to remove this assumption. For the case of known density function $f$, the debiased estimator takes the form

$$\bar{\boldsymbol{\beta}}(\tau) = \widehat{\boldsymbol{\beta}}(\tau) + n^{-1}\widetilde{\mathbf{D}}_\tau \sum_{i \in [n]} \mathbf{X}_i \Psi_\tau\left(Y_i - \mathbf{X}_i^T \widehat{\boldsymbol{\beta}}(\tau)\right). \quad (3.2)$$

The stimator $\bar{\boldsymbol{\beta}}(\tau)$ is analogous to $\check{\boldsymbol{\beta}}(\tau)$ in (2.2), however, $\widehat{\mathbf{D}}_\tau$ is replaced with $\widetilde{\mathbf{D}}_\tau = \varsigma(\tau)\widehat{\mathbf{D}}$. The following theorem provides Bahadur representation for the estimator $\bar{\boldsymbol{\beta}}(\tau)$.

**Theorem 3.1.** *Suppose that the assumptions* **(D)**, **(X)**, *and* **(S)** *hold. Furthermore, let $\widehat{\boldsymbol{\beta}}(\tau)$ be a minimizer of* (1.2) *with the penalty parameter set as in* (2.1) *with $\lambda_0 = c_1\sqrt{n^{-1}\log(p)}$ and $\widehat{\mathbf{D}}$ be* a minimizer of (2.4) with $\gamma_n = c_2 M \sqrt{\log(p)/n}$, $M = \|\mathbf{\Sigma}^{-1}\|_1$ and $L_n = c_3 \Lambda_{\min}^{-1/2}(\mathbf{\Sigma}) \sqrt{\log(p \vee n)}$. Then the debiased estimator $\bar{\boldsymbol{\beta}}(\tau)$ in (3.2) satisfies

$$\sqrt{n}(\bar{\boldsymbol{\beta}}(\tau) - \boldsymbol{\beta}^*(\tau)) = n^{-1/2} \widetilde{\mathbf{D}}_\tau \sum_{i \in [n]} \mathbf{X}_i \psi_\tau \left(u_i - F^{-1}(\tau)\right) + \mathcal{O}_P \left( \frac{L_n s^{3/4} \log^{3/4}(p \vee n)}{n^{1/4}} \vee \frac{L_n s \log(p \vee n)}{n^{1/2}} \right) \quad (3.3)$$

uniformly in $\tau \in [\epsilon, 1-\epsilon]$ for $\epsilon \in (0, 1)$.

Theorem 3.1 is a generalization of Theorem 2.1 of Zhou and Portnoy (1996) as it provides a Bahadur representation of regression quantiles for a high dimensional linear model. The theoretical development for the high dimensional case cannot rely on the traditional empirical process arguments. We devise new chaining arguments to circumvent both the non-smoothness of the process at hand and its high dimensionality. We observe that in the high dimensional setting, the residual term in the Bahadur representation is of higher order compared to the classical quantile estimator. Zhou and Portnoy (1996) establish the residual to be of the order of $(\log n/n)^{3/4}$, whereas we establish a rate of $(s^3 \log^5 p/n^3)^{1/4}$ whenever $s \log p/n < 1$. Moreover, the result is uniform in nature, as it holds true for all $\epsilon \leq \tau \leq 1 - \epsilon$ and $\epsilon \in (0, 1)$. To the best of our knowledge this is the first result establishing uniform and high dimensional guarantees. The proof of Theorem 3.1 begins with Proposition 2.1 and shows uniform bounds on the terms $\boldsymbol{\Delta}_1(\tau), \boldsymbol{\Delta}_2(\tau), \boldsymbol{\Delta}_3(\tau)$ appearing therein. These bounds are given in Lemma 1, 2, and 3 below.

**Lemma 1.** *Suppose the assumptions of Theorem 3.1 are satisfied and $f_{\max} \Lambda_{\max} r_n = \mathcal{O}_P(1)$. Then there exists a fixed constant $C$, such that*

$$\sup_{\tau \in [\epsilon, 1-\epsilon]} \sup_{\boldsymbol{\delta} \in \mathcal{C}(r_n, t)} \|G_n(\tau, \boldsymbol{\delta})\|_\infty \leq C \left( \sqrt{\frac{f_{\min}^{-2} L_n^2 f_{\max} \Lambda_{\max}^{1/2}(\mathbf{\Sigma}) r_n t \log(np/\delta)}{n}} \vee \frac{f_{\min}^{-1} L_n t \log(np/\delta)}{n} \right).$$

*with probability $1 - \delta$, where*

$$\mathcal{C}(r, t) = \{\mathbf{w} \in \mathbb{R}^{p+1} \mid \|\mathbf{w}\|_2 \leq r, \|\mathbf{w}\|_0 \leq t\}$$

*denote a ball around the origin of vectors that are $t$-sparse and $G_n(\tau, \boldsymbol{\delta})$ is defined in (2.6).*

A few remarks are in order. Lemma 1 establishes a doubly uniform tail probability bound for a growing supremum of an empirical process $G_n(\tau, \boldsymbol{\delta})$. It is doubly uniform as it covers both a large range of $\boldsymbol{\delta}$ and of $\tau$; and it is growing as supremum is taken over $p$ coordinates of the process, which are allowed to grow with $n$. The proof of Lemma 1 is further challenged by the non-smooth components of the process $G_n(\tau, \boldsymbol{\delta})$ itself. It proceeds in two steps. First, we show that for a fixed $\tau$ and $\boldsymbol{\delta}$ the term $\|G_n(\tau, \boldsymbol{\delta})\|_\infty$ is small. In the second step, we devise a new epsilon net argument to control the non-smooth terms uniformly for all $\tau$ and $\boldsymbol{\delta}$ simultaneously. In conclusion, with $r_n = C\sqrt{s \log(p \vee n)/n}$ and $t = Cs$ (see (3.2)), Lemma 1 establishes a uniform bound $\sup_{\tau \in [\epsilon, 1-\epsilon]} \sqrt{n} \Delta_1(\tau) = o_P(1)$.

**Lemma 2.** *Suppose the assumptions of Theorem 3.1 are satisfied. Then the optimization program in (2.4) has a feasible solution with probability at least $1 - Cp^{-c_3}$.*

The proof of Lemma 2 is a modification of the result of Javanmard and Montanari (2014) and shows that the population precision matrix $\mathbf{\Sigma}^{-1}$ is feasible for the optimization program with high-probability.

**Lemma 3.** *Let $\{\bar{w}_i\}_{i \in [n}$ be such that each $\bar{w}_i$ is between $\mathbf{X}_i^T \boldsymbol{\delta} + F^{-1}(\tau)$ and $F^{-1}(\tau)$. Under the conditions of Theorem 3.1, we have that*

$$\sup_{\tau \in [\epsilon, 1-\epsilon]} \left\| n^{-1} \sum_{i \in [n]} f(\bar{w}_i) \widetilde{\mathbf{D}}_\tau \mathbf{X}_i \left( \mathbf{X}_i^T \left( \widehat{\boldsymbol{\beta}}(\tau) - \boldsymbol{\beta}^*(\tau) \right) \right)^2 \right\|_\infty = \mathcal{O}_P \left( \frac{L_n s \log(p)}{n} \right).$$

Result of Lemma 3 is based on the above two Lemmas and equation (3.1). Lemma 3 shows that $\sup_{\tau \in [\epsilon, 1-\epsilon]} \sqrt{n} \Delta_3(\tau) = o_P(1)$.

# 4 Density Estimation Through Regression Rank Scores

In this section we introduce a high dimensional regression rank scores in the regression model (1.1) with $p \geq n$ that generalize regression rank scores of Gutenbrunner and Jurečková (1992). We provide a uniform, finite sample, Bahadur representation of such high dimensional regression rank scores, generalizing the asymptotic results of Gutenbrunner and Jurečková (1992) to the case of $p \geq n$. The introduced regression rank scores provide a foundation for the new theory of linear rank statistics which we use to provide a new, distribution free estimator of the sparsity function $\zeta(\tau)$ defined in (2.3).

## 4.1 High Dimensional Regression Rank Scores

We define high-dimensional regression rank scores, $\widehat{\boldsymbol{\xi}}(\tau)$, as the solution to the dual of the primal problem (1.2). For a fixed value of the tuning parameter $\lambda_n$ and the quantile $\tau$, we define $\widehat{\boldsymbol{\xi}}(\tau)$ as the solution to the following optimization problem

$$\begin{aligned}
\max_{\boldsymbol{\xi}} \quad & \mathbf{Y}^T \boldsymbol{\xi}(\tau) \\
\text{s.t.} \quad & \|\mathbf{X}^T \boldsymbol{\xi}(\tau) - (1-\tau) \mathbf{X}^T \mathbf{1}_n\|_\infty \leq \lambda_n, \\
& \boldsymbol{\xi}(\tau) \in [0,1]^n.
\end{aligned} \quad (4.1)$$

In the display above, it is clear that the high dimensional rank scores $\widehat{\boldsymbol{\xi}}(\tau)$ depend on the value of the tuning parameter $\lambda_n$, $\widehat{\boldsymbol{\xi}}(\tau) = \widehat{\boldsymbol{\xi}}(\tau, \lambda_n)$. This dependence on $\lambda_n$ is suppressed in the notation whenever possible.

While the regression quantiles are suitable mainly for estimation, the regression rank scores are used for testing hypotheses in the linear model, particularly when the hypothesis concerns only some components of $\boldsymbol{\beta}^*$ and other components are considered as nuisance. In what follows, we show that the introduced penalized regression rank scores $\widehat{\boldsymbol{\xi}}(\tau)$ are uniformly close to the regression rank scores $\boldsymbol{\alpha}(\tau)$ of Gutenbrunner and Jurečková (1992). Let $S$ and $s$ be defined as in Condition **(S)**. Then the regression rank scores $\boldsymbol{\alpha}(\tau)$ are defined as the solution to the following optimization problem

$$\begin{aligned}
\max_{\boldsymbol{\alpha}} \quad & \mathbf{Y}^T \boldsymbol{\alpha}(\tau) \\
\text{s.t.} \quad & \mathbf{X}_S^T \boldsymbol{\alpha}(\tau) = (1-\tau) \mathbf{X}_S^T \mathbf{1}_n \\
& \boldsymbol{\alpha}(\tau) \in [0,1]^n.
\end{aligned} \quad (4.2)$$

While allowing the size of the set $S$, $s$, to be a function of $n$ and to diverge with $n$ we show the following result.

**Lemma 4.** *Assume that the conditions of Theorem 3.1 hold. Let $\boldsymbol{\alpha}(\tau)$ be a solution to the dual problem in (4.2). Furthermore, let $E_{i\tau} = u_i - F^{-1}(\tau)$, and $\bar{\alpha}_i(t) = \mathbb{I}\{E_{i\tau} \geq 0\}$. Then,*

$$\sup_{\tau \in [\epsilon, 1-\epsilon]} \frac{1}{n} \sum_{i \in [n]} E_{i\tau} |\alpha_i(\tau) - \bar{\alpha}_i(\tau)| = \mathcal{O}_P\left(s\sqrt{\frac{\log n}{n}}\right), \quad (4.3)$$

*for any fixed $\epsilon \in (0, 1)$.*

Lemma 4 shows that the regression rank scores of Gutenbrunner and Jurečková (1992) can be well approximated by an empirical process. The uniform upper bound indicates that the uniform consistency of the regression rank scores is up to an order of $s\sqrt{\log n / n}$. It is the first result that explicitly establishes the rate of convergence of such uniform approximation that is the function of both $s$ and $n$. If we only consider a finite number of $\tau$'s we obtain the rate of $\sqrt{s^2/n}$. When $s$ is not diverging this rate matches the optimal rate of Gutenbrunner and Jurečková (1992) (see Theorem 1 (ii) therein).

**Lemma 5.** *Assume that the conditions of Theorem 3.1 hold. Let $\boldsymbol{\alpha}(\tau)$ be a solution to the dual problem in (4.2) and let $\widehat{\boldsymbol{\xi}}(\tau)$ be a solution to (4.1). Suppose that the assumptions (D), (X), and (S) hold. Then*

$$\sup_{\tau \in [\epsilon, 1-\epsilon]} \left| n^{-1} \sum_{i \in [n]} Y_i \left(\widehat{\xi}_i(\tau) - \alpha_i(\tau)\right) \right| = \mathcal{O}_P\left(\lambda_n \left(\sup_{\tau \in [\epsilon, 1-\epsilon]} \|\boldsymbol{\beta}^*(\tau)\|_1\right) + s\sqrt{\frac{\log(p \vee n)}{n}}\right), \quad (4.4)$$

*for any $\epsilon \in (0, 1)$.*

Lemma 5 shows that if $\lambda_n$ is chosen appropriately, then the high dimensional regression rank scores can be well approximated by the regression rank scores. Moreover, this approximation is uniform in that the upper bound is established over a large set of values of $\tau$. The reminder term in the approximation indicates that the bias in this approximation is of the order of $\lambda_n \left(\sup_{\tau \in [\epsilon, 1-\epsilon]} \|\boldsymbol{\beta}^*(\tau)\|_1\right)$ and is diminishing as long as $\lambda_n$ is chosen appropriately. This establishes a uniform approximation of two linear rank statistics based on the penalized regression rank scores and the regression rank scores. Regression rank statistics are useful for designing rank estimators while results above are useful in establishing asymptotic properties of these estimators. This is the first result on regression rank scores developed in the setting where $s$ and $p$ are allowed to grow with $n$.

### 4.2 Rank Density Estimation

In this section we propose a new estimator for the sparsity function $\varsigma(\tau) = 1/f(F^{-1}(\tau))$, which is based on the regression rank scores. The sparsity function is commonly estimated using the Koenker's quotient estimator. Let $\widehat{Q}_Y(\tau) = n^{-1} \sum_{i=1}^n \mathbf{X}_i^\top \widehat{\boldsymbol{\beta}}(\tau)$ be the estimated conditional quantile function, where $\widehat{\boldsymbol{\beta}}(\tau)$ is a consistent estimator of $\beta^*(\tau)$. Then Koenker's estimator of $\varsigma(\tau)$ is defined as

$$\left(\widehat{Q}_Y(\tau + h) - \widehat{Q}_Y(\tau - h)\right)/(2h),$$

where $h$ is a bandwidth parameter. Belloni et al. (2013b) employ such an estimator in a high-dimensional setting. In contrast we take a new approach and design a rank statistics based on the introduced high-dimensional regression rank scores. Both approaches achieve consistency, but the estimator based on ranks has a smaller bias (see Theorem 4.1 bellow). Furthermore, the estimator is less sensitive in situations when data are not normally distributed.

One of the main advantages of regression rank scores is their regression equivariance, which is useful in designing a number of distribution-free tests. The ranks allow for a distribution free inference and are widely used in robust statistics (Jurečková et al., 2012). Furthermore, ranks are robust to outliers in the response variable and invariant with respect to the regression shift of the nuisance parameter. The invariance property is highly desirable in high dimensional models where perfect model selection is not possible.

To estimate the sparsity function, $\varsigma(\tau)$, we propose a scale rank statistics and then connect it with $\varsigma(\tau)$. The scale rank statistics, $S_n(\tau)$, is defined as a weighted linear combination of penalized regression rank scores. In particular, we define $S_n(\tau)$ as

$$S_n(\tau) = n^{-1} \sum_{i \in [n]} Y_i \int_0^\tau \phi(\alpha) d\widehat{\xi}_i(\alpha), \tag{4.5}$$

where $\phi(t) = \text{sign}\left(t - \frac{1}{2}\right)$ is the scaled score function and the sign function $\text{sign}(x)$ is defined as 1 if $x \geq 0$ and $-1$ if $x < 0$. Notice that if $\phi$ is taken to be an identity function, then the scale rank statistic becomes a linear combination of penalized regression ranks scores and as such preserves a distribution free property.

We define the population quantity $S_F(\tau)$ as

$$S_F(\tau) = \int_0^\tau \phi(\alpha) F^{-1}(\alpha) d\alpha \tag{4.6}$$

and observe that the sparsity function $\varsigma(\tau)$ is related to $S_F(\tau)$ via the following equation

$$\varsigma(\tau) = \partial^2 S_F(\tau)/\partial \tau^2.$$

Later, we show $S_n(\tau) \xrightarrow{\mathbb{P}} S_F(\tau)$ as $n, p \to \infty$, even in the regime where $p \gg n$. This suggests the following estimator for the sparsity function

$$\widehat{\varsigma}(\tau) = h^{-2} \left( S_n(\tau + h) - 2 S_n(\tau) + S_n(\tau - h) \right). \tag{4.7}$$

In the display above, $h$ is an appropriately chosen bandwidth parameter that converges to zero. If the conditional quantile function is three times continuously differentiable, the estimator is based on the second order partial difference of the scale rank statistics $S_n(\tau)$ and it can be shown to have the bias of order $\mathcal{O}(h^2)$. The size of the bias can be improved by considering the following corrected estimator

$$\widetilde{\varsigma}(\tau) = \frac{24}{21} h^{-2} \left( S_n(\tau + h) - 2 S_n(\tau) + S_n(\tau - h) \right) - \frac{1}{14} h^{-2} \left( S_n(\tau + 2h) - 2 S_n(\tau) + S_n(\tau - 2h) \right), \tag{4.8}$$

whose bias is of the order $\mathcal{O}(h^4)$.

The following Lemma establishes a first order approximation of the scale rank statistics $S_n(\tau)$.

**Lemma 6.** *Assume that the conditions of Theorem 3.1 hold. Let $E_{i\tau} = u_i - F^{-1}(\tau)$, with $\mathbf{E}_\tau = (E_{1\tau}, \ldots, E_{n\tau})^T$ and let $\mathbf{w} \in \mathbb{R}^n$ be defined as*

$$\mathbf{w} = \left(\mathbf{I} - \mathbf{X}_S \left(\mathbf{X}_S^T \mathbf{X}_S\right)^{-1} \mathbf{X}_S^T\right) \mathbf{E}_\tau.$$

*Then the scale statistics can be represented as*

$$S_n(\tau) = n^{-1} \sum_{i \in [n]} w_i \phi\left(F(u_i)\right) \mathbb{1}\left\{F^{-1}(\tau) \leq u_i \leq F^{-1}(0)\right\} \\ + \mathcal{O}_P\left(\lambda_n \left(\sup_{\tau \in [\epsilon, 1-\epsilon]} ||\boldsymbol{\beta}^*(\tau)||_1\right) + s\sqrt{\frac{\log(p \vee n)}{n}}\right), \quad (4.9)$$

*uniformly for all $\tau \in [\epsilon, 1-\epsilon]$ and any $\epsilon \in (0,1)$.*

In particular, if $\sup_{\tau \in [\epsilon, 1-\epsilon]} ||\boldsymbol{\beta}^*(\tau)||_1 = \mathcal{O}(s)$ and $\lambda_n$ is chosen to be of the order $O(\sqrt{\frac{\log(p \vee n)}{n}})$ then the result of Lemma 6 implies

$$S_n(\tau) = n^{-1} \sum_{i \in [n]} w_i \phi\left(F(u_i)\right) \mathbb{1}\left\{F^{-1}(\tau) \leq u_i \leq F^{-1}(0)\right\} + o_P(1),$$

uniformly for all $\tau \in [\epsilon, 1-\epsilon]$ and $\epsilon \in (0,1)$. Hence, we have approximated a linear rank process $S_n(\tau)$ with a non-linear residual process. In a classical setting, similar results were obtain by Gutenbrunner and Jurečková (1992). In contrast to their result, we allow both $s$ and $p$ to grow with $n$. The proof is challenging due to the high-dimensional nature of the rank scores $\widehat{\boldsymbol{\xi}}(\tau)$. The proof has three parts. In the first part, we establish a connection between high-dimensional rank scores process and their oracle counterpart process. Moreover, this result is established uniformly and not pointwise. The second and third steps project the oracle process to the residual process $E_{i\tau}$ and control the size of the projection uniformly in $\tau$.

To show consistency of the estimator $\widehat{\varsigma}(\tau)$ we require the following regularity condition.

**(D$'$)** The density of the error, $f$, is three times continuously differentiable with the derivative $f'''$ bounded by a constant.

**Theorem 4.1.** *Assume that the conditions of Theorem 3.1 hold. In addition, assume that the condition (D$'$) is satisfied. Then the sparsity function estimator $\widehat{\varsigma}(\tau)$ in (4.7), satisfies*

$$\sup_{\epsilon \leq \tau \leq 1-\epsilon} |\widehat{\varsigma}(\tau) - \varsigma(\tau)| = \mathcal{O}_P\left(h^2 + h^{-2}\left(\lambda_n \left(\sup_{\tau \in [\epsilon, 1-\epsilon]} ||\boldsymbol{\beta}^*(\tau)||_1\right) + s\sqrt{\frac{\log(p \vee n)}{n}}\right)\right)$$

*for every $\epsilon \in (0,1)$.*

Theorem 4.1 gives us guidance on how to choose the bandwidth parameter that balances the bias and variance of the density estimator. Suppose that $\sup_{\tau \in [\epsilon, 1-\epsilon]} ||\boldsymbol{\beta}^*(\tau)||_1 = O(s)$. Then the choice of the bandwidth $h^4 = \mathcal{O}\left(s\sqrt{n^{-1}\log(p \vee n)}\right)$ gives us

$$\sup_{\epsilon \leq \tau \leq 1-\epsilon} |\widehat{\varsigma}(\tau) - \varsigma(\tau)| = \mathcal{O}_P\left(s^{1/2}\left(\frac{\log(p \vee n)}{n}\right)^{1/4}\right).$$

We are unaware of any existing result that establishes uniform rates of convergence for the sparsity function estimation. Belloni et al. (2013b) establish a point-wise convergence rate of the Koenker's density estimator that is of the order of $h^{-1}\sqrt{s\log(p\vee n)/n}+h^\kappa$ (see equation (C.48) therein). The parameter $\kappa$ depends on the initial estimator $\widehat{\boldsymbol{\beta}}(\tau)$. The main difference from their work is that we establish a uniform rate of convergence that neither depends on the initial estimator nor the quantile $\tau$. Lemma 4 provides an easy access to the asymptotic variance of the proposed estimator. It is easy to see that the asymptotic variance of $S_n(\tau)$ will be

$$\int_{F^{-1}(0)}^{F^{-1}(\tau)}\int_{F^{-1}(0)}^{F^{-1}(\tau)}\left(F(u\wedge v)-F(u)F(v)\right)\phi(F(u))\phi(F(v))dudv,$$

and hence is bounded. As $\widehat{\zeta}(\tau)$ is a linear functional of $S_n(\tau)$, its asymptotic variance is clearly bounded and is of the order of

$$\mathcal{O}\left(s^3\left(\frac{\log(p\vee n)}{n}\right)^{3/2}\right),$$

for the optimal choice of $h$ above.

### 4.3 Bahadur Representation Revisited

For the case of the unknown density function $f$, the debiased estimator, $\check{\boldsymbol{\beta}}(\tau)$, takes the form in (2.2). The following theorem provides Bahadur representation for the estimator $\check{\boldsymbol{\beta}}(\tau)$, when the sparsity function estimator $\widehat{\zeta}(\tau)$ takes the form in (4.7).

**Theorem 4.2.** *Suppose that conditions of Theorem 3.1 and Theorem 4.1 are satisfied. Moreover, suppose the sparsity function estimator $\widehat{\zeta}(\tau)$ defined in (4.7) be obtained with $h=c_4 s^{1/4}n^{-1/8}\log^{1/8}(p\vee n)$ where $c_4$ is a positive constant independent of $n$ and $p$. Then, the debiased estimator $\check{\boldsymbol{\beta}}(\tau)$ in (2.2) satisfies*

$$\sqrt{n}(\check{\boldsymbol{\beta}}(\tau)-\boldsymbol{\beta}^*(\tau))=n^{-1/2}\widehat{\mathbf{D}}_\tau\sum_{i\in[n]}\mathbf{X}_i\psi_\tau\left(u_i-F^{-1}(\tau)\right)$$
$$+\mathcal{O}_P\left(\frac{L_n s^{3/4}\log^{3/4}(p\vee n)}{n^{1/4}}\bigvee\frac{L_n s\log(p\vee n)}{n^{1/2}}\bigvee\frac{s\log^{3/4}(p\vee n)}{n^{1/4}}\right)$$
(4.10)

*uniformly in $\tau\in[\epsilon,1-\epsilon]$ for all $\epsilon>0$.*

A few comments are in order. Compared to the residual size of the Bahadur representation with known sparsity function, the case of the unknown sparsity function has a residual whose size contains an additional term of the order of $s\log^{3/4}(p\vee n)/n^{1/4}$. This term has an effect on the rate only when $\sqrt{\log(p\vee n)}<s^{1/4}$. Whenever, $\log(p\vee n)>\sqrt{s}$, the rates of convergence in Theorem 4.2 and 3.1 are the same and of the order of $s^{3/4}\log^{5/4}(p\vee n)/n^{1/4}$.

## 5 Confidence Intervals and Hypothesis Testing

In this section we provide uniform inference results on the quantile process $\boldsymbol{\beta}(\tau)$. Although other inference problems can be analyzed, we focus on those of paramount importance where $\boldsymbol{\beta}(\tau)$ is treated as a function of $\tau$.

## 5.1 Pointwise Confidence Intervals

Fix $\tau$ and $\alpha$ to be in the interval $(0,1)$ and let $z_\alpha$ denote the $(1-\alpha)$th standard normal percentile point. Let $\mathbf{x}$ be a fixed vector in $\mathbb{R}^{p+1}$. A $(1-2\alpha)100\%$ confidence interval for $\mathbf{x}^\top \beta(\tau)$ can be constructed as

$$\widehat{I}_n = \left( \mathbf{x}^\top \widecheck{\boldsymbol{\beta}}(\tau) - \widehat{a}_n, \mathbf{x}^\top \widecheck{\boldsymbol{\beta}}(\tau) + \widehat{a}_n \right), \tag{5.1}$$

where $\widecheck{\boldsymbol{\beta}}(\tau)$ is defined in (2.2) and

$$\widehat{a}_n = \widehat{\zeta}(\tau) z_\alpha \sqrt{\tau(1-\tau) \mathbf{x}^\top \widehat{\mathbf{D}} \widehat{\boldsymbol{\Sigma}} \widehat{\mathbf{D}} \mathbf{x}}/\sqrt{n},$$

with $\widehat{\mathbf{D}}$ defined in (2.4) and $\widehat{\zeta}(\tau)$ defined in (4.7). A confidence interval for one coordinate, $\beta_j(\tau)$, can be obtained by choosing $\mathbf{x}$ to be a standard basis vector. The following lemma provides coverage of $\widehat{I}_{1n}$.

**Lemma 7.** *Suppose that conditions of Theorem 3.1 and 4.1 hold. In addition, assume that $sM^4 \log(p) = o(n^{1/5})$, then for all $\tau \in (\epsilon, 1-\epsilon)$, $\epsilon \in (0,1)$ and all vectors $\mathbf{x}$, such that $\|\mathbf{x}\|_1 = o(n^{1/5})$ and $\|\mathbf{x}\|_2 = 1$, we have*

$$\sup_{\boldsymbol{\beta}(\tau): \|\boldsymbol{\beta}(\tau)\|_0 \leq s} \mathbb{P}_{\boldsymbol{\beta}(\tau)} \left( \mathbf{x}^\top \boldsymbol{\beta}(\tau) \in \widehat{I}_{1n} \right)$$

$$= 1 - \alpha + \mathcal{O}\left( \|\mathbf{x}\|_1 \cdot \Lambda_{\max}^{1/2}(\boldsymbol{\Sigma}) \cdot \left( n^{-1/2} L_n \vee n^{-1/4} s \log^{3/4}(p \vee n) \right) \bigvee \frac{L_n s^{3/4} \log^{3/4}(p \vee n)}{n^{1/4}} \bigvee \frac{L_n s \log(p \vee n)}{n^{1/2}} \right).$$

Lemma 7 indicates that the coverage errors of $\widehat{I}_{1n}$ are of the order $\mathcal{O}\left( \left( n^{-1} s^3 \log^5(p \vee n) \right)^{1/4} \right)$ whenever the constant $M$ and the dimension and size of the vector $\mathbf{x}$ are considered fixed. Classical results, with $p \ll n$, have the coverage error of the confidence interval as $\mathcal{O}\left( \left( n^{-1} \log^3(n) \right)^{1/4} \right)$ (Zhou and Portnoy, 1996). Hence, we obtain rates with an additional term that is a polynomial in $s$ and $\log(p \vee n)$. Controlling the coverage is difficult if one applies pointwise tests over a number of quantile indices. Thus, methods developed in Zhao et al. (2014) and Belloni et al. (2013b), which control coverage of single parameters, cannot guarantee coverage of the confidence intervals studied here and a new proof technique needs to be developed. The dimension of the loadings $\mathbf{x}$ is allowed to grow with $n$ and $p$, making the problem significantly more challenging from the existing literature on testing for one or simultaneous coordinates. Moreover, the size of the matrix $l_1$ norm, $M$, stated in the assumption (**X**) is also allowed to grow with $n$ and $p$.

We observe a particular phase transition. Whenever the loadings vector is sparse, with $\|\mathbf{x}\|_0 = \mathcal{O}(n^{1/5})$, and $\|\mathbf{x}\|_{\max} = \mathcal{O}(1)$, together with $M \leq C\bar{s}$ for some constant $C$, we have that the coverage error is of the order $\mathcal{O}\left( \max(\bar{s}^4, s^{3/4}) \log^{5/4}(p \vee n)/n^{1/4} \right)$. These conditions are satisfied for a wide range of seetings; for example, it is satisfied for a design with Toeplitz correlation matrix and whenever one is interested in testing for a linear combination of a sparse number of coordinates of the regression parameter.

Whenever the loading vector, $\mathbf{x}$ is non-sparse with $\|\mathbf{x}\|_0 = O(p)$, we see that the rates of convergence are of the order of $\mathcal{O}\left( \|\mathbf{x}\|_1 \bar{s} \log^{3/4}(p \vee n)/n^{1/4} \right)$, which can be substantially slower than the rates obtained for the sparse loadings above. The behavior of the problem is significantly different in the two regimes of the loading vector $\mathbf{x}$. An important example of a dense loading vector is a design vector, in which case the above confidence intervals provide intervals for a quantile regression function, $Q_\tau$ for a fixed $\tau$.

## 5.2 Uniform Confidence Bands

Let $\mathbf{e}$ be a unit vector in $\mathbb{R}^{p+1}$. We are interested in constructing confidence bands for regression quantile process $T_n$ defined as
$$T_n = \left\{ \mathbf{e}^\top \boldsymbol{\beta}(\tau), \tau \in \mathcal{T} \right\}$$
where $\mathcal{T}$ is a compact subset of $(0, 1)$. To that end, let
$$\nu_\alpha = \inf \left\{ t : \mathbb{P}\left( \sup_{0 < \tau < 1} |B(\tau)| \leq t \right) \geq 1 - \alpha \right\}$$
with $B(t)$ denoting the 1-dimensional Brownian bridge. Now the confidence band for $T_n$ can be constructed as
$$J_n = \left\{ \mathbf{e}^\top \check{\boldsymbol{\beta}}(\tau) \pm j_n(\tau), \tau \in \mathcal{T} \right\} \tag{5.2}$$
with
$$j_n(\tau) = \widehat{\zeta}(\tau) \nu_\alpha \sqrt{\tau(1-\tau) \mathbf{e}^\top \widehat{\mathbf{D}} \widehat{\boldsymbol{\Sigma}} \widehat{\mathbf{D}} \mathbf{e} / n}.$$

The following lemma characterizes the coverage.

**Lemma 8.** *Let $J_n$ be defined in (5.2) with $\mathbf{e}$ satisfying $\|\mathbf{e}\|_1 \leq C$ for a positive constant $C$. Under conditions of Theorems 3.1 and 4.1, for a constant $c > 0$, we have*
$$\sup_{\boldsymbol{\beta}(\tau): \|\boldsymbol{\beta}(\tau)\|_0 \leq s} \mathbb{P}_{\boldsymbol{\beta}(\tau)} (T_n \in J_n) = 1 - 2\alpha + \mathcal{O}\left(n^{-c}\right).$$

Observe that, whenever the design vectors $\mathbf{X}_i$ are such that the conditions of Lemma 8 are satisfied, the interval $J_n$ provides a valid confidence interval for the whole quantile regression function. In the case of random designs, such conditions are satisfied for independent Gaussian designs with variance of the order of $1/\sqrt{p}$ or Gaussian designs with sparse correlation matrix. In the case of fixed designs, conditions will be met whenever a design is a sparse matrix itself.

Next, we present a uniform convergence result that allows for an asymptotics of growing number of linear combination of high-dimensional quantile estimators.

**Lemma 9.** *Let $\mathbf{V} \in \mathbb{R}^{d \times (p+1)}$ be a fixed matrix with $\|\mathbf{V}\|_F = d$. Suppose that $d\bar{s}^2 M^3 \sqrt{\log(p \vee n)/n} = o(1)$ and $d^7 \log^3(p \vee n) = o(n^{1-\delta})$ for some $\delta \in (0, 1)$. Under conditions of Theorems 3.1 and 4.1 when $n \to \infty$ we have the following:*
$$\sup_{\tau \in \mathcal{T}} \sqrt{n} \left\| \left( \mathbf{V} \widehat{\mathbf{D}} \widehat{\boldsymbol{\Sigma}} \widehat{\mathbf{D}} \mathbf{V}^\top \right)^{-1/2} \mathbf{V} \left( \check{\boldsymbol{\beta}}(\tau) - \boldsymbol{\beta}(\tau) \right) \right\|_2 \xrightarrow{\mathbb{D}} \sup_{\tau \in \mathcal{T}} \|B_d(\tau)\|_2, \tag{5.3}$$
*with $B_d(t)$ denoting the $d$-dimensional Brownian bridge and $\xrightarrow{\mathbb{D}}$ denoting convergence in distribution.*

Lemma 9 is useful for establishing our next task, where we construct simultaneous uniform confidence bands with respect to both the quantile of interest and the loading vector. Let $\mathbf{w}$ denote the loading vector of interest and define the set

$$\mathcal{D} = \left\{ \mathbf{w} = (1, \widetilde{\mathbf{w}}), \widetilde{\mathbf{w}} \in \mathbb{R}^{p-1} : \|\widetilde{\mathbf{w}}\|_0 \leq d \text{ and } \widetilde{\mathbf{w}}^\top \mathbf{\Sigma}^{-1} \widetilde{\mathbf{w}} \leq K \right\}$$

for a constant $K > 0$ that is independent of $n$. Moreover, define the confidence interval of interest

$$M_n = \left\{ \mathbf{w}^\top \boldsymbol{\beta}(\tau), \tau \in \mathcal{T}, \mathbf{w} \in \mathcal{D} \right\}, \tag{5.4}$$

and its candidate estimator as

$$K_n = \left\{ \mathbf{w}^\top \check{\boldsymbol{\beta}}(\tau) \pm l_n(\tau), \tau \in \mathcal{T}, \mathbf{w} \in \mathcal{D} \right\}, \tag{5.5}$$

where $l_n = \widehat{\zeta}(\tau) \nu_\alpha \sqrt{\tau(1-\tau) \mathbf{w}^\top \widehat{\mathbf{D}} \widehat{\mathbf{\Sigma}} \widehat{\mathbf{D}} \mathbf{w}} / \sqrt{n}$ and $\nu_\alpha = \inf \{ t : \mathbb{P} (\sup_{0 < \tau < 1} \|B_d(\tau)\|_2 \leq t) \geq 1 - \alpha \}$, with $B_d(t)$ denoting the $d$-dimensional Brownian bridge, is a simultaneous uniform confidence band for $M_n$.

**Lemma 10.** *Let $M_n$ and $K_n$ be defined in* (5.4) *and* (5.5). *Under conditions of Lemma 9, for all $\tau \in \mathcal{T}$ when $n \to \infty$ we have the following*

$$\sup_{\boldsymbol{\beta}(\tau):\|\boldsymbol{\beta}(\tau)\|_0 \leq s} \mathbb{P}_{\boldsymbol{\beta}(\tau)} (M_n \in K_n) = 1 - 2\alpha + \mathcal{O}\left(n^{-c}\right), \qquad \text{for } c \in (0, 1).$$

Lemma 10 provides validity of a set of uniform confidence bands over the range of quantile values and designs when $p \geq n$. Differently from Lemma 8 we observe that the width of the pessimistic interval of Lemma 10 depends on the $l_2$ quantile of a $d$-dimensional Brownian bridge, whereas the interval of Lemma 8 depends on a quantile of a univariate Brownian bridge. We observe that the conditions imposed are matching those of the single-parameter testing. Much like Corollaries 3.2 and 3.3 of Zhou and Portnoy (1996), which give a uniform confidence band for regression quantile in case of $p \leq n$, Lemmas 8 and 10 provide confidence bands that are uniform in $\tau$ and simultaneously uniform in the loading vectors, respectively, when $p \geq n$.

## 5.3 Linear Testing

Let $\mathbf{M} \in \mathbb{R}^{d \times (p+1)}$ as defined in Lemma 9. We allow for $d$ to depend and grow with the sample size $n$. We are interested in testing

$$H_0 : \mathbf{M}\boldsymbol{\beta}(\tau) = r \qquad \text{vs} \qquad H_a : \mathbf{M}\boldsymbol{\beta}(\tau) \neq r,$$

for $r \in \mathbb{R}^d$ known and for possibly a wide range of quantile $\tau$ values, $\tau \in \mathcal{T}$. The linear hypotheses above are defined in abstract terms, however they include many cases of interests that haven't been addressed in the existing literature. Among others, they include single-parameter testing, simultaneous testing, and a test of linear combination of many parameters of interest. An example, with a particular choice of $\mathbf{M}$ and $r$, leads to a test for structural changes (Qu, 2008) in the quantile process

$$H_0 : \beta_k(\tau) = \beta_j(\tau), \qquad k \neq j$$

for a fixed value of $\tau \in (0,1)$. A different choice of $\mathbf{M}$ and $r$ leads to a test based on indefinite integrals of the process $\boldsymbol{\beta}(\tau)$

$$H_0 : \int_0^1 \beta_k(\tau)d\tau > \int_0^1 \beta_j(\tau)d\tau,$$

(see for example Qu and Yoon (2015)). Another specifically important class of examples include the partial orderings of conditional distributions using stochastic dominance. The simplest examples is a treatment control model where the test above can be used to test wheather

$$H_0 : \boldsymbol{\beta}_J(\tau) > 0, \qquad \tau \in \mathcal{T}, J \subset \{1,\ldots,p\}, |J| \leq q,$$

that is, to test weather the treatment distribution stochastically dominates the control distribution (Pin Ng, 2011). Lastly, our setup includes testing the complete quantile regression function, for example a test of the following kind

$$H_0 : Q_\tau(Y|\mathbf{X}_i) > 0,$$

is included in the setting above. It is worth pointing that testing complete quantile regression function, although a problem of a great practical interest and importance, has not been covered in the existing literature of high-dimensional models.

Under the linear hypothesis $H_0$, for any fixed $\tau$, we construct a regression Wald test statistics as follows

$$W_n(\tau) = n(\mathbf{M}\widehat{\boldsymbol{\beta}}(\tau) - r)^\top \left[\tau(1-\tau)\mathbf{M}\widehat{\mathbf{D}}_\tau \widehat{\boldsymbol{\Sigma}}\widehat{\mathbf{D}}_\tau^\top \mathbf{M}^\top\right]^{-1} (\mathbf{M}\widehat{\boldsymbol{\beta}}(\tau) - r).$$

Additionally, when performing the test for a range of values $\tau \in \mathcal{T} \subset (0,1)$, we look at a Kolmogorov-Smirnov test, or a sup-Wald test, defined as

$$\sup_{\tau \in \mathcal{T}} W_n(\tau).$$

We note that the Kolmogorov-Smirnov test is more appropriate for general alternatives and outperforms $t$ test for a variety of non-gaussian error distributions (Koenker and Xiao, 2002). Next, we present theoretical properties of the two tests introduced above.

**Theorem 5.1.** *Suppose that conditions of Theorems 3.1 and 4.1, as well as those of Lemmas 7 - 10 are satisfied. Then,*

(a) *Under the null hypothesis $H_0$ and a fixed value of $\tau$, we have*

$\sup_z \left|\mathbb{P}\left(W_n(\tau) \leq z\right) - \mathbb{P}\left(\chi_d^2 \leq z\right)\right| = o(1),$

(b) *Under the null hypothesis $H_0$ and a range of values of $\tau \in \mathcal{T} \subset (0,1)$, we have*

$\sup_z \left|\mathbb{P}\left(\sup_{\tau \in \mathcal{T}} W_n(\tau) \leq z\right) - \mathbb{P}\left(\sup_{\tau \in \mathcal{T}} Q_d^2(\tau) \leq z\right)\right| = o(1),$

*where $Q_d(\tau) = \|B_d(\tau)\|/\sqrt{\tau(1-\tau)}$ is a Bessel process of order $d$, $B_d(t)$ denotes a d-dimensional Brownian bridge and $\|\cdot\|$ stands for the Euclidean norm.*

Part (a) of Theorem 5.1 provides a way to approximate the quantiles of the Wald test statistic for a fixed value of $\tau$. Part (b) establishes a much stronger result, which approximates quantiles of the sup-Wald test statistics.

# 6 Numerical Experiments

In this section we provide Monte-Carlo simulations to illustrate finite sample properties of the confidence intervals constructed in Section 5. Data are generated from a model in (1.1) with $(n, p) = (1000, 1500)$. The number of non-zero coefficient of $\boldsymbol{\beta}^*$ is $s = 10$ with $\text{supp}(\boldsymbol{\beta}^*) = \{1, \ldots, 10\}$. The intercept is $\beta_0^* = 0$ and the non-zero coefficients are $\beta_j^* = 1 - 1/18 * (j - 1)$ for $j = 1, \ldots, 10$. Because of the decay of the coefficients, the small coefficients cannot be recovered reliably. However, as we illustrate below, our procedure is robust to model selection mistakes. Each $\mathbf{X}_i$ is drawn independently from $N(\mathbf{0}, \boldsymbol{\Sigma})$ where the covariance $\boldsymbol{\Sigma}$ is one of the following: an equi-correlation matrix $\boldsymbol{\Sigma} = (1 - \rho)I + \rho \mathbf{1}_p \mathbf{1}_p^T$ with $\rho = 0.5$ or a Toeplitz matrix with elements $(\sigma_{ab})_{ab}$ where $\sigma_{ab} = \rho^{|a-b|}$ with $\rho = 0.1$. We also illustrate three quantile settings $\tau = 0.3, 0.5$ and $0.6$. The noise $u_i$ follows one of the following distributions: standard Normal distribution or t-distribution with one degrees of freedom. Empirical coverage of constructed confidence intervals is reported over 500 independent simulation runs.

In Figures 1 - 6 we present distribution of our test statistic under the null under various quantile levels and distributional errors. We observe a persistently good behavior of our test statistics and its closeness to the theoretical null distribution - both when the error is normal or extremely heavy tailed with only existence of the first moment. Moreover, we compared the distribution also with the oracle one, that is the distribution of the Wald test statistics composed of an estimator using only the true non-zero indices. What we observe is that the effect of a large number of nuisance parameters is negligible and that the two distributions match. Additionally, the distribution doesn't vary much across different coordinates of parameter beta - both signal and noise variables behave uniformly well. However we observe that our $p$ and $n$ needed to be large in order to see asymptotic in the finite samples – this may be due to the non-smoothness of the loss function. Lastly, we see robustness of the distribution with respect to the choice of the quantile value $\tau$ as well.

Next, we perform a finite sample analysis of the newly proposed robust density estimator, i.e. sparsity function estimator. Figure 7 collects the results of our analysis where the black line presents the true sparsity function and the gray lines present 100 replications of the estimator $\widehat{S}(\tau)$. Figure 7 presents an estimator $\widehat{S}(\tau)$ as a function of the quantiles $\tau$ and over a range of bandwidth choices $h$. Moreover we show results of the estimation of a density of a standard normal distribution and Student $t_1$ distribution. It is worth pointing that the estimator of the latter is extremely difficult due to its heavy-tailed properties. However, Figure 7 presents remarkable results and a consistency of estimation of the underlying sparsity function $1/f(F^{-1}(\tau))$.

We also present a power curve of the Wald test developed in Section 5 for testing the null. Results are presented in Figure 8. We show plots for light-tailed and heavy-tailed error and observe that the test reaches power very quickly for both settings and preserves Type I error simultaneously.

Lastly, we present coverage probability of the proposed confidence intervals. Results are presented in Table 1. We observe that the coverage probabilities are extremely close to the nominal 95% both for the strong signal, weak signal and noise variables. Overall coverage is not disturbed by a change in the error distribution – going from light to heavy tails – or a change in the choice of the quantile $\tau$.

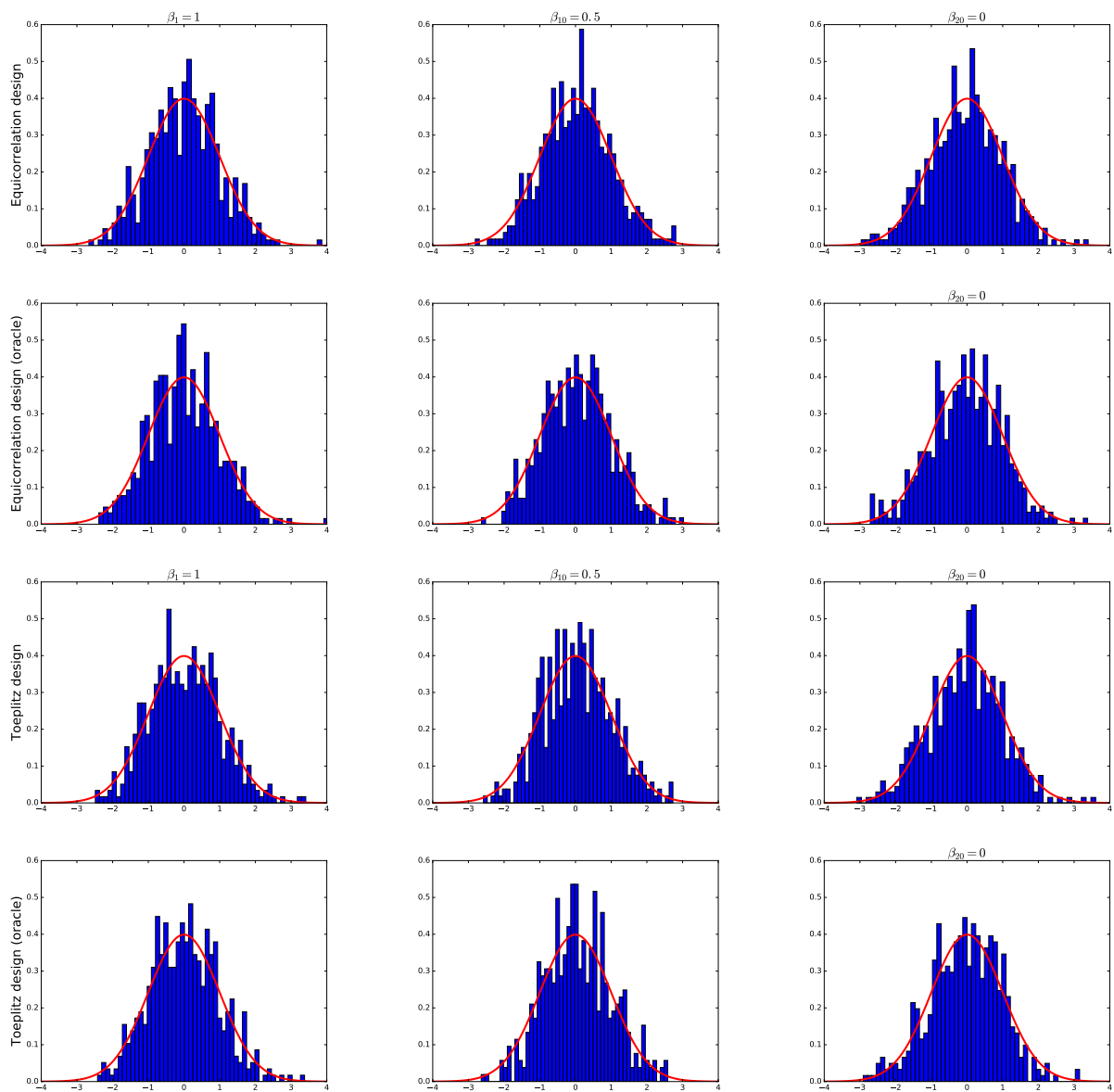

Figure 1: Histogram of $\widehat{\sigma}_j^{-1} \cdot \sqrt{n}\left(\check{\beta}_j - \beta_j\right)$ under equi-correlation and Toeplitz design with Gaussian noise and $\tau = 0.5$. The oracle procedure solves the quantile regression problem using a subset of predictor $\{1, \ldots, 10\} \cup \{j\}$. The red curve indicates the density of a standard Normal random variable centered.

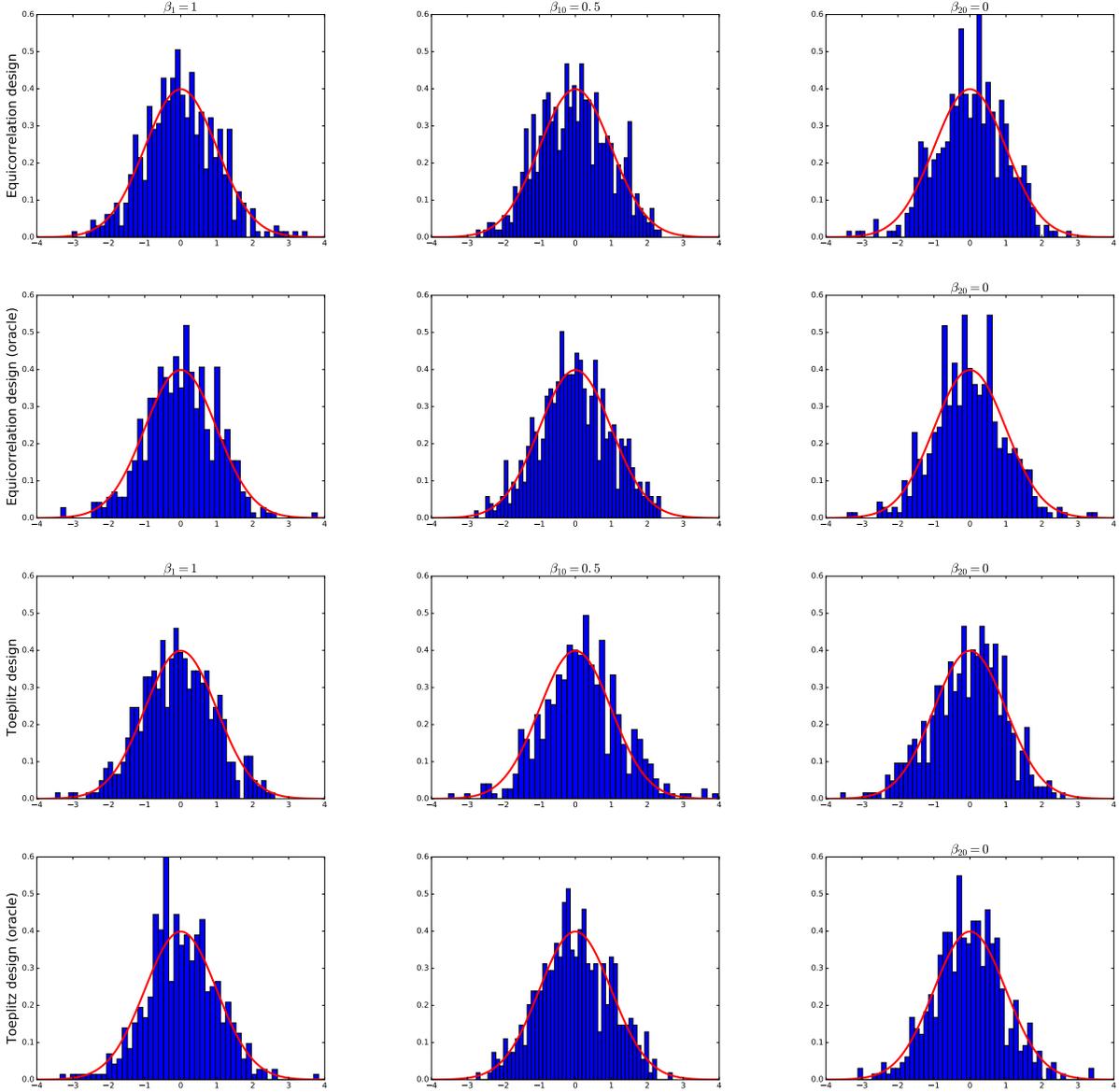

Figure 2: Histogram of $\widehat{\sigma}_j^{-1} \cdot \sqrt{n} \left( \check{\beta}_j - \beta_j \right)$ under equi-correlation and Toeplitz design with $t_1$ distributed noise and $\tau = 0.5$. The oracle procedure solves the quantile regression problem using a subset of predictor $\{1, \ldots, 10\} \cup \{j\}$. The red curve indicates the density of a standard Normal random variable centered.

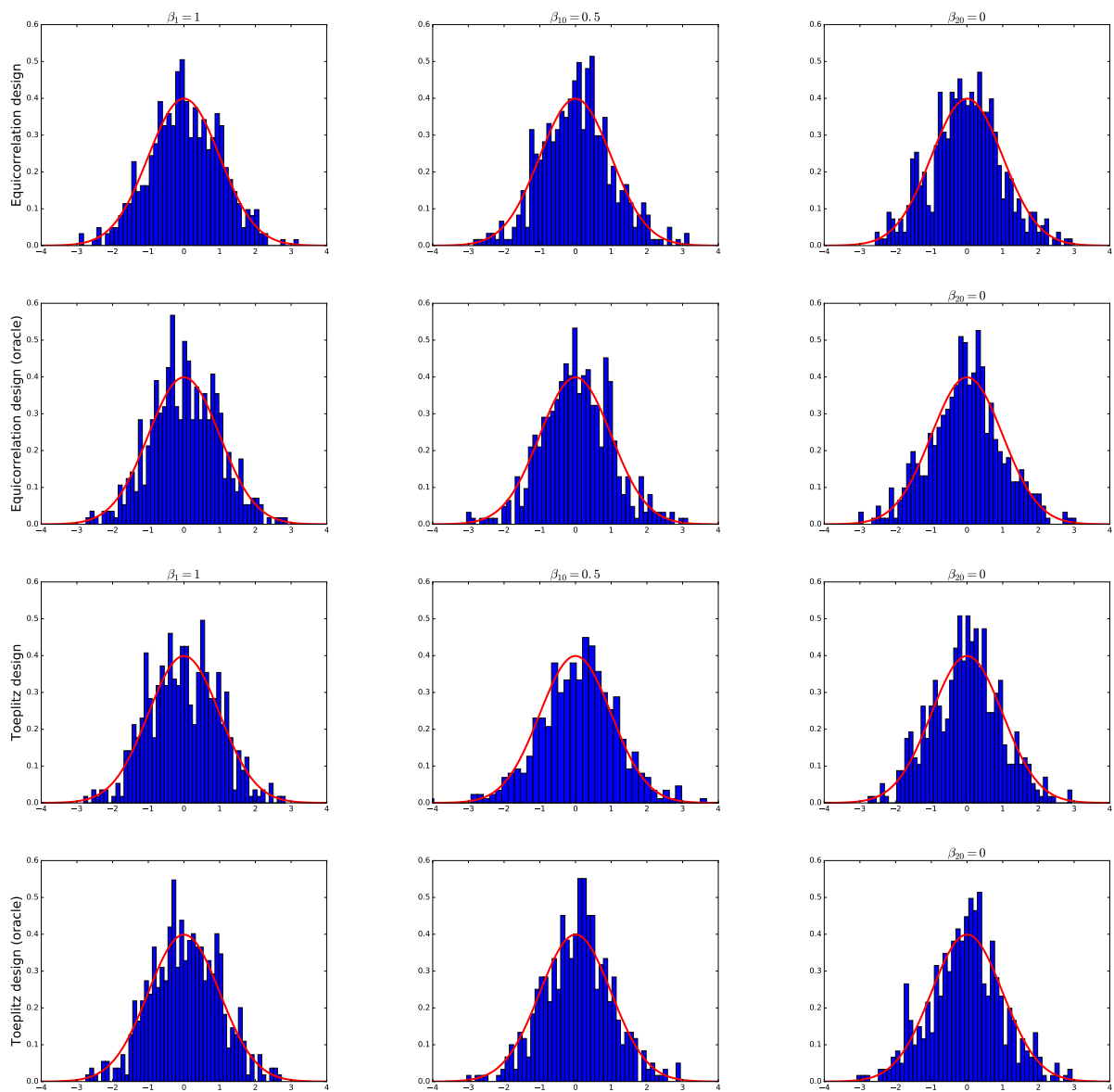

Figure 3: Histogram of $\widehat{\sigma}_j^{-1} \cdot \sqrt{n}\left(\check{\beta}_j - \beta_j\right)$ under equi-correlation and Toeplitz design with Gaussian noise and $\tau = 0.3$. The oracle procedure solves the quantile regression problem using a subset of predictor $\{1, \ldots, 10\} \cup \{j\}$. The red curve indicates the density of a standard Normal random variable centered.

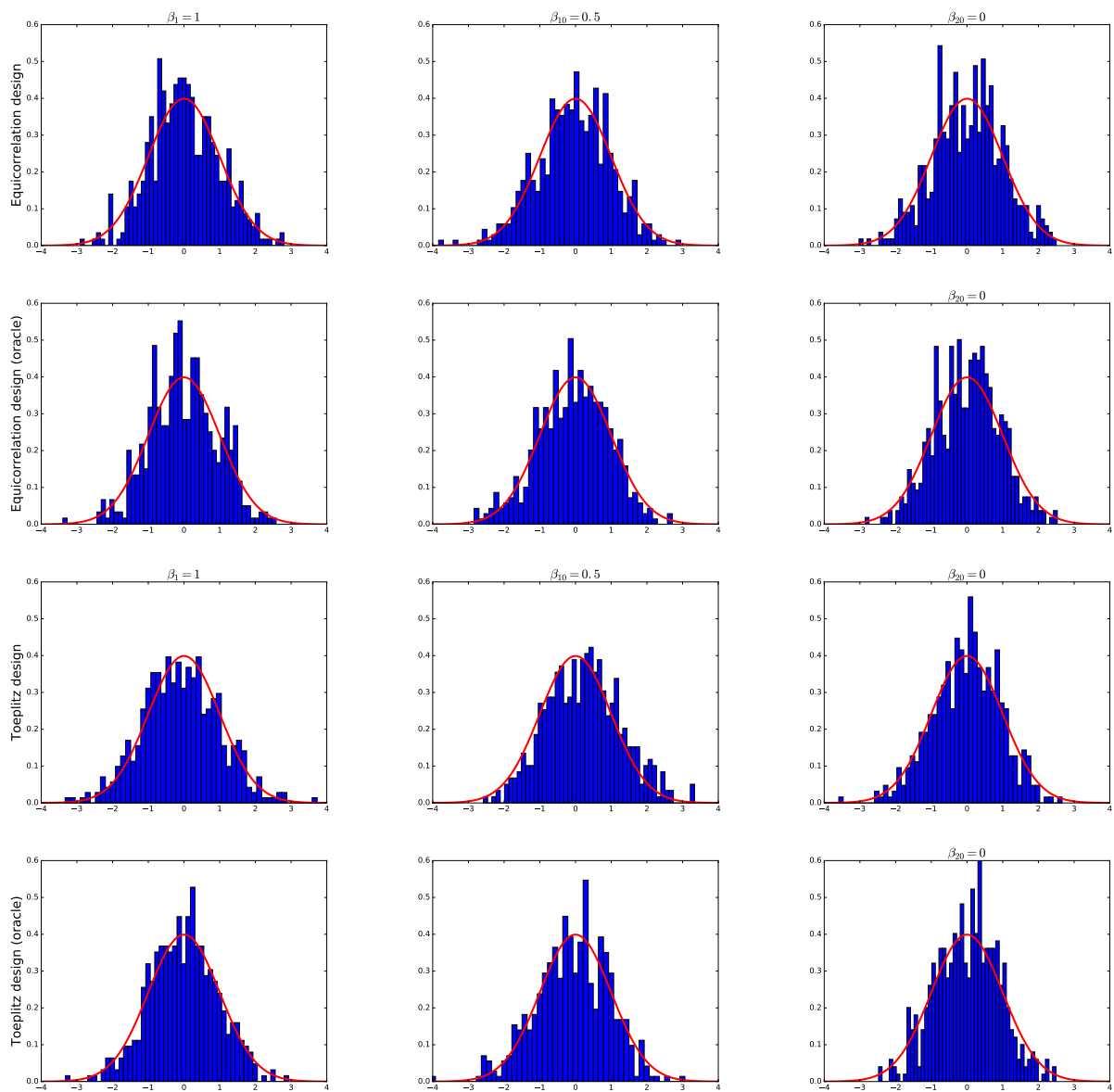

Figure 4: Histogram of $\widehat{\sigma}_j^{-1} \cdot \sqrt{n}\left(\check{\beta}_j - \beta_j\right)$ under equi-correlation and Toeplitz design with $t_1$ distributed noise and $\tau = 0.3$. The oracle procedure solves the quantile regression problem using a subset of predictor $\{1, \ldots, 10\} \cup \{j\}$. The red curve indicates the density of a standard Normal random variable centered.

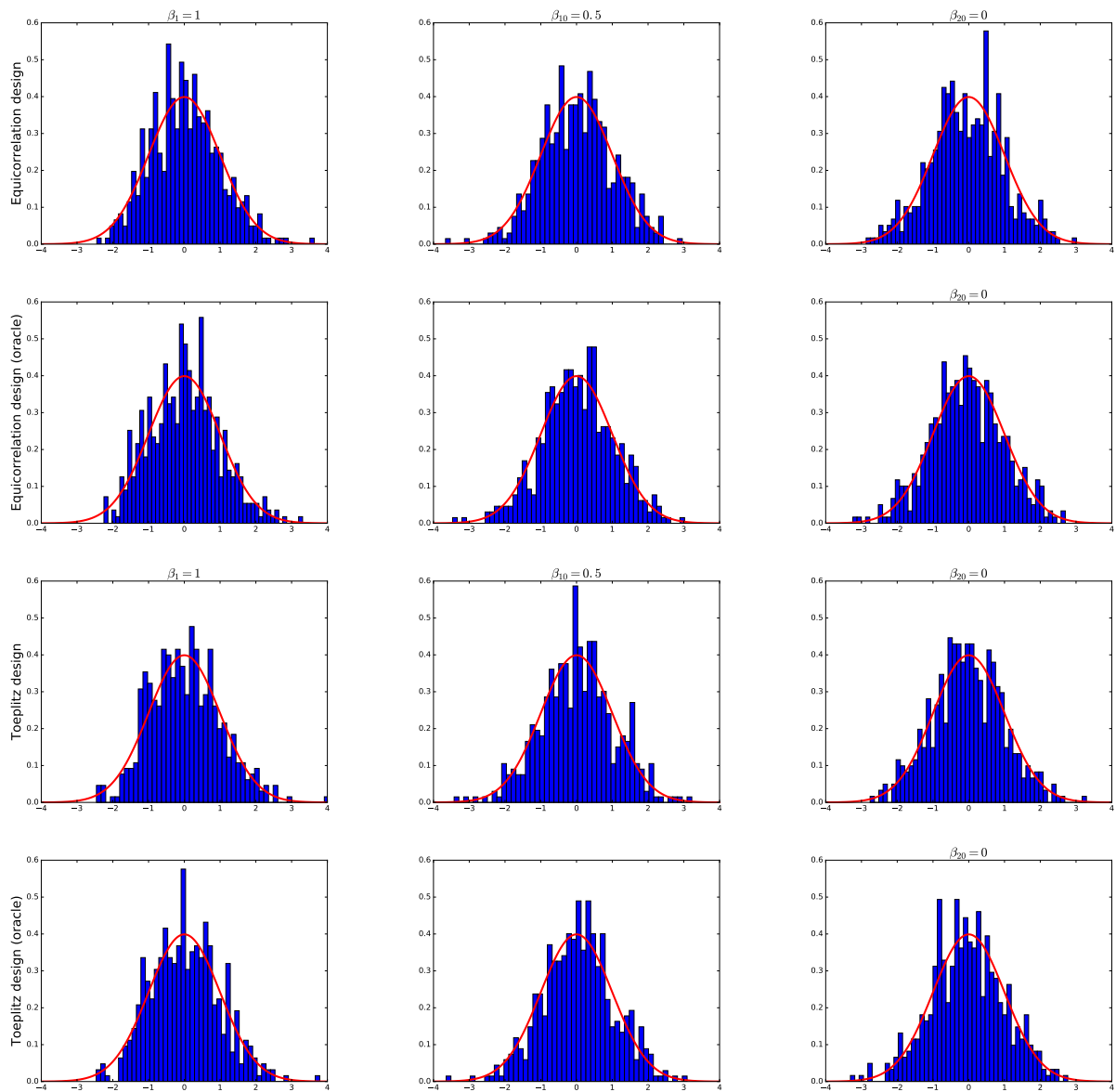

Figure 5: Histogram of $\widehat{\sigma}_j^{-1} \cdot \sqrt{n}\left(\check{\beta}_j - \beta_j\right)$ under equi-correlation and Toeplitz design with Gaussian noise and $\tau = 0.6$. The oracle procedure solves the quantile regression problem using a subset of predictor $\{1, \ldots, 10\} \cup \{j\}$. The red curve indicates the density of a standard Normal random variable centered.

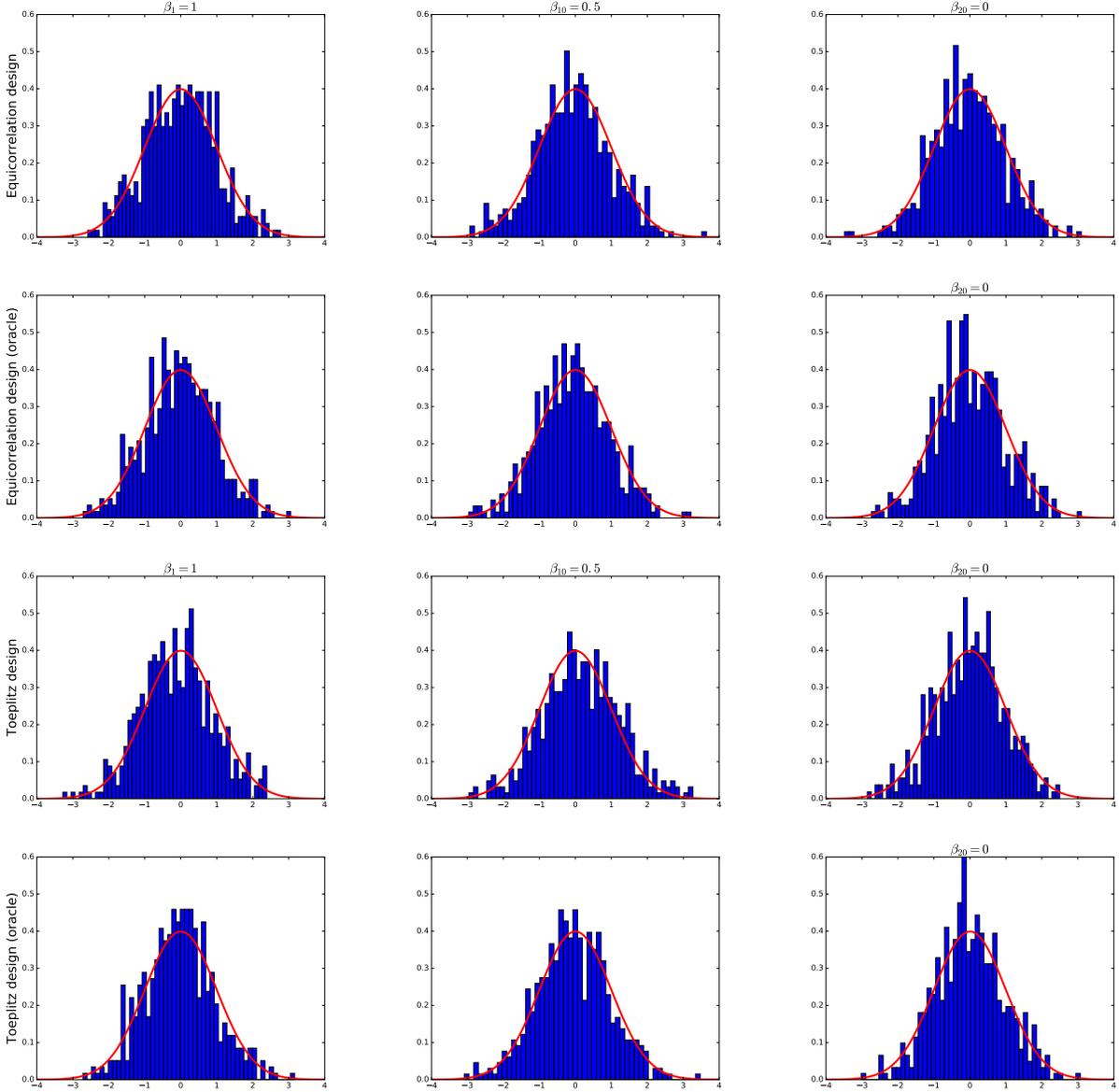

Figure 6: Histogram of $\widehat{\sigma}_j^{-1} \cdot \sqrt{n}\left(\check{\beta}_j - \beta_j\right)$ under equi-correlation and Toeplitz design with $t_1$ distributed noise and $\tau = 0.6$. The oracle procedure solves the quantile regression problem using a subset of predictor $\{1, \ldots, 10\} \cup \{j\}$. The red curve indicates the density of a standard Normal random variable centered.

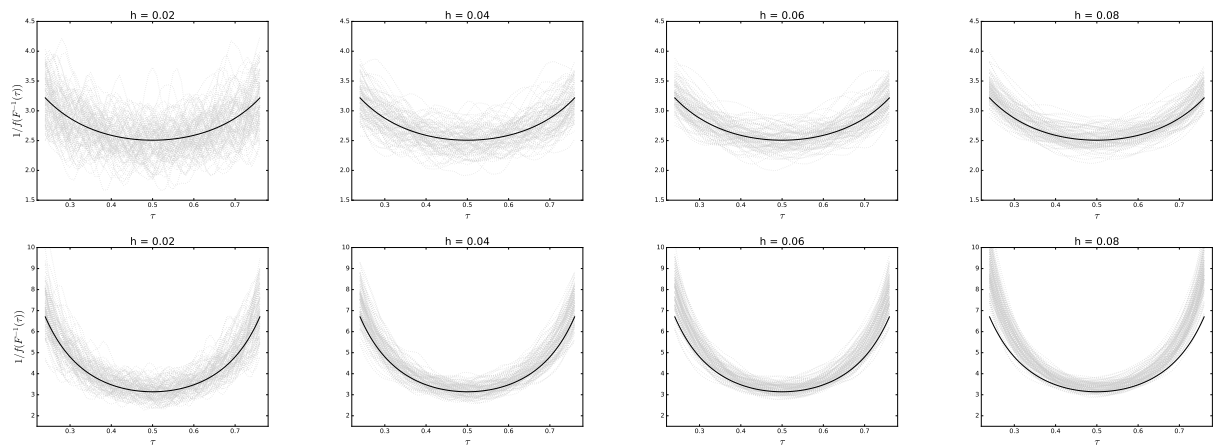

Figure 7: Robust Sparsity Function Estimator as a function of $\tau$. The first row is for the Normal Errors and the second is for the Student t Errors.

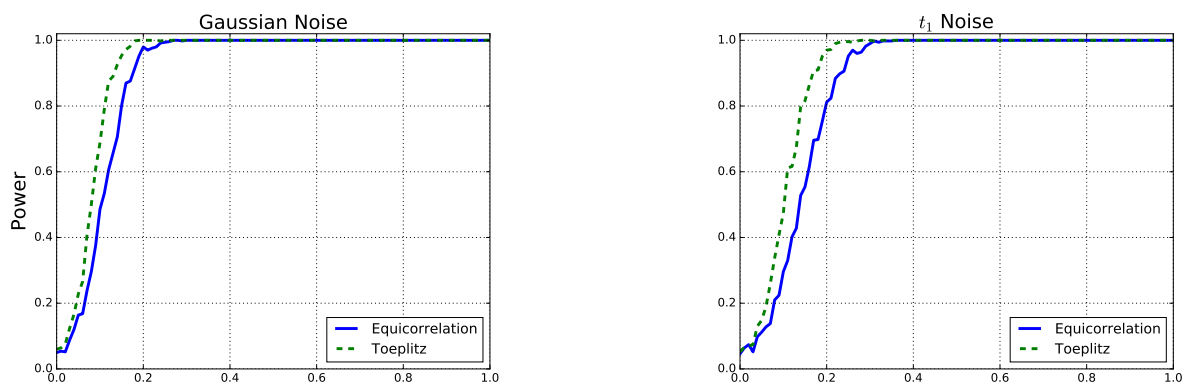

Figure 8: Power curves of the Linear Tests with Normal and Student $t$ Error Distributions

|  | Setting | $\beta_1$ | | $\beta_{10}$ | | $\beta_{20}$ | |
|---|---|---|---|---|---|---|---|
| $\tau = 0.3$ | EQ, $N(0,1)$ | 95.4 | (0.45) | 94.8 | (0.45) | 94.6 | (0.22) |
| | Toeplitz, $N(0,1)$ | 96.2 | (0.26) | 93.2 | (0.26) | 96.8 | (0.26) |
| | EQ, $t_1$ | 94.8 | (1.50) | 95.2 | (1.50) | 95.6 | (0.75) |
| | Toeplitz, $t_1$ | 93.6 | (1.90) | 93.0 | (1.92) | 96.6 | (1.92) |
| Oracle | EQ, $N(0,1)$ | 95.8 | (0.41) | 94.8 | (0.41) | 94.6 | (0.23) |
| | Toeplitz, $N(0,1)$ | 95.8 | (0.23) | 96.2 | (0.23) | 94.4 | (0.23) |
| | EQ, $t_1$ | 96.2 | (1.34) | 94.8 | (1.34) | 97.4 | (0.75) |
| | Toeplitz, $t_1$ | 96.2 | (0.75) | 95.6 | (0.75) | 97.2 | (0.75) |
| $\tau = 0.5$ | EQ, $N(0,1)$ | 96.2 | (0.40) | 95.8 | (0.40) | 95.4 | (0.20) |
| | Toeplitz, $N(0,1)$ | 96.2 | (0.20) | 95.8 | (0.20) | 94.8 | (0.20) |
| | EQ, $t_1$ | 94.8 | (0.68) | 96.2 | (0.68) | 97.0 | (0.34) |
| | Toeplitz, $t_1$ | 94.8 | (0.73) | 93.2 | (0.73) | 94.8 | (0.73) |
| Oracle | EQ, $N(0,1)$ | 96.6 | (0.36) | 96.8 | (0.36) | 94.4 | (0.20) |
| | Toeplitz, $N(0,1)$ | 96.2 | (0.20) | 96.0 | (0.20) | 94.8 | (0.20) |
| | EQ, $t_1$ | 96.0 | (0.61) | 95.0 | (0.61) | 96.2 | (0.34) |
| | Toeplitz, $t_1$ | 95.8 | (0.34) | 94.6 | (0.34) | 95.6 | (0.34) |
| $\tau = 0.6$ | EQ, $N(0,1)$ | 96.2 | (0.41) | 95.0 | (0.41) | 93.0 | (0.20) |
| | Toeplitz, $N(0,1)$ | 95.4 | (0.21) | 94.2 | (0.21) | 94.8 | (0.21) |
| | EQ, $t_1$ | 95.4 | (0.82) | 93.6 | (0.82) | 95.6 | (0.41) |
| | Toeplitz, $t_1$ | 94.6 | (0.95) | 92.2 | (0.96) | 95.2 | (0.96) |
| Oracle | EQ, $N(0,1)$ | 95.8 | (0.37) | 94.8 | (0.37) | 94.0 | (0.21) |
| | Toeplitz, $N(0,1)$ | 96.2 | (0.21) | 95.0 | (0.21) | 95.0 | (0.21) |
| | EQ, $t_1$ | 94.6 | (0.74) | 95.2 | (0.74) | 95.4 | (0.41) |
| | Toeplitz, $t_1$ | 95.8 | (0.41) | 94.6 | (0.41) | 96.2 | (0.41) |

Table 1: Coverage of the confidence intervals and $\sqrt{n}$ times the width of the confidence interval (reported in parenthesis).

# 7 Discussion

We aim to provide a rigorous statistical approach to formalizing confidence sets and hypothesis testing statistics for all quantiles at the percentile level in a predetermined set $\mathcal{T}$. Our work does not address how $\mathcal{T}$ should be chosen because the choice of $\mathcal{T}$ should align with the scientific problem at hand. In practice, $\mathcal{T}$ can be chosen as a large interval say $[0.1, 0.9]$ to reflect an interest in normal outcomes or a smaller interval, say $[0.75, 0.9]$ to reflect our interest in the upper tail of the response distribution. Even in the case where one is interested in a single quantile level $\tau$, our work suggests that applying the proposed method with $\mathcal{T}$ chosen as a small window interval containing $\tau$ leads to a more stable confidence sets when the sample size is limited. In practice, the scientist may have more than one reasonable choice of $\tau$. For example, choice of $[0.7, 0.9]$ and $[0.72, 0.88]$ should not induce huge changes in the testing decision and confidence intervals.

We showcase a particular transition phase for linear testing which is affected by the properties of the loading vector of the linear hypothesis. If it is a sparse vector we obtain rates that are in line with classical results, however when the loading vector is a dense vector we obtain significantly slower rates of convergence. As linear tests are practically very important it is of interest to present full analysis of their properties. For example, our results apply for construction of confidence intervals of a complete quantile regression line, a result we believe is new in existing high-dimensional literature.

## Acknowledgments

This work was supported in part by an NSF grant 1205296. This work is also supported by an IBM Corporation Faculty Research Fund at the University of Chicago Booth School of Business. This work was completed in part with resources provided by the University of Chicago Research Computing Center.

## References

R. R. Bahadur. A note on quantiles in large samples. *Ann. Math. Statist.*, 37:577–580, 1966.

R. F. Barber and M. Kolar. Rocket: Robust confidence intervals via kendall's tau for transelliptical graphical models. *ArXiv e-prints, arXiv:1502.07641*, 2015.

A. Belloni, V. Chernozhukov, D. Chetverikov, and I. Fernandez-Val. Conditional Quantile Processes based on Series or Many Regressors. *ArXiv e-prints*, 2011.

A. Belloni, V. Chernozhukov, and K. Kato. Valid Post-Selection Inference in High-Dimensional Approximately Sparse Quantile Regression Models. *ArXiv e-prints*, 2013.

A. Belloni and V. Chernozhukov. $\ell_1$-penalized quantile regression in high-dimensional sparse models. *Ann. Stat.*, 39(1):82–130, 2011.

A. Belloni, V. Chernozhukov, and C. B. Hansen. Inference on treatment effects after selection amongst high-dimensional controls. *Rev. Econ. Stud.*, 81(2):608–650, 2013a.

A. Belloni, V. Chernozhukov, and K. Kato. Robust inference in high-dimensional approximately sparse quantile regression models. *arXiv preprint arXiv:1312.7186*, 2013b.

A. Belloni, V. Chernozhukov, and K. Kato. Uniform post selection inference for lad regression models. *arXiv preprint arXiv:1304.0282*, 2013c.

A. Belloni, V. Chernozhukov, and Y. Wei. Honest confidence regions for logistic regression with a large number of controls. *arXiv preprint arXiv:1304.3969*, 2013d.

J. Bradic. Support recovery via weighted maximum-contrast subagging. *ArXiv e-prints, arXiv:1306.3494*, 2013.

T. T. Cai, W. Liu, and X. Luo. A constrained $\ell_1$ minimization approach to sparse precision matrix estimation. *J. Am. Stat. Assoc.*, 106(494):594–607, 2011.

T. T. Cai, W. Liu, and H. H. Zhou. Estimating sparse precision matrix: Optimal rates of convergence and adaptive estimation. *aos*, 44(2):455–488, 2016.

A. Chatterjee and S. N. Lahiri. Rates of convergence of the adaptive LASSO estimators to the oracle distribution and higher order refinements by the bootstrap. *Ann. Stat.*, 41(3):1232–1259, 2013.

Z. Chen and C. Leng. Dynamic covariance models. *Journal of the American Statistical Association*, 111(515):1196–1207, 2016.

V. Chernozhukov, D. Chetverikov, and K. Kato. Gaussian approximations and multiplier bootstrap for maxima of sums of high-dimensional random vectors. *Ann. Stat.*, 41(6):2786–2819, 2013.

M. Csörgő and P. Révész. Two approaches to constructing simultaneous confidence bounds for quantiles. *Probab. Math. Statist.*, 4(2):221–236, 1984.

J. Fan and R. Li. Variable selection via nonconcave penalized likelihood and its oracle properties. *J. Am. Stat. Assoc.*, 96(456):1348–1360, 2001.

J. Fan and J. Lv. Nonconcave penalized likelihood with np-dimensionality. *IEEE Trans. Inf. Theory*, 57(8):5467–5484, 2011.

J. Fan, L. Xue, and H. Zou. Strong oracle optimality of folded concave penalized estimation. *Ann. Stat.*, 42(3):819–849, 2014.

M. H. Farrell. Robust inference on average treatment effects with possibly more covariates than observations. *arXiv preprint arXiv:1309.4686*, 2013.

W. Feller. On the berry-esseen theorem. *Zeitschrift für Wahrscheinlichkeitstheorie und Verwandte Gebiete*, 10(3):261–268, 1968.

E. Gautier and A. B. Tsybakov. High-dimensional instrumental variables regression and confidence sets. *arXiv preprint arXiv:1105.2454*, 2011.

E. Gautier and A. B. Tsybakov. Pivotal estimation in high-dimensional regression via linear programming. In *Empirical inference*, pages 195–204. Springer, Heidelberg, 2013.

C. Gutenbrunner and J. Jurečková. Regression rank scores and regression quantiles. *Ann. Stat.*, 20(1):305–330, 1992.

P. Hall. *The bootstrap and Edgeworth expansion*. Springer Series in Statistics. Springer-Verlag, New York, 1992.

A. Javanmard and A. Montanari. Confidence intervals and hypothesis testing for high-dimensional regression. *J. Mach. Learn. Res.*, 15(Oct):2869–2909, 2014.

A. Juditsky, F. Kilinç Karzan, A. Nemirovski, and B. Polyak. Accuracy guarantees for $\ell_1$ recovery of block-sparse signals. *Ann. Stat.*, 40(6):3077–3107, 2012.

J. Jurečková, P. K. Sen, and J. Picek. *Methodology in Robust and Nonparametric Statistics*. CRC Press, 2012.

R. Koenker. *Quantile regression*, volume 38 of *Econometric Society Monographs*. Cambridge University Press, Cambridge, 2005.

R. Koenker and G. Bassett, Jr. Regression quantiles. *Econometrica*, 46(1):33–50, 1978.

R. Koenker and Z. Xiao. Inference on the quantile regression process. *Econometrica*, 70(4):1583–1612, 2002.

Y. Li and J. Zhu. $L_1$-norm quantile regression. *J. Comp. Graph. Stat.*, 17(1):163–185, 2008.

H. Liu and B. Yu. Asymptotic properties of Lasso+mLS and Lasso+Ridge in sparse high-dimensional linear regression. *Electron. J. Stat.*, 7:3124–3169, 2013.

M. Lopes. A residual bootstrap for high-dimensional regression with near low-rank designs. In Z. Ghahramani, M. Welling, C. Cortes, N. Lawrence, and K. Weinberger, editors, *Advances in Neural Information Processing Systems 27*, pages 3239–3247. Curran Associates, Inc., 2014.

L. Mackey, M. I. Jordan, R. Y. Chen, B. Farrell, and J. A. Tropp. Matrix concentration inequalities via the method of exchangeable pairs. *Annals of Probability*, 42(3):906–945, 2014.

Z. X. Pin Ng, Wing-Keung Wong. Stochastic dominance via quantile regression. *Working Paper Series–11-01*, 2011.

Z. Qu. Testing for structural change in regression quantiles. *Journal of Econometrics*, 146(1):170 – 184, 2008.

Z. Qu and J. Yoon. Nonparametric estimation and inference on conditional quantile processes. *Journal of Econometrics*, 185(1):1 – 19, 2015.

M. Rudelson and R. Vershynin. Hanson-wright inequality and sub-gaussian concentration. *Electron. Commun. Probab.*, 18(0), 2013.

S. A. van de Geer. *Applications of empirical process theory*, volume 6 of *Cambridge Series in Statistical and Probabilistic Mathematics*. Cambridge University Press, Cambridge, 2000.

S. A. van de Geer, P. Bühlmann, Y. Ritov, and R. Dezeure. On asymptotically optimal confidence regions and tests for high-dimensional models. *Ann. Stat.*, 42(3):1166–1202, 2014.

A. W. van der Vaart and J. A. Wellner. *Weak Convergence and Empirical Processes: With Applications to Statistics*. Springer, 1996.

M. J. Wainwright. Sharp thresholds for high-dimensional and noisy sparsity recovery using $\ell_1$-constrained quadratic programming (lasso). *IEEE Trans. Inf. Theory*, 55(5):2183–2202, 2009.

L. Wang, Y. Wu, and R. Li. Quantile regression for analyzing heterogeneity in ultra-high dimension. *J. Am. Stat. Assoc.*, 107(497):214–222, 2012.

L. Wang, Y. Kim, and R. Li. Calibrating nonconvex penalized regression in ultra-high dimension. *Ann. Statist.*, 41(5):2505–2536, 2013.

Y. Wu and Y. Liu. Variable selection in quantile regression. *Stat. Sinica*, 19(2):801–817, 2009.

C.-H. Zhang. Nearly unbiased variable selection under minimax concave penalty. *Ann. Stat.*, 38(2):894–942, 2010.

C.-H. Zhang and S. S. Zhang. Confidence intervals for low dimensional parameters in high dimensional linear models. *J. R. Stat. Soc. B*, 76(1):217–242, 2013.

P. Zhao and B. Yu. On model selection consistency of lasso. *J. Mach. Learn. Res.*, 7:2541–2563, 2006.

T. Zhao, M. Kolar, and H. Liu. A General Framework for Robust Testing and Confidence Regions in High-Dimensional Quantile Regression. *ArXiv e-prints*, 2014.

Q. Zheng, C. Gallagher, and K. B. Kulasekera. Adaptive penalized quantile regression for high dimensional data. *J. Statist. Plann. Inference*, 143(6):1029–1038, 2013.

K. Q. Zhou and S. L. Portnoy. Direct use of regression quantiles to construct confidence sets in

linear models. *Ann. Stat.*, 24(1):287–306, 1996.

H. Zou and M. Yuan. Composite quantile regression and the oracle model selection theory. *Ann. Stat.*, 36(3):1108–1126, 2008.

# Supplementary Material

In this document we provide detailed proofs of all of the results presented in the main text. The proofs are organized by sections with the proof of the main theorem of the section first, followed by the proofs and statements of necessary lemmas.

## 8 Technical Proofs

### 8.1 Preparatory Lemmas

For the proof we will need the following Hanson-Wright inequality that can be found in Rudelson and Vershynin (2013).

**Lemma 11** (Preparatory lemma: Hanson-Wright inequality). *Let $\mathbf{X} = (X_1, \ldots, X_n) \in \mathbb{R}^n$ be a random vector with independent component $X_i$ which satisfy $\mathbb{E}X_i = 0$ and $\|X_i\|_{\Psi_2} \leq K$. Let $\mathbf{A}$ be an $n \times n$ matrix. Then, for every $t \geq 0$*

$$\mathbb{P}\left(\left|\mathbf{X}^\top \mathbf{A} \mathbf{X} - \mathbb{E}\mathbf{X}^\top \mathbf{A} \mathbf{X}\right| > t\right) \leq 2\exp\left\{-c\min\left(\frac{t^2}{K^4\|\mathbf{A}\|_F^2}, \frac{t}{K^2\|\mathbf{A}\|^2}\right)\right\}.$$

### 8.2 Proofs for Section 2

*Proof of Proposition 2.1.* Using Knight's identity (2) and the representation of the loss function (2.5), we have that

$$\widehat{\boldsymbol{\beta}}(\tau) = \arg\min_{\boldsymbol{\beta} \in \mathbb{R}^{p+1}} \Bigg\{ \frac{1}{n} \sum_{i \in [n]} \bigg[ -\mathbf{X}_i^T (\boldsymbol{\beta} - \boldsymbol{\beta}^*(\tau)) \psi_\tau \left(u_i - F^{-1}(\tau)\right) $$
$$+ \int_0^{\mathbf{X}_i^T(\boldsymbol{\beta} - \boldsymbol{\beta}^*(\tau))} \left[\mathbb{I}\left\{u_i \leq s + F^{-1}(\tau)\right\} - \mathbb{I}\{u_i \leq F^{-1}(\tau)\}\right] ds \bigg]$$
$$+ \sum_{j \in [p]} \lambda_j |\beta_j| \Bigg\}.$$

To simplify the notation, let $\boldsymbol{\delta} = \widehat{\boldsymbol{\beta}}(\tau) - \boldsymbol{\beta}^*(\tau)$. Since the function $\psi_\tau$ can be represented as

$$\psi_\tau(Y_i - \mathbf{X}_i^T \widehat{\boldsymbol{\beta}}(\tau)) = \psi_\tau(u_i - F^{-1}(\tau)) - \mathbb{I}\left\{u_i \leq \mathbf{X}_i^T \boldsymbol{\delta} + F^{-1}(\tau)\right\} + \mathbb{I}\left\{u_i \leq F^{-1}(\tau)\right\},$$

from (2.2) we obtain the following representation

$$\check{\boldsymbol{\beta}}(\tau) - \boldsymbol{\beta}^*(\tau) = n^{-1}\widehat{\mathbf{D}}_\tau \sum_{i \in [n]} \mathbf{X}_i \psi_\tau \left(u_i - F^{-1}(\tau)\right) - \frac{\widehat{\varsigma}(\tau)}{\varsigma(\tau)} G_n(\tau, \boldsymbol{\delta}) - \frac{\widehat{\varsigma}(\tau)}{\varsigma(\tau)} \mathbb{E}_{u_i}\left[\nu_n(\tau, \boldsymbol{\delta})\right] + \boldsymbol{\delta}. \quad (8.1)$$

Note that

$$\mathbb{E}_{u_i}\left[\nu_n(\tau, \boldsymbol{\delta})\right] = n^{-1}\sum_{i \in [n]} \widetilde{\mathbf{D}}_\tau \mathbf{X}_i \left(\mathbb{P}\left[u_i \leq \mathbf{X}_i^T \boldsymbol{\delta} + F^{-1}(\tau)\right] - \mathbb{P}\left[u_i \leq F^{-1}(\tau)\right]\right)$$
$$= n^{-1}\sum_{i \in [n]} \widehat{\mathbf{D}}\mathbf{X}_i \mathbf{X}_i^T \boldsymbol{\delta} + n^{-1}\sum_{i \in [n]} f(\bar{w}_i) \widetilde{\mathbf{D}}_\tau \mathbf{X}_i \left(\mathbf{X}_i^T \boldsymbol{\delta}\right)^2,$$

using the mean value theorem and $\bar{w}_i$ is between $\mathbf{X}_i^T \boldsymbol{\delta} + F^{-1}(\tau)$ and $F^{-1}(\tau)$. The desired representation immediately follows by plugging the above display into (8.1) and rearranging terms. □

## 8.3 Proofs for Section 3

*Proof of Theorem 3.1.* Starting from Proposition 2.1, we have uniformly in $\tau$ that

$$\sqrt{n}(\bar{\boldsymbol{\beta}}(\tau) - \boldsymbol{\beta}^*(\tau)) = n^{-1/2}\widetilde{\mathbf{D}}_\tau \sum_{i\in[n]} \mathbf{X}_i \psi_\tau\left(u_i - F^{-1}(\tau)\right) - \sqrt{n}\left(\boldsymbol{\Delta}_1(\tau) + \boldsymbol{\Delta}_2(\tau) + \boldsymbol{\Delta}_3(\tau)\right), \quad (8.2)$$

where

$$\boldsymbol{\Delta}_1(\tau) = G_n(\tau, \boldsymbol{\delta}), \qquad \boldsymbol{\Delta}_2(\tau) = \left(\widehat{\mathbf{D}}\widehat{\boldsymbol{\Sigma}} - \mathbf{I}\right)\left(\widehat{\boldsymbol{\beta}}(\tau) - \boldsymbol{\beta}^*(\tau)\right),$$

and

$$\boldsymbol{\Delta}_3(\tau) = n^{-1} \sum_{i\in[n]} \frac{f(\bar{w}_i)}{f(F^{-1}(\tau))} \widehat{\mathbf{D}} \mathbf{X}_i \left(\mathbf{X}_i^T \boldsymbol{\delta}\right)^2$$

with $\boldsymbol{\delta} = \widehat{\boldsymbol{\beta}}(\tau) - \boldsymbol{\beta}^*(\tau)$ and $\bar{w}_i$ between $\mathbf{X}_i^T \boldsymbol{\delta} + F^{-1}(\tau)$ and $F^{-1}(\tau)$.

Using Lemma 1 with $r_n = C\sqrt{s\log(p\vee n)/n}$ and $t = Cs$, we have that

$$\sup_{\tau\in[\epsilon,1-\epsilon]} \sqrt{n}\boldsymbol{\Delta}_1(\tau) = \mathcal{O}_P\left(\frac{L_n s^{3/4} \log^{3/4}(p\vee n)}{n^{1/4}} \bigvee \frac{L_n s \log(p\vee n)}{n^{1/2}}\right).$$

Lemma 2 gives us

$$\sup_{\tau\in[\epsilon,1-\epsilon]} \sqrt{n}\boldsymbol{\Delta}_2(\tau) \leq \sqrt{n}\gamma_n \sup_{\tau\in[\epsilon,1-\epsilon]} ||\widehat{\boldsymbol{\beta}}(\tau) - \boldsymbol{\beta}(\tau)||_1 = \mathcal{O}_P\left(\frac{s\log(p\vee n)}{n^{1/2}}\right).$$

Finally, Lemma 3 gives us

$$\sup_{\tau\in[\epsilon,1-\epsilon]} \sqrt{n}\boldsymbol{\Delta}_3(\tau) = \mathcal{O}_P\left(\frac{L_n s \log(p\vee n)}{n^{1/2}}\right).$$

Combining all the results with (8.2), we have that

$$\sqrt{n}(\bar{\boldsymbol{\beta}}(\tau) - \boldsymbol{\beta}^*(\tau)) = n^{-1/2} \mathbf{D}_\tau \sum_{i\in[n]} \mathbf{X}_i \psi_\tau\left(u_i - F^{-1}(\tau)\right)$$

$$+ \mathcal{O}_P\left(\frac{L_n s^{3/4} \log^{3/4}(p\vee n)}{n^{1/4}} \bigvee \frac{L_n s \log(p\vee n)}{n^{1/2}}\right)$$

uniformly in $\tau$, which completes the proof. □

*Proof Lemma 1.* Recall that

$$Y_i = \mathbf{X}_i^T \boldsymbol{\beta}^*(\tau) - F^{-1}(\tau) + u_i$$

where $\boldsymbol{\beta}^*(\tau) = (\beta_0^* + F^{-1}(\tau), \beta_1^*, \ldots, \beta_p^*)^T \in \mathbb{R}^{p+1}$. To facilitate the proof, we introduce some additional notation. Denote

$$\widehat{P}_i(\tau, \boldsymbol{\delta}) = \mathbb{I}\left\{u_i \leq F^{-1}(\tau) + \mathbf{X}_i^T \boldsymbol{\delta}\right\}$$

and

$$P_i(\tau, \boldsymbol{\delta}) = \mathbb{P}[u_i \leq F^{-1}(\tau) + \mathbf{X}_i^T \boldsymbol{\delta} \mid \mathbf{X}_i].$$

Then

$$\nu_n(\tau, \boldsymbol{\delta}) = n^{-1} \sum_{i \in [n]} \widetilde{\mathbf{D}}_\tau \mathbf{X}_i \left(\widehat{P}_i(\tau, \boldsymbol{\delta}) - \widehat{P}_i(\tau, \mathbf{0})\right).$$

Let $\{\widetilde{\tau}_l\}_{l \in [N_\tau]}$ with $\widetilde{\tau}_l = \epsilon + \frac{1-2\epsilon}{2N_\tau}(2l-1)$, where $N_\tau \in \mathbb{N}$ will be chosen later. Furthermore, let $\{\widetilde{\boldsymbol{\delta}}_k\}_{k \in [N_\delta]}$ be centers of the balls of radius $r_n \xi_n$ that cover the set $\mathcal{C}(r_n, t)$. Such a cover can be constructed with $N_\delta \leq \binom{p}{t}(3/\xi_n)^t$ (see, for example, Lemma 2.5 in van de Geer, 2000). Furthermore, let

$$\mathcal{B}^{\text{supp}}(\widetilde{\boldsymbol{\delta}}_k, r) = \left\{\boldsymbol{\delta} \mid \|\widetilde{\boldsymbol{\delta}}_k - \boldsymbol{\delta}\|_2 \leq r \wedge \text{supp}(\boldsymbol{\delta}) \subseteq \text{supp}(\widetilde{\boldsymbol{\delta}}_k)\right\}$$

be a ball or radius $r$ centered at $\widetilde{\boldsymbol{\delta}}_k$ with points that have the same support as $\widetilde{\boldsymbol{\delta}}_k$. In what follows, we will bound $\sup_{\tau \in [\epsilon, 1-\epsilon]} \sup_{\boldsymbol{\delta} \in \mathcal{C}(r_n, t)} \|G_n(\tau, \boldsymbol{\delta})\|_\infty$ using a standard $\epsilon$-net argument. In particular, using the above introduced notation, we have the following decomposition

$$\begin{aligned}
&\sup_{\tau \in [\epsilon, 1-\epsilon]} \sup_{\boldsymbol{\delta} \in \mathcal{C}(r_n, t)} \|G_n(\tau, \boldsymbol{\delta})\|_\infty \\
&= \max_{l \in [N_\tau]} \max_{k \in [N_\delta]} \sup_{\tau \in \left[\widetilde{\tau}_l - \frac{1-2\epsilon}{2N_\tau}, \widetilde{\tau}_l + \frac{1-2\epsilon}{2N_\tau}\right]} \sup_{\boldsymbol{\delta} \in \mathcal{B}^{\text{supp}}(\widetilde{\boldsymbol{\delta}}_k, r_n \xi_n)} \|G_n(\tau, \boldsymbol{\delta})\|_\infty \\
&\leq \underbrace{\max_{l \in [N_\tau]} \max_{k \in [N_\delta]} \|G_n(\widetilde{\tau}_l, \widetilde{\boldsymbol{\delta}}_k)\|_\infty}_{T_1} \\
&\quad + \underbrace{\max_{l \in [N_\tau]} \max_{k \in [N_\delta]} \sup_{\tau \in \left[\widetilde{\tau}_l - \frac{1-2\epsilon}{2N_\tau}, \widetilde{\tau}_l + \frac{1-2\epsilon}{2N_\tau}\right]} \sup_{\boldsymbol{\delta} \in \mathcal{B}^{\text{supp}}(\widetilde{\boldsymbol{\delta}}_k, r_n \xi_n)} \|G_n(\tau, \boldsymbol{\delta}) - G_n(\widetilde{\tau}_l, \widetilde{\boldsymbol{\delta}}_k)\|_\infty}_{T_2}.
\end{aligned} \qquad (8.3)$$

Notice that the term $T_1$ arises from discretization of the sets $[\epsilon, 1-\epsilon]$ and $\mathcal{C}(r_n, t)$. To control it, we will apply the union bound for each fixed $l$ and $k$. The term $T_2$ captures the deviation of the process in a small neighborhood around the fixed center $\widetilde{\tau}_l$ and $\widetilde{\boldsymbol{\delta}}_k$. In the remainder of the proof, we provide details for bounding $T_1$ and $T_2$.

We first bound the term $T_1$ in (8.3). Let $a_{ij}(\tau) = \mathbf{e}_j^T \widetilde{\mathbf{D}}_\tau \mathbf{X}_i$,

$$Z_{ijlk} = a_{ij}(\widetilde{\tau}_l) \left( \left( \widehat{P}_i(\widetilde{\tau}_l, \widetilde{\boldsymbol{\delta}}_k) - P_i(\widetilde{\tau}_l, \widetilde{\boldsymbol{\delta}}_k) \right) - \left( \widehat{P}_i(\widetilde{\tau}_l, \mathbf{0}) - P_i(\widetilde{\tau}_l, \mathbf{0}) \right) \right)$$

and

$$\widetilde{Z}_{ijlk} = a_{ij}(\widetilde{\tau}_l) \left( P_i\left(\widetilde{\tau}_l, \widetilde{\boldsymbol{\delta}}_k\right) - P_i\left(\widetilde{\tau}_l, \mathbf{0}\right) \right) - \mathbb{E}\left[ a_{ij}(\widetilde{\tau}_l) \left( \widehat{P}_i(\widetilde{\tau}_l, \widetilde{\boldsymbol{\delta}}_k) - \widehat{P}_i(\widetilde{\tau}_l, \mathbf{0}) \right) \right]$$

Then

$$T_1 = \max_{l \in [N_\tau]} \max_{k \in [N_\delta]} \max_{j \in [p]} \left| n^{-1} \sum_{i \in [n]} \left( Z_{ijlk} + \widetilde{Z}_{ijlk} \right) \right|$$

$$\leq \underbrace{\max_{l \in [N_\tau]} \max_{k \in [N_\delta]} \max_{j \in [p]} \left| n^{-1} \sum_{i \in [n]} Z_{ijlk} \right|}_{T_{11}} + \underbrace{\max_{l \in [N_\tau]} \max_{k \in [N_\delta]} \max_{j \in [p]} \left| n^{-1} \sum_{i \in [n]} \widetilde{Z}_{ijlk} \right|}_{T_{12}}.$$

Note that $\mathbb{E}[Z_{ijlk} \mid \{\mathbf{X}_i\}_{i \in [n]}] = 0$ and

$$\begin{aligned}
\text{Var}&[Z_{ijlk} \mid \{\mathbf{X}_i\}_{i \in [n]}] \\
&= a_{ij}^2(\widetilde{\tau}_l) \Big( P_i(\widetilde{\tau}_l, \widetilde{\boldsymbol{\delta}}_k) - P_i^2(\widetilde{\tau}_l, \widetilde{\boldsymbol{\delta}}_k) + P_i(\widetilde{\tau}_l, \mathbf{0}) - P_i^2(\widetilde{\tau}_l, \mathbf{0}) \\
&\qquad\qquad - 2\left( P_i(\widetilde{\tau}_l, \mathbf{0}) \vee P_i(\widetilde{\tau}_l, \widetilde{\boldsymbol{\delta}}_k) \right) + 2 P_i\left(\widetilde{\tau}_l, \widetilde{\boldsymbol{\delta}}_k\right) P_i\left(\widetilde{\tau}_l, \mathbf{0}\right) \Big) \\
&\overset{(i)}{\leq} a_{ij}^2(\widetilde{\tau}_l) \left( P_i(\widetilde{\tau}_l, \widetilde{\boldsymbol{\delta}}_k) + P_i(\widetilde{\tau}_l, \mathbf{0}) - 2\left( P_i(\widetilde{\tau}_l, \mathbf{0}) \vee P_i(\widetilde{\tau}_l, \widetilde{\boldsymbol{\delta}}_k) \right) \right) \\
&\overset{(ii)}{\leq} a_{ij}^2(\widetilde{\tau}_l) \left| \mathbf{X}_i^T \widetilde{\boldsymbol{\delta}}_k \right| f_i\left( F^{-1}(\widetilde{\tau}_l) + \eta_i \mathbf{X}_i^T \widetilde{\boldsymbol{\delta}}_k \right) \quad (\eta_i \in [0,1]) \\
&\overset{(iii)}{\leq} a_{ij}^2(\widetilde{\tau}_l) \left| \mathbf{X}_i^T \widetilde{\boldsymbol{\delta}}_k \right| f_{\max}
\end{aligned}$$

where $(i)$ follows by dropping a negative term, $(ii)$ follows by the mean value theorem, and $(iii)$ from the assumption that the conditional density is bounded stated in Condition **(D)**.

Furthermore, conditional on $\{\mathbf{X}_i\}_{i \in [n]}$ we have that $|Z_{ijlk}| \leq 4 \max_{ij} |a_{ij}(\widetilde{\tau}_l)|$ a.s. For a fixed $j$, $k$, and $l$, Bernstein's inequality (see, for example, Section 2.2.2 of van der Vaart and Wellner, 1996) gives us

$$\left| n^{-1} \sum_{i \in [n]} Z_{ijlk} \right| \leq C \left( \sqrt{\frac{f_{\max} \log(2/\delta)}{n^2} \sum_{i \in [n]} a_{ij}^2(\widetilde{\tau}_l) \left| \mathbf{X}_i^T \widetilde{\boldsymbol{\delta}}_k \right|} \bigvee \frac{\max_{i \in [n], j \in [p]} |a_{ij}(\widetilde{\tau}_l)|}{n} \log(2/\delta) \right)$$

with probability $1 - \delta$. By construction in (2.4), we have

$$\max_{i \in [n], j \in [p]} |a_{ij}(\widetilde{\tau}_l)| \leq f_{\min}^{-1} L_n$$

and therefore

$$\sum_{i\in[n]} a_{ij}^2(\widetilde{\tau}_l)\left|\mathbf{X}_i^T\widetilde{\boldsymbol{\delta}}_k\right| \leq f_{\min}^{-2}L_n^2 n\sqrt{\widetilde{\boldsymbol{\delta}}_k^T\widehat{\boldsymbol{\Sigma}}\widetilde{\boldsymbol{\delta}}_k}$$

$$\leq (1+o_P(1))f_{\min}^{-2}L_n^2 n r_n \Lambda_{\max}^{1/2}(\boldsymbol{\Sigma}),$$

where the line follows using the Cauchy-Schwartz inequality and inequality (58a) of Wainwright (2009). Combining the results, with probability $1-2\delta$ we have that

$$\left|n^{-1}\sum_{i\in[n]} Z_{ijlk}\right| \leq C\left(\sqrt{\frac{f_{\max}\Lambda_{\max}^{1/2}(\boldsymbol{\Sigma})f_{\min}^{-2}L_n^2 r_n \log(2/\delta)}{n}} \vee \frac{f_{\min}^{-1}L_n \log(2/\delta)}{n}\right).$$

Using the union bound over $j \in [p]$, $l \in [N_\tau]$ and $k \in [N_\delta]$, with probability $1-2\delta$, we have

$$T_{11} \leq C\left(\sqrt{\frac{f_{\max}\Lambda_{\max}^{1/2}(\boldsymbol{\Sigma})f_{\min}^{-2}L_n^2 r_n \log(2N_\tau N_\delta p/\delta)}{n}} \vee \frac{f_{\min}^{-1}L_n \log(2N_\tau N_\delta p/\delta)}{n}\right).$$

We deal with the term $T_{12}$ in a similar way. We will work on the event

$$\mathcal{A} = \left\{\sup_{\tau\in[\epsilon,1-\epsilon]}\max_{i\in[n],j\in[p]}|a_{ij}(\tau)| \leq f_{\min}^{-1}L_n\right\}, \qquad (8.4)$$

which holds with probability at $1-\delta$ using Lemma 2. For a fixed $l$, $k$ and $j$, we apply Bernstein's inequality to obtain

$$\left|n^{-1}\sum_{i\in[n]} \widetilde{Z}_{ijlk}\right| \leq C\left(\sqrt{\frac{f_{\min}^{-2}L_n^2 f_{\max}^2 \Lambda_{\max}(\boldsymbol{\Sigma})r_n^2 \log(2/\delta)}{n}} \vee \frac{f_{\min}^{-1}L_n \log(2/\delta)}{n}\right)$$

with probability $1-\delta$, since on the event $\mathcal{A}$ in (8.4) we have that $\left|\widetilde{Z}_{ijlk}\right| \leq 4f_{\min}^{-1}L_n$ and

$$\mathrm{Var}\left[\widetilde{Z}_{ijlk}\right] \leq \mathbb{E}\left[a_{ij}^2(\widetilde{\tau}_l)\left(P_i(\widetilde{\tau}_l,\widetilde{\boldsymbol{\delta}}_k) - P_i(\widetilde{\tau}_l,\mathbf{0})\right)^2\right]$$

$$\leq f_{\min}^{-2}L_n^2 f_{\max}^2 r_n^2 \Lambda_{\max}(\boldsymbol{\Sigma}).$$

The union bound over $l \in [N_\tau]$, $k \in [N_\delta]$, and $j \in [p]$, gives us

$$T_{12} \leq C\left(\sqrt{\frac{f_{\min}^{-2}L_n^2 f_{\max}^2 \Lambda_{\max}(\boldsymbol{\Sigma})r_n^2 \log(2N_\tau N_\delta p/\delta)}{n}} \vee \frac{f_{\min}^{-1}L_n \log(2N_\tau N_\delta p/\delta)}{n}\right)$$

with probability at least $1-2\delta$. Combining the bounds on $T_{11}$ and $T_{12}$, with probability $1-4\delta$, we have

$$T_1 \leq C\left(\sqrt{\frac{f_{\min}^{-2}L_n^2 f_{\max}\Lambda_{\max}^{1/2}(\boldsymbol{\Sigma})r_n \log(2N_\tau N_\delta p/\delta)}{n}} \vee \frac{f_{\min}^{-1}L_n \log(2N_\tau N_\delta p/\delta)}{n}\right),$$

since $f_{\max}\Lambda_{\max}^{1/2}(\Sigma)r_n = \mathcal{O}_P(1)$.

Let us now focus on bounding $T_2$ term. Note that $a_{ij}(\tau) = a_{ij}(\widetilde{\tau}_l) + a'_{ij}(\bar{\tau}_l)(\tau - \widetilde{\tau}_l)$ for some $\bar{\tau}_l$ between $\tau$ and $\tau_l$. Let

$$W_{ijl}(\tau, \boldsymbol{\delta}) = a'_{ij}(\bar{\tau}_l)(\tau - \widetilde{\tau}_l)\left(\widehat{P}_i(\tau, \boldsymbol{\delta}) - \widehat{P}_i(\tau, \mathbf{0})\right).$$

For a fixed $j, l$, and $k$ we have

$$\sup_{\tau \in \left[\widetilde{\tau}_l - \frac{1-2\epsilon}{2N_\tau}, \widetilde{\tau}_l + \frac{1-2\epsilon}{2N_\tau}\right]} \sup_{\boldsymbol{\delta} \in \mathcal{B}^{\text{supp}}(\widetilde{\boldsymbol{\delta}}_k, r_n\xi_n)} \left|\mathbf{e}_j^T \left(G_n(\tau, \boldsymbol{\delta}) - G_n(\widetilde{\tau}_l, \widetilde{\boldsymbol{\delta}}_k)\right)\right|$$

$$\leq \underbrace{\sup_{\tau \in \left[\widetilde{\tau}_l - \frac{1-2\epsilon}{2N_\tau}, \widetilde{\tau}_l + \frac{1-2\epsilon}{2N_\tau}\right]} \sup_{\boldsymbol{\delta} \in \mathcal{B}^{\text{supp}}(\widetilde{\boldsymbol{\delta}}_k, r_n\xi_n)} \left|\mathbf{e}_j^T \left(G_n(\widetilde{\tau}_l, \boldsymbol{\delta}) - G_n(\widetilde{\tau}_l, \widetilde{\boldsymbol{\delta}}_k)\right)\right|}_{T_{21}}$$

$$+ \underbrace{\sup_{\tau \in \left[\widetilde{\tau}_l - \frac{1-2\epsilon}{2N_\tau}, \widetilde{\tau}_l + \frac{1-2\epsilon}{2N_\tau}\right]} \sup_{\boldsymbol{\delta} \in \mathcal{B}^{\text{supp}}(\widetilde{\boldsymbol{\delta}}_k, r_n\xi_n)} \left|n^{-1}\sum_{i \in [n]} W_{ij}(\tau, \boldsymbol{\delta}) - \mathbb{E}\left[W_{ijl}(\tau, \boldsymbol{\delta})\right]\right|}_{T_{22}}.$$

We will deal with the two terms separately. For $T_{21}$, we will use the fact that $\mathbb{I}\{a < x\}$ and $\mathbb{P}\{Z < x\}$ are monotone function in $x$. Note that

$$\max_{l \in [N_\tau]} \max_{k \in [N_\delta]} \max_{i \in [n]} \sup_{\tau \in \left[\widetilde{\tau}_l - \frac{1-2\epsilon}{2N_\tau}, \widetilde{\tau}_l + \frac{1-2\epsilon}{2N_\tau}\right]} \sup_{\boldsymbol{\delta} \in \mathcal{B}^{\text{supp}}(\widetilde{\boldsymbol{\delta}}_k, r_n\xi_n)} \left|F^{-1}(\tau) + \mathbf{X}_i^T\boldsymbol{\delta} - F^{-1}(\widetilde{\tau}_l) + \mathbf{X}_i^T\widetilde{\boldsymbol{\delta}}_k\right|$$

$$\leq \left(\frac{f_{\min}^{-1}(1-2\epsilon)}{2N_\tau} + r_n\xi_n\sqrt{t}\max_{i,j}|x_{ij}|\right)$$

$$\leq C\left(\frac{f_{\min}^{-1}}{N_\tau} + r_n\xi_n\sqrt{\left(\max_j \Sigma_{jj}\right)t\log(2np\delta^{-1})}\right)$$

$$=: \widetilde{L}_n,$$

with probability $1 - \delta$, since $|F^{-1}(\tau) - F^{-1}(\widetilde{\tau}_l)| \leq f_{\min}^{-1}|\tau - \widetilde{\tau}_l|$ using the mean value theorem,

$$\left|\mathbf{X}_i^T(\boldsymbol{\delta} - \widetilde{\boldsymbol{\delta}}_k)\right| \leq \|\boldsymbol{\delta} - \widetilde{\boldsymbol{\delta}}_k\|_2\sqrt{|\text{supp}(\boldsymbol{\delta} - \widetilde{\boldsymbol{\delta}}_k)|}\max_{i,j}|x_{ij}|$$

and finally $\max_{i,j}|x_{ij}| \leq C\sqrt{(\max_j \Sigma_{jj})\log(2np\delta^{-1})}$ using a tail bound for subgaussian random variables (see Section 2.2.1. of van der Vaart and Wellner, 1996). Now,

$$T_{21} \leq n^{-1} \sum_{i \in [n]} \left[ |a_{ij}(\widetilde{\tau}_l)| \left( \mathbb{1}\left\{u_i \leq F^{-1}(\widetilde{\tau}_l) + \mathbf{X}_i^T \widetilde{\boldsymbol{\delta}}_k + \widetilde{L}_n\right\} - \mathbb{1}\left\{u_i \leq F^{-1}(\widetilde{\tau}_l) - \widetilde{L}_n\right\} \right.\right.$$
$$- \mathbb{1}\left\{u_i \leq F^{-1}(\widetilde{\tau}_l) + \mathbf{X}_i^T \widetilde{\boldsymbol{\delta}}_k\right\} + \mathbb{1}\left\{u_i \leq F^{-1}(\widetilde{\tau}_l)\right\}$$
$$- \mathbb{P}\left[u_i \leq F^{-1}(\widetilde{\tau}_l) + \mathbf{X}_i^T \widetilde{\boldsymbol{\delta}}_k - \widetilde{L}_n\right] + \mathbb{P}\left[u_i \leq F^{-1}(\widetilde{\tau}_l) + \widetilde{L}_n\right]$$
$$\left.\left.+ \mathbb{P}\left[u_i \leq F^{-1}(\widetilde{\tau}_l) + \mathbf{X}_i^T \widetilde{\boldsymbol{\delta}}_k\right] - \mathbb{P}\left[u_i \leq F^{-1}(\widetilde{\tau}_l)\right] \right) \right]$$

$$\leq n^{-1} \sum_{i \in [n]} \left[ |a_{ij}(\widetilde{\tau}_l)| \left( \mathbb{1}\left\{u_i \leq F^{-1}(\widetilde{\tau}_l) + \mathbf{X}_i^T \widetilde{\boldsymbol{\delta}}_k + \widetilde{L}_n\right\} - \mathbb{1}\left\{u_i \leq F^{-1}(\widetilde{\tau}_l) - \widetilde{L}_n\right\} \right.\right.$$
$$- \mathbb{1}\left\{u_i \leq F^{-1}(\widetilde{\tau}_l) + \mathbf{X}_i^T \widetilde{\boldsymbol{\delta}}_k\right\} + \mathbb{1}\left\{u_i \leq F^{-1}(\widetilde{\tau}_l)\right\}$$
$$- \mathbb{P}\left[u_i \leq F^{-1}(\widetilde{\tau}_l) + \mathbf{X}_i^T \widetilde{\boldsymbol{\delta}}_k + \widetilde{L}_n\right] + \mathbb{P}\left[u_i \leq F^{-1}(\widetilde{\tau}_l) - \widetilde{L}_n\right]$$
$$\left.\left.+ \mathbb{P}\left[u_i \leq F^{-1}(\widetilde{\tau}_l) + \mathbf{X}_i^T \widetilde{\boldsymbol{\delta}}_k\right] - \mathbb{P}\left[u_i \leq F^{-1}(\widetilde{\tau}_l)\right] \right) \right]$$

$$+ n^{-1} \sum_{i \in [n]} \left[ |a_{ij}(\widetilde{\tau}_l)| \left( \mathbb{P}\left[u_i \leq F^{-1}(\widetilde{\tau}_l) + \mathbf{X}_i^T \widetilde{\boldsymbol{\delta}}_k + \widetilde{L}_n\right] - \mathbb{P}\left[u_i \leq F^{-1}(\widetilde{\tau}_l) - \widetilde{L}_n\right]\right.\right.$$
$$\left.\left.- \mathbb{P}\left[u_i \leq F^{-1}(\widetilde{\tau}_l) + \mathbf{X}_i^T \widetilde{\boldsymbol{\delta}}_k - \widetilde{L}_n\right] + \mathbb{P}\left[u_i \leq F^{-1}(\widetilde{\tau}_l) + \widetilde{L}_n\right] \right) \right].$$

The first term in the display above can be bounded in a similar way to $T_1$ by applying Bernstein's inequality and hence the details are omitted. For the second term we have a bound $Cf_{\min}^{-1}f_{\max}L_n\widetilde{L}_n$, since $|a_{ij}(\widetilde{\tau}_l)| \leq f_{\min}^{-1}L_n$ by construction and $F(x + \widetilde{L}_n) - F(x - \widetilde{L}_n) \leq Cf_{\max}\widetilde{L}_n$ for all $x \in \mathbb{R}$. Therefore, with probability $1 - 2\delta$,

$$T_{21} \leq C \left( \sqrt{\frac{f_{\max}f_{\min}^{-2}L_n^2 \widetilde{L}_n \log(2/\delta)}{n}} \bigvee \frac{f_{\min}^{-1}L_n \log(2/\delta)}{n} \bigvee f_{\min}^{-1}f_{\max}L_n\widetilde{L}_n \right).$$

A bound on $T_{22}$ is obtain similarly to that on $T_{21}$. The only difference is that we need to bound $|a'_{ij}(\bar{\tau})(\tau - \widetilde{\tau}_l)|$ for $\tau \in \left[\widetilde{\tau}_l - \frac{1-2\epsilon}{2N_\tau}, \widetilde{\tau}_l + \frac{1-2\epsilon}{2N_\tau}\right]$ instead of $|a_{ij}(\widetilde{\tau}_l)|$. We have $|a'_{ij}(\bar{\tau})(\tau - \widetilde{\tau}_l)| \leq CN_\tau^{-1}f'_{\max}L_n$, since $|X_i^T d| \leq L_n$ by construction and $|\tau - \widetilde{\tau}| \leq CN_\tau^{-1}$. Now,

$$T_{22} \leq C\left( \sqrt{\frac{f_{\max}(f'_{\min})^{-2}L_n^2 N_\tau^{-2} \left(r_n \Lambda_{\max}^{1/2}(\boldsymbol{\Sigma}) + \widetilde{L}_n\right) \log(2/\delta)}{n}} \right.$$
$$\left. \bigvee \frac{N_\tau^{-1}(f'_{\min})^{-1} L_n \log(2/\delta)}{n} \bigvee N_\tau^{-1}(f'_{\min})^{-1} f_{\max} L_n \widetilde{L}_n \right).$$

A bound on $T_2$ now follows using a union bound over $j \in [p]$, $l \in [N_\tau]$ and $k \in [N_\delta]$.

We can choose $N_\tau = n^2$ and $\xi_n = n^{-1}$, which gives us $N_\delta \lesssim (pn^2)^t$. With these choices, the term $T_2$ is negligible compared to $T_1$ and we obtain

$$T \leq C \left( \sqrt{\frac{f_{\min}^{-2} L_n^2 f_{\max} \Lambda_{\max}^{1/2}(\mathbf{\Sigma}) r_n t \log(np/\delta)}{n}} \vee \frac{f_{\min}^{-1} L_n t \log(np/\delta)}{n} \right),$$

which completes the proof. $\square$

*Proof of Lemma 2.* The lemma is proven by showing that $\mathbf{\Sigma}^{-1}$ is a feasible point for the optimization problem in (2.4). From Lemma 6.2 of Javanmard and Montanari (2014), we have that

$$||\mathbf{\Sigma}^{-1} \widehat{\mathbf{\Sigma}} - \mathbf{I}||_{\max} \leq \gamma_n$$

with probability at least $1 - C'p^{-c'}$. Using a standard tail bound for the sub-gaussian random variables (see Section 2.2.1. of van der Vaart and Wellner, 1996), we have that

$$\max_{i \in [n]} ||\mathbf{\Sigma}^{-1} \mathbf{x}_i||_\infty \leq C \Lambda_{\min}^{-1}(\mathbf{\Sigma}) \sqrt{\log(p \vee n)}$$

with probability at least $1 - C'p^{-c'}$ as desired. $\square$

*Proof of Lemma 3.* To simplify the notation, let $\xi_i(\tau) = \left( \mathbf{X}_i^T \left( \widehat{\boldsymbol{\beta}}(\tau) - \boldsymbol{\beta}^*(\tau) \right) \right)^2$. With this we have

$$\left\| n^{-1} \sum_{i \in [n]} f(\bar{w}_i) \widetilde{\mathbf{D}}_\tau \mathbf{X}_i \xi_i \right\|_\infty \leq \frac{f_{\max}}{f_{\min}} \frac{1}{n} \sum_{i \in [n]} \left\| \widehat{\mathbf{D}} \mathbf{X}_i \xi_i \right\|_\infty$$

$$\leq \frac{f_{\max} L_n}{f_{\min}} \frac{1}{n} \sum_{i \in [n]} |\xi_i|,$$

where the first inequality follows since $f(\bar{w}_i)/f(F^{-1}(\tau)) \leq f_{\max} f_{\min}^{-1}$ and the second inequality follows from the construction of $\widehat{\mathbf{D}}$ in (2.4). Recall that by Lemma 2 there is a solution to (2.4). Finally, using Theorem 2 in Belloni and Chernozhukov (2011), we have that

$$n^{-1} \sum_{i \in [n]} |\xi_i| \leq C \frac{s \log(p)}{n}$$

with probability at least $1 - Cp^{-c}$. $\square$

### 8.4 Proofs for Section 4

*Proof of Theorem 4.1.* Since the effect of penalization mostly disappears from the second order derivative $\widehat{\xi}_i''(\tau)$, we will make a connection between penalized regression rank scores $\widehat{\xi}_i(\tau)$ and regression rank scores $\widehat{\alpha}_i(\tau)$ of Gutenbrunner and Jurečková (1992). We split the proof into three parts. First, we approximate $S_n(\tau)$ by a linear function. Then, we continue by showing that the linear approximation is uniform over a range of values of $\tau$. Lastly, approximate the second order derivatives of $S_n(\tau)$.

**Part I.** Let $\mathbf{w} = \mathbf{P}_S^\perp \mathbf{E}_\tau$ with $\mathbf{P}_S^\perp = \mathbf{I} - \mathbf{X}_S \left(\mathbf{X}_S^T \mathbf{X}_S\right)^{-1} \mathbf{X}_S^T$. From Lemma 6, we have that

$$\sup_{\tau \in [\epsilon, 1-\epsilon]} |S_n(\tau) - \bar{T}_n(\tau)| = \mathcal{O}_P(R_n),$$

where

$$\bar{T}_n(\tau) = n^{-1} \sum_{i \in [n]} w_i \phi(F(u_i)) \mathbb{1}[F^{-1}(\tau) \leq u_i \leq F^{-1}(0)]$$

and

$$R_n = \mathcal{O}_P \left( \lambda_n \left( \sup_{\tau \in [\epsilon, 1-\epsilon]} ||\boldsymbol{\beta}^*(\tau)||_1 \right) + s \sqrt{\frac{\log(p \vee n)}{n}} \right).$$

We proceed to show that

$$\bar{T}_n(\tau) - S_F(\tau) = o_{\mathrm{P}}(1).$$

Gutenbrunner and Jurečková (1992) show that statistic $\bar{T}_n(\tau)$ is regression invariant. Therefore, we can write

$$\bar{T}_n(\tau) = n^{-1} \sum_{i \in [n]} \eta_i = n^{-1} \sum_{i \in [n]} u_i \phi(F(u_i)) \mathbb{1}[F^{-1}(\tau) \leq u_i \leq F^{-1}(0)]$$

with

$$\begin{aligned}
\mathbb{E}\eta_i &= \mathbb{E}\left(u_i \phi(F(u_i)) \mathbb{1}[\tau \leq F(u_i) \leq 0]\right) \\
&= \int_{-\infty}^{\infty} u\phi(F(u)) \mathbb{1}[\tau \leq F(u) \leq 0] f(u) du \\
&= \int_{\tau}^{0} F^{-1}(z) \phi(z) dz \\
&= S_F(\tau)
\end{aligned}$$

where second to last equality follow by a change of variables $F(u) = z$. Since $\eta_i$ is a bounded random variable, taking values in $[F^{-1}(\tau), F^{-1}(1/2)]$, Hoefding's inequality gives us

$$\mathbb{P}\left(|\bar{T}_n(\tau) - S_F(\tau)| \geq t\right) \leq 2 \exp\left\{-2 \frac{nt^2}{(F^{-1}(\tau) - F^{-1}(0))^2}\right\}.$$

This implies that for a fixed $\tau \in [\epsilon, 1-\epsilon]$ we have $S_n(\tau) = S_F(\tau) + \mathcal{O}_P(R_n)$.

**Part II.** Result established in Part I is pointwise in nature. Here, we establish the uniform convergence. For that purpose, due to Lemmas 4 and 5 it suffices to show stochastic equicontinuity of the sequence $T_n(\tau)$, that is

$$\mathbb{E}\left[\sup_{\epsilon \leq v \leq 1-\epsilon} \left\{ \frac{\partial}{\partial \tau} \left(-n^{-1} \sum_{i \in [n]} \int_{\tau}^{0} \phi(t) d\alpha_i(t)\right) \bigg|_{\tau=v} \right\} \right] < \infty. \tag{8.5}$$

Since

$$-\int_{\tau}^{0} \phi(t) d\alpha_i(t) = +\int_{0}^{\tau} \phi(t) \alpha_i'(t) dt \quad \text{and} \quad \frac{\partial}{\partial \tau} -\int_{\tau}^{0} \phi(t) d\alpha_i(t) = \phi(\tau) \alpha_i'(\tau),$$

together with $\sup_{\epsilon \leq \tau \leq 1-\epsilon} \phi(\tau) < 1$ we have that

$$\sup_{\epsilon \leq \tau \leq 1-\epsilon} \left( \frac{\partial}{\partial \tau} - n^{-1} \sum_{i \in [n]} \int_\tau^0 \phi(t) d\alpha_i(t) \right) \leq \sup_{\epsilon \leq \tau \leq 1-\epsilon} n^{-1} \sum_{i \in [n]} \alpha_i'(\tau).$$

Right hand derivative of the regression rank scores $\alpha$ are one-step functions defined as

$$\sum_{i \in N_n(\tau)} \mathbf{X}_{i,S} \alpha_i'(\tau) + (1-\tau) \mathbf{X}_S^T \mathbf{1}_n = \mathbf{0}_S$$

for $N_n(\tau) = \{i \mid Y_i = X_{i,S}^T \widehat{\boldsymbol{\beta}}^o(\tau)\}$ where $\widehat{\boldsymbol{\beta}}^o(\tau) = \arg\min \sum_{i \in [n]} \rho_\tau(Y_i - \mathbf{X}_{i,S}^T \boldsymbol{\beta})$ is the solution of the oracle problem. By the identifiability condition of the quantile loss problem, we have that $\exists \delta_s > 0$ such that

$$\|\boldsymbol{\alpha}'(\tau)\|_2 \leq \delta_s^{-1} \left\| n^{-1} \sum_{i \in N_n(\tau)} \mathbf{X}_{i,S} \alpha_i'(\tau) \right\|_2.$$

Let $\boldsymbol{\Sigma}_S$ be a $s \times s$ sub matrix of $\boldsymbol{\Sigma}$ indexed by $S$. Combining with the previous equation we have

$$\|\boldsymbol{\alpha}'(\tau)\|_2 \leq \frac{1-\tau}{\delta_s n} \|\mathbf{X}_S^T \mathbf{1}_n\|_2 = \frac{1}{\delta_s n} \sqrt{\mathbf{1}_n^\top \mathbf{X}_S \mathbf{X}_S^\top \mathbf{1}_n} \tag{8.6}$$

$$\leq \frac{1}{\delta_s n} \sqrt{\lambda_{\max}\left(\mathbf{X}_S \mathbf{X}_S^\top\right) \|\mathbf{1}_n\|_2^2} = \frac{1}{\delta_s} \sqrt{\lambda_{\max}\left(n^{-1} \mathbf{X}_S \mathbf{X}_S^\top\right)} \tag{8.7}$$

By Weyl's inequality we have that

$$|\lambda_{\max}(n^{-1} \mathbf{X}_S^T \mathbf{X}_S) - \lambda_{\max}(\boldsymbol{\Sigma}_S)| \leq \|n^{-1} \mathbf{X}_S^T \mathbf{X}_S - \boldsymbol{\Sigma}_S\|^2,$$

where $\|\|$ stands for a spectral norm of a matrix. By a matrix Bernstein inequality, see Corrolary 5.2 of Mackey et al. (2014), we have $\|n^{-1} \mathbf{X}_S^T \mathbf{X}_S - \boldsymbol{\Sigma}_S\| \leq \sqrt{\log p / n}$. Hence, from all of the above

$$\|\boldsymbol{\alpha}'(\tau)\|_\infty \leq \|\boldsymbol{\alpha}'(\tau)\|_2 \leq \frac{1}{\delta_s} \sqrt{\lambda_{\max}(\boldsymbol{\Sigma}) + \frac{\log p}{n}},$$

concluding that (8.5) holds and that representation of Part I holds uniformly in $\tau \in (\epsilon, 1-\epsilon)$.

**Part III.** Observe that

$$S_F(\tau + h) - S_F(\tau) = S_F'(\tau)h + \frac{1}{2} S_F''(\tau) h^2 + \frac{1}{6} S_F'''(\tau) h^3 + \frac{1}{24} S_F''''(\bar\tau) h^4, \tag{8.8}$$

and

$$S_F(\tau - h) - S_F(\tau) = -S_F'(\tau)h + \frac{1}{2} S_F''(\tau) h^2 - \frac{1}{6} S_F'''(\tau) h^3 + \frac{1}{24} S_F''''(\widetilde\tau) h^4, \tag{8.9}$$

for $\bar\tau \in (\tau, \tau+h)$ and $\widetilde\tau \in (\tau-h, \tau)$. Recall, that $S_F''(\tau) = 1/f(F^{-1}(\tau)) = \varsigma(\tau)$. Combining the last two equations gives us

$$\frac{S_F(\tau+h) - 2S_F(\tau) + S_F(\tau-h)}{h^2} = \varsigma(\tau) + \frac{1}{24} h^2 \left[ S_F''''(\bar\tau) + S_F''''(\widetilde\tau) \right].$$

Recall that $\widehat{\varsigma}(\tau) = h^{-2}\left(S_n(\tau+h) - 2S_n(\tau) + S_n(\tau-h)\right)$. Therefore,

$$\widehat{\varsigma}(\tau) - \varsigma(\tau) = \frac{1}{h^2}\left[S_n(\tau+h) - S_F(\tau+h) - 2\Big(S_n(\tau) - S_F(\tau)\Big) + S_n(\tau-h) - S_F(\tau-h)\right] + \mathcal{O}(h^2), \quad (8.10)$$

since

$$\sup_{\tau \in [\epsilon, 1-\epsilon]} \left|S_F''''(\bar{\tau}) + S_F''''(\widetilde{\tau})\right| \leq \sup_{\tau \in [\epsilon, 1-\epsilon]} \left|\frac{f(F^{-1}(\tau))f''(F^{-1}(\tau)) - 3f'^2(F^{-1}(\tau))}{f^5(F^{-1}(\tau))}\right| = \mathcal{O}(1)$$

and the RHS is bounded above by a constant using Condition $(\mathbf{D}')$.

Combining the results from the three steps, we have that

$$\sup_{\epsilon \leq \tau \leq 1-\epsilon} |\widehat{\varsigma}(\tau) - \varsigma(\tau)| = \mathcal{O}_P\left(h^{-2}R_n + h^2\right),$$

which completes the proof. □

*Proof of Lemma 4.* To simplify notation we define $\boldsymbol{\delta} \in \mathbb{R}^{p+1}$ as $\boldsymbol{\delta}_S = \widehat{\boldsymbol{\beta}}^o(\tau) - \boldsymbol{\beta}_S^*(\tau)$ and zeros elsewhere. Observe that we can write the check function as $\rho_\tau(z) = z\left(\mathbb{I}\{z \geq 0\} - (1-\tau)\right)$. Therefore

$$n^{-1} \sum_{i \in [n]} \left(E_{i\tau} - \mathbf{X}_i^T \boldsymbol{\delta}\right)[\alpha_i(\tau) - (1-\tau)] = n^{-1} \sum_{i \in [n]} \rho_\tau\left(E_{i\tau} - \mathbf{X}_i^T \boldsymbol{\delta}\right),$$

and

$$n^{-1} \sum_{i \in [n]} E_{i\tau}[\bar{\alpha}_i(\tau) - (1-\tau)] = n^{-1} \sum_{i \in [n]} \rho_\tau(E_{i\tau}).$$

Taking difference between the last two displays and combining with Knight's identity (2), we obtain

$$n^{-1} \sum_{i \in [n]} E_{i\tau}[\alpha_i(\tau) - \bar{\alpha}_i(\tau)] = n^{-1} \sum_{i \in [n]} \mathbf{X}_i^T \boldsymbol{\delta}[\alpha_i(\tau) - (1-\tau)] - n^{-1} \sum_{i \in [n]} \psi_\tau(E_{i\tau})\mathbf{X}_i^T \boldsymbol{\delta}$$
$$+ n^{-1} \sum_{i \in [n]} \int_0^{\mathbf{X}_i^T \boldsymbol{\delta}} [\mathbb{I}\{E_{i\tau} \leq s\} - \mathbb{I}\{E_{i\tau} \leq 0\}]\,ds.$$

Observe that $|\alpha_i(\tau)| \leq 1$ and $\psi_\tau(E_{i\tau}) \geq \tau - 1$ uniformly in $\tau$. Therefore,

$$n^{-1} \sum_{i \in [n]} E_{i\tau}[\alpha_i(\tau) - \bar{\alpha}_i(\tau)] \leq \tau n^{-1} \sum_{i \in [n]} \left|\mathbf{X}_i^T \boldsymbol{\delta}\right| + (1-\tau)n^{-1} \sum_{i \in [n]} \left|\mathbf{X}_i^T \boldsymbol{\delta}\right| + 2n^{-1} \sum_{i \in [n]} \left|\mathbf{X}_i^T \boldsymbol{\delta}\right|$$
$$\leq 3n^{-1} \left\|\mathbf{X}\boldsymbol{\delta}\right\|_1,$$

where we have used $\left|\int_a^b g(x)dx\right| \leq |b-a| \sup_{x \in [a,b]} |g(x)|$. The lemma follows since

$$n^{-1} \|\mathbf{X}\boldsymbol{\delta}\|_1 = n^{-1} \left\|\mathbf{X}_S\left(\widehat{\boldsymbol{\beta}}^o(\tau) - \boldsymbol{\beta}_S^*(\tau)\right)\right\|_1 = \mathcal{O}_P\left(s\sqrt{\frac{\log(n)}{n}}\right)$$

for all $\tau \in [\epsilon, 1-\epsilon]$ by adapting the proof of Theorem 2 in Belloni and Chernozhukov (2011). □

*Proof of Lemma 5.* Since the loss function $\rho_\tau(\cdot)$ is Lipschitz, we have that

$$\left|\rho_\tau\left(E_{i\tau} - \mathbf{X}_i^T \widehat{\boldsymbol{\beta}}^\circ(\tau)\right) - \rho_\tau\left(E_{i\tau} - \mathbf{X}_i^T \widehat{\boldsymbol{\beta}}(\tau)\right)\right| \leq \max\{\tau, 1-\tau\} |\mathbf{X}_i^T \widehat{\boldsymbol{\beta}}^\circ(\tau) - \mathbf{X}_i^T \widehat{\boldsymbol{\beta}}(\tau)|. \qquad (8.11)$$

The proof of the lemma is continued by analyzing separately the left and the right hand side of the previous inequality.

Since the strong duality holds, we have that

$$n^{-1} \sum_{i \in [n]} \rho_\tau(Y_i - \mathbf{X}_i^\top \widehat{\boldsymbol{\beta}}(\tau)) + \lambda_n \|\widehat{\boldsymbol{\beta}}(\tau)\|_1$$

$$= n^{-1} \sum_{i \in [n]} Y_i \left[\widehat{\xi}_i(\tau) - (1-\tau)\right]$$

$$= n^{-1} \sum_{i \in [n]} (Y_i - \mathbf{X}_i^T \widehat{\boldsymbol{\beta}}(\tau)) \left[\widehat{\xi}_i(\tau) - (1-\tau)\right] + n^{-1} \sum_{i \in [n]} \mathbf{X}_i^T \widehat{\boldsymbol{\beta}}(\tau) \left[\widehat{\xi}_i(\tau) - (1-\tau)\right].$$

Furthermore, we have that

$$\max_{j \in [p]} \left| \sum_{i \in [n]} X_{ij} \left[\widehat{\xi}_i(\tau) - (1-\tau)\right] \right| \leq \lambda_n. \qquad (8.12)$$

Let $\boldsymbol{\delta} = \widehat{\boldsymbol{\beta}}(\tau) - \boldsymbol{\beta}^*(\tau)$. We can write the display above as

$$n^{-1} \sum_{i \in [n]} \left(E_{i\tau} - \mathbf{X}_i^T \boldsymbol{\delta}\right) \left[\widehat{\xi}(\tau) - (1-\tau)\right] + n^{-1} \sum_{i \in [n]} \mathbf{X}_i^T \widehat{\boldsymbol{\beta}}(\tau) \left[\widehat{\xi}_i(\tau) - (1-\tau)\right] - \lambda_n \|\widehat{\boldsymbol{\beta}}(\tau)\|_1$$

$$= n^{-1} \sum_{i \in [n]} \rho_\tau(E_{i\tau} - \mathbf{X}_i^T \boldsymbol{\delta}).$$

Similarly, let $\boldsymbol{\delta}^\circ = \widehat{\boldsymbol{\beta}}^\circ(\tau) - \boldsymbol{\beta}(\tau)$ and write

$$n^{-1} \sum_{i \in [n]} \left(E_{i\tau} - \mathbf{X}_i^T \boldsymbol{\delta}^\mathbf{o}\right) (\alpha_i(\tau) - (1-\tau)) = n^{-1} \sum_{i \in [n]} \rho_\tau \left(E_{i\tau} - \mathbf{X}_i^T \boldsymbol{\delta}^\circ\right),$$

for the oracle problem. Taking the difference of the right hand side of the last two equations we get that

$$n^{-1} \sum_{i \in [n]} \rho_\tau\left(E_{i\tau} - \mathbf{X}_i^T \boldsymbol{\delta}^\circ\right) - n^{-1} \sum_{i \in [n]} \rho_\tau\left(E_{i\tau} - \mathbf{X}_i^T \boldsymbol{\delta}\right)$$

$$= n^{-1} \sum_{i \in [n]} E_{i\tau} \left(\alpha_i(\tau) - \widehat{\xi}_i(\tau)\right) - n^{-1} \sum_{i \in [n]} \mathbf{X}_i^T \boldsymbol{\delta}^\mathbf{o} \left[\alpha_i(\tau) - (1-\tau)\right]$$

$$- n^{-1} \sum_{i \in [n]} \mathbf{X}_i^T \boldsymbol{\beta}^*(\tau) \left[\widehat{\xi}_i(\tau) - (1-\tau)\right] + \lambda_n \|\widehat{\boldsymbol{\beta}}(\tau)\|_1.$$

Using (8.11), it follows from the display above that

$$n^{-1} \sum_{i \in [n]} E_{i\tau} \left( \alpha_i(\tau) - \widehat{\xi}_i(\tau) \right)$$

$$\leq \underbrace{n^{-1} \sum_{i \in [n]} \mathbf{X}_i^T \boldsymbol{\delta}^\circ \left[ \alpha_i(\tau) - (1-\tau) \right]}_{T_1} - \underbrace{n^{-1} \sum_{i \in [n]} \mathbf{X}_i^T \boldsymbol{\beta}^*(\tau) \left[ \widehat{\xi}_i(\tau) - (1-\tau) \right]}_{T_2} \qquad (8.13)$$

$$- \underbrace{\lambda_n \|\widehat{\boldsymbol{\beta}}(\tau)\|_1}_{T_3} + \underbrace{\max\{\tau, 1-\tau\} n^{-1} \sum_{i \in [n]} |\mathbf{X}_i^T [\boldsymbol{\delta}^\circ - \boldsymbol{\delta}]|}_{T_4}.$$

We bound each of the four terms individually. From the optimality conditions for the oracle problem in (4.2), we have that

$$n^{-1} \sum_{i \in [n]} X_{ij} [\alpha_i(\tau) - (1-\tau)] = 0, \qquad (8.14)$$

for each $j \in S$. Therefore, $T_1 = 0$.

For the term $T_2$, we have

$$T_2 = -n^{-1} \sum_{i \in [n]} \sum_{j \in [p]} X_{ij} \beta_j^*(\tau) [\widehat{\xi}_i(\tau) - (1-\tau)]$$

$$\overset{(i)}{\leq} n^{-1} \max_{j \in [p]} \left| \sum_{i \in [n]} X_{ij} [\widehat{\xi}_i(\tau) - (1-\tau)] \right| \cdot \|\boldsymbol{\beta}^*(\tau)\|_1$$

$$\overset{(ii)}{\leq} n^{-1} \lambda_n \|\boldsymbol{\beta}^*(\tau)\|_1,$$

where $(i)$ is a simple application of Hoelder's inequality and $(ii)$ an application of (8.12).

For the term $T_3$, the triangle inequality leads to

$$T_3 \leq \lambda_n \left( \|\boldsymbol{\delta}\|_1 + \|\boldsymbol{\beta}^*(\tau)\|_1 \right) \leq \lambda_n \left( \|\boldsymbol{\beta}^*(\tau)\|_1 + \mathcal{O}_P(s\lambda_n) \right)$$

where the last inequality follows from (3.1).

For the term $T_4$, we have

$$T_4 \leq n^{-1} \|\mathbf{X}\boldsymbol{\delta}^\circ\|_1 + n^{-1} \|\mathbf{X}\boldsymbol{\delta}\|_1 = \mathcal{O}_P \left( s \sqrt{\frac{\log(p \vee n)}{n}} \right)$$

similar as in the proof of Lemma 4 by adapting the proof of Theorem 2 and 3 in Belloni and Chernozhukov (2011)

Combining the bounds on $T_1, T_2, T_3$ and $T_4$, we have

$$\left| n^{-1} \sum_{i \in [n]} E_{i\tau} \left( \alpha_i(\tau) - \widehat{\xi}_i(\tau) \right) \right| = \mathcal{O}_P \left( \lambda_n \|\boldsymbol{\beta}^*(\tau)\|_1 + s \sqrt{\frac{\log(p \vee n)}{n}} \right). \qquad (8.15)$$

Next, observe that

$$\left| n^{-1} \sum_{i \in [n]} Y_i(\alpha_i(\tau) - \widehat{\xi}_i(\tau)) \right| \leq \left| n^{-1} \sum_{i \in [n]} E_{i\tau}(\alpha_i(\tau) - \widehat{\xi}_i(\tau)) \right| + \left| n^{-1} \sum_{i \in [n]} \mathbf{X}_i^T \boldsymbol{\beta}^*(\tau)(\alpha_i(\tau) - \widehat{\xi}_i(\tau)) \right|$$
$$\leq T_5 + T_6.$$

Now, $T_5$ is bounded with (8.15). Moreover

$$T_6 \leq \left| n^{-1} \sum_{i \in [n]} \mathbf{X}_i^T(\alpha_i(\tau) - (1-\tau))\boldsymbol{\beta}^*(\tau) \right| + \max_{j \in [p]} \left| n^{-1} \sum_{i \in [n]} X_{ij}(\widehat{\xi}_i(\tau) - (1-\tau)) \right| \cdot \|\boldsymbol{\beta}^*(\tau)\|_1$$
$$\leq n^{-1} \lambda_n \|\boldsymbol{\beta}^*(\tau)\|_1,$$

using (8.14) and (8.12), respectively. Therefore, we have

$$\left| \frac{1}{n} \sum_{i \in [n]} Y_i(\alpha_i(\tau) - \widehat{\xi}_i(\tau)) \right| = \mathcal{O}_P\left( \lambda_n \|\boldsymbol{\beta}^*(\tau)\|_1 + s\sqrt{\frac{\log(p \vee n)}{n}} \right), \tag{8.16}$$

which concludes the proof. $\square$

*Proof of Lemma 6.* The proof requires three steps. In the first we approximate the penalized rank process $\widehat{\boldsymbol{\xi}}(\tau)$ with the the rank process $\boldsymbol{\alpha}(\tau)$. In the second step, we approximate the process $\boldsymbol{\alpha}(\tau)$ with a process $\mathbb{I}\{E_{i\tau} \geq 0\}$. In the third step we show that the representation of linear combination of $\mathbb{I}\{E_{i\tau} \geq 0\}$ has the exact format as in (4.9).

**Step 1.** The first step of the proof is given in Lemma 5.

**Step 2.** Let the score statistic $T_n(\tau)$ be defined as

$$T_n(\tau) = -n^{-1} \sum_{i \in [n]} Y_i \int_\tau^0 \phi(t) d\alpha_i(t), \tag{8.17}$$

where $\boldsymbol{\alpha}(\tau)$ is the dual oracle solution in (4.2). Notice that $T_n(\tau)$ is analogous to $S_n(\tau)$, however, with $\boldsymbol{\alpha}(\tau)$ substituting $\widehat{\boldsymbol{\xi}}(\tau)$. The score statistics $T_n(\tau)$ is regression invariant and hence can be represented as

$$T_n(\tau) = -n^{-1} \sum_{i \in [n]} w_i \int_\tau^0 \phi(t) d\alpha_i(t),$$

where $\mathbf{w} \in \mathbb{R}^n$ is defined as

$$\mathbf{w} = \mathbf{P}_S^\perp \mathbf{Y} = \mathbf{P}_S^\perp \mathbf{E}_\tau$$

with $\mathbf{P}_S^\perp = \mathbf{I} - \mathbf{X}_S \left(\mathbf{X}_S^T \mathbf{X}_S\right)^{-1} \mathbf{X}_S^T$.

We proceed to show that

$$\sup_{\tau \in [\epsilon, 1-\epsilon]} n^{-1} \sum_{i \in [n]} w_i \int_\tau^0 \phi(t) |d\alpha_i(t) - d\bar{\alpha}_i(t)| = O_P\left( s\sqrt{\frac{\log n}{n}} \right), \tag{8.18}$$

where $\bar{\alpha}_i(t) = \mathbb{1}\{E_{i\tau} \geq 0\}$. Let

$$W(\tau) = n^{-1} \sum_{i \in [n]} w_i \left[\alpha_i(\tau) - (1-\tau)\right]$$

and

$$\bar{W}(\tau) = n^{-1} \sum_{i \in [n]} w_i \left[\bar{\alpha}_i(\tau) - (1-\tau)\right]$$

and $\mathbf{\Delta}_\alpha(\tau) = \boldsymbol{\alpha}(\tau) - \bar{\boldsymbol{\alpha}}(\tau)$. Observe that

$$\begin{aligned}
W(\tau) - \bar{W}(\tau) &= n^{-1} \sum_{i \in [n]} w_i \left[\alpha_i(\tau) - \bar{\alpha}_i(\tau)\right] \\
&= n^{-1} \operatorname{tr}(\mathbf{P}_{\bar{S}}^{\perp} \mathbf{E}_\tau \mathbf{\Delta}_\alpha^\top) \\
&\overset{(i)}{\leq} n^{-1} \sum_{i \in [n]} \sigma_i\left(\mathbf{P}_{\bar{S}}^{\perp}\right) \sigma_i\left(\mathbf{E}_\tau \mathbf{\Delta}_\alpha^\top\right) \\
&\overset{(ii)}{\leq} \sigma_{\max}\left(\mathbf{P}_{\bar{S}}^{\perp}\right) n^{-1} \operatorname{tr}\left(\mathbf{E}_\tau \mathbf{\Delta}_\alpha^\top\right) \\
&\overset{(iii)}{=} n^{-1} \operatorname{tr}\left(\mathbf{E}_\tau \mathbf{\Delta}_\alpha^\top\right),
\end{aligned}$$

where $(i)$ is due to Von-Neumann's trace inequality, $(ii)$ is due to the equality $\sum_{i \in [n]} \sigma_i\left(\mathbf{E}_\tau \mathbf{\Delta}_\alpha^\top\right) = \operatorname{tr}(\mathbf{E}_\tau \mathbf{\Delta}_\alpha^\top)$, and $(iii)$ follows since the largest singular value of the projection matrix is equal to 1. Moreover, by Lemma 4, $\sup_{\tau \in [\epsilon, 1-\epsilon]} n^{-1} \operatorname{tr}(\mathbf{E}_\tau \mathbf{\Delta}_\alpha^\top) = \mathcal{O}_P(s\sqrt{\log n/n})$, which gives us (8.18) easily.

Substituting (8.18) into the definition of the scale statistics $T_n(\tau)$, (8.17), we obtain

$$T_n(\tau) = \underbrace{-n^{-1} \sum_{i \in [n]} w_i \int_\tau^0 \phi(t) d\bar{\alpha}_i(t)}_{\bar{T}_n(\tau)} + \mathcal{O}_P\left(s\sqrt{\frac{\log n}{n}}\right), \quad (8.19)$$

which completes the proof of the second step.

**Step 3.** The proof is finalized by showing that $\bar{T}_n(\tau)$ has the desired form in (4.9). Let

$$\bar{b}_i = -\int_\tau^0 \phi(t) d\bar{\alpha}_i(t), \quad \text{and} \quad \bar{T}_n(\tau) = n^{-1} \sum_{i \in [n]} w_i \bar{b}_i.$$

By partial integration we have

$$\bar{b}_i = \bar{\alpha}_i(\tau)\phi(\tau) - \bar{\alpha}_i(0)\phi(0) + \int_\tau^0 \bar{\alpha}_i(t) d\phi(t).$$

The function $\phi$ satisfies $-\phi(1-\tau) = \phi(\tau)$, leading to

$$\bar{T}_n(\tau) = \underbrace{n^{-1} \sum_{i \in [n]} w_i \bar{\alpha}_i(\tau)\phi(\tau)}_{\bar{T}_1(\tau)} + \underbrace{n^{-1} \sum_{i \in [n]} w_i \bar{\alpha}_i(0)\phi(0)}_{\bar{T}_2(\tau)} + \underbrace{n^{-1} \sum_{i \in [n]} w_i \int_\tau^0 \bar{\alpha}_i(t) d\phi(t)}_{\bar{T}_3(\tau)}.$$

Let us concentrate on the third term first. Plugging in $\bar{\alpha}_i(t) = \mathbb{1}\{u_i \geq F^{-1}(t)\} = \mathbb{1}\{F(u_i) \geq t\}$, we obtain

$$\bar{T}_3(\tau) = n^{-1} \sum_{i \in [n]} w_i \int_\tau^0 \mathbb{1}\{u_i \geq F^{-1}(t)\} d\phi(t)$$

$$= n^{-1} \sum_{i \in [n]} w_i \int_{F^{-1}(\tau)}^{F^{-1}(0)} \mathbb{1}\{u_i \geq z\} \phi'(F(z)) f(z) dz,$$

where the second line follows from a change of variables, $t = F(z)$. We split the term $\bar{T}_3(\tau)$ into three factors, $\bar{T}_{31}$, $\bar{T}_{32}$ and $\bar{T}_{33}$, according to the value of $u_i$: the first term for $u_i \leq F^{-1}(t)$, the second for $F^{-1}(t) \leq u_i \leq F^{-1}(0)$ and the last for $u_i \geq F^{-1}(0)$. If $u_i \leq F^{-1}(t)$, then $\bar{\alpha}_i(t) = 0$ and $\bar{T}_{31} = 0$. If $F^{-1}(t) \leq u_i \leq F^{-1}(0)$, then

$$\bar{T}_{32} = n^{-1} \sum_{i \in [n]} w_i \int_{F^{-1}(\tau)}^{u_i} \phi'(F(z)) f(z) dz.$$

With a change of variable $z = F^{-1}(t)$ we have $dz = dt/f(F^{-1}(t))$ and

$$\bar{T}_{32} = n^{-1} \sum_{i \in [n]} w_i \int_\tau^{F(u_i)} \phi'(t) f(F^{-1}(t)) \frac{dt}{f(F^{-1}(t))} = n^{-1} \sum_{i \in [n]} w_i \int_\tau^{F(u_i)} \phi'(t) dt$$

$$= n^{-1} \sum_{i \in [n]} w_i [\phi(F(u_i)) - \phi(\tau)].$$

Finally, if $u_i \geq F^{-1}(0)$, then $\bar{\alpha}_i(t) = 1$ and

$$\bar{T}_{33} = n^{-1} \sum_{i \in [n]} w_i \int_{F^{-1}(\tau)}^{F^{-1}(0)} \phi'(F(z)) f(z) dz = n^{-1} \sum_{i \in [n]} w_i \int_\tau^0 \phi'(t) dt$$

$$= n^{-1} \sum_{i \in [n]} w_i [\phi(0) - \phi(\tau)].$$

Putting things together, we have

$$\bar{T}_3(\tau) = n^{-1} \sum_{i \in [n]} w_i \phi(F(u_i)) \mathbb{1}\{F^{-1}(\tau) \leq u_i \leq F^{-1}(0)\}$$

$$- n^{-1} \sum_{i \in [n]} w_i \phi(\tau) \mathbb{1}\{u_i \geq F^{-1}(\tau)\}$$

$$+ n^{-1} \sum_{i \in [n]} w_i \phi(1 - \tau) \mathbb{1}\{u_i \geq F^{-1}(0)\}$$

Writing down explicitly $\bar{T}_1(\tau)$ and $\bar{T}_2(\tau)$ we have

$$\bar{T}_1(\tau) = n^{-1} \sum_{i \in [n]} w_i \phi(\tau) \mathbb{1}\{u_i \geq F^{-1}(\tau)\}$$

and

$$\bar{T}_2(\tau) = -n^{-1} \sum_{i \in [n]} w_i \phi(1-\tau) \, \mathbb{I}\left\{u_i \geq F^{-1}(0)\right\},$$

which leads to

$$\bar{T}_n(\tau) = n^{-1} \sum_{i \in [n]} w_i \phi(F(u_i)) \, \mathbb{I}\left\{F^{-1}(\tau) \leq u_i \leq F^{-1}(0)\right\},$$

thus proving (4.9). □

*Proof of Theorem 4.2.* Starting from Proposition 2.1, we need to bound

$$\sqrt{n}\left(\boldsymbol{\Delta}_1(\tau) + \boldsymbol{\Delta}_2(\tau) + \boldsymbol{\Delta}_3(\tau)\right)$$

uniformly in $\tau$.

Using Lemma 4.1, we have that $\sup_{\tau \in [\epsilon, 1-\epsilon]} \widehat{\varsigma}(\tau)/\varsigma(\tau) = \mathcal{O}_P(1)$. Combined with Lemma 1 with $r_n = C\sqrt{s \log(p \vee n)/n}$ and $t = Cs$, we obtain

$$\sup_{\tau \in [\epsilon, 1-\epsilon]} \sqrt{n}\boldsymbol{\Delta}_1(\tau) = \sup_{\tau \in [\epsilon, 1-\epsilon]} \sqrt{n} G_n\left(\tau, \widehat{\boldsymbol{\beta}}(\tau) - \boldsymbol{\beta}^*(\tau)\right) = \mathcal{O}_P\left(\frac{L_n s^{3/4} \log^{3/4}(p \vee n)}{n^{1/4}} \vee \frac{L_n s \log(p \vee n)}{n^{1/2}}\right).$$

Similarly

$$\sup_{\tau \in [\epsilon, 1-\epsilon]} \sqrt{n}\boldsymbol{\Delta}_3(\tau) = \sup_{\tau \in [\epsilon, 1-\epsilon]} n^{-1/2} \sum_{i \in [n]} f(\bar{w}_i) \widetilde{\mathbf{D}}_\tau \mathbf{X}_i \left(\mathbf{X}_i^T \left(\widehat{\boldsymbol{\beta}}(\tau) - \boldsymbol{\beta}^*(\tau)\right)\right)^2 = \mathcal{O}_P\left(\frac{L_n s \log(p \vee n)}{n^{1/2}}\right).$$

Finally, Lemma 2 provides

$$\sup_{\tau \in [\epsilon, 1-\epsilon]} \sqrt{n}\boldsymbol{\Delta}_2(\tau) \leq \sqrt{n}\gamma_n \sup_{\tau \in [\epsilon, 1-\epsilon]} ||\widehat{\boldsymbol{\beta}}(\tau) - \boldsymbol{\beta}(\tau)||_1 \left|\frac{\widehat{\varsigma}(\tau)}{\varsigma(\tau)}\right| + \sqrt{n} \sup_{\tau \in [\epsilon, 1-\epsilon]} ||\widehat{\boldsymbol{\beta}}(\tau) - \boldsymbol{\beta}(\tau)||_1 \left|\frac{\widehat{\varsigma}(\tau)}{\varsigma(\tau)} - 1\right|$$

$$= \mathcal{O}_p\left(\frac{s \log(p \vee n)}{n^{1/2}} \vee \frac{s \log^{3/4} p}{n^{1/4}}\right),$$

where the last step follows from

$$\sup_{\tau \in [\epsilon, 1-\epsilon]} \left|\frac{\widehat{\varsigma}(\tau)}{\varsigma(\tau)} - 1\right| = \mathcal{O}_P(\sqrt{s} \log^{1/4} p / n^{1/4})$$

using Lemma 4.1. Combining all the results, the proof is complete.

□

## 8.5 Proofs for Section 5

*Proof of Theorem 5.1.* The proof follows easily from Lemmas 7 - 10 and Theorem 4.1. □

*Proof of Lemma 7.* From Proposition 2.1, we have that

$$\frac{\sqrt{n}\mathbf{x}^\top(\check{\boldsymbol{\beta}}(\tau) - \boldsymbol{\beta}^*(\tau))}{\widehat{\zeta}(\tau)\sqrt{\tau(1-\tau)\mathbf{x}^\top\widehat{\mathbf{D}}\widehat{\boldsymbol{\Sigma}}\widehat{\mathbf{D}}\mathbf{x}}} = n^{-1/2}\sum_{i\in[n]}\frac{\mathbf{x}^\top\widehat{\mathbf{D}}\mathbf{X}_i\psi_\tau\left(u_i - F^{-1}(\tau)\right)}{\sqrt{\tau(1-\tau)\mathbf{x}^\top\widehat{\mathbf{D}}\widehat{\boldsymbol{\Sigma}}\widehat{\mathbf{D}}\mathbf{x}}}$$
$$- \frac{\sqrt{n}\mathbf{x}^\top\left(\boldsymbol{\Delta}_1(\tau) + \boldsymbol{\Delta}_2(\tau) + \boldsymbol{\Delta}_3(\tau)\right)}{\widehat{\zeta}(\tau)\sqrt{\tau(1-\tau)\mathbf{x}^\top\widehat{\mathbf{D}}\widehat{\boldsymbol{\Sigma}}\widehat{\mathbf{D}}\mathbf{x}}} \qquad (8.20)$$
$$=: n^{-1/2}\sum_{i\in[n]} Z_i + B.$$

Given $\{X_i\}_{i\in[n]}$, we have that $Z_i$ has mean zero and variance equal to one. The Berry-Esseen theorem for independent but not identically distributed random variables (Feller, 1968) gives us

$$\sup_t\left|\mathbb{P}\left(n^{-1/2}\sum_{i\in[n]} Z_i \leq t\right) - \Phi(t)\right| \leq Cn^{-3/2}\sum_{i=1}^n \mathbb{E}_{u_i}|Z_i|^3,$$

where $\mathbb{E}_{u_i}$ denotes the expectation with respect to the probability measure generated by the random noise $u_i$. Let $a_i = \frac{\mathbf{x}^\top\widehat{\mathbf{D}}\mathbf{X}_i}{\sqrt{\tau(1-\tau)\mathbf{x}^\top\widehat{\mathbf{D}}\widehat{\boldsymbol{\Sigma}}\widehat{\mathbf{D}}\mathbf{x}}}$. Then

$$n^{-3/2}\sum_{i\in[n]}\mathbb{E}_{u_i}\left[\left|a_i\psi_\tau\left(u_i - F^{-1}(\tau)\right)\right|^3\right]$$
$$\leq n^{-3/2}||\mathbf{x}||_1 L_n\left(\frac{\sum_{i\in[n]}\mathbf{x}^\top\widehat{\mathbf{D}}X_iX_i^\top\widehat{\mathbf{D}}\mathbf{x}}{\left(\tau(1-\tau)\mathbf{x}^\top\widehat{\mathbf{D}}\widehat{\boldsymbol{\Sigma}}\widehat{\mathbf{D}}\mathbf{x}\right)^{3/2}}\right)\cdot\mathbb{E}_{u_i}\left[\left|\psi_\tau\left(u_i - F^{-1}(\tau)\right)\right|^3\right] \qquad (8.21)$$
$$\leq Cn^{-1/2}||\mathbf{x}||_1 L_n\left(\mathbf{x}^\top\widehat{\mathbf{D}}\widehat{\boldsymbol{\Sigma}}\widehat{\mathbf{D}}\mathbf{x}\right)^{-1/2}.$$

We proceed to establish a lower bound on $\mathbf{x}^\top\widehat{\mathbf{D}}\widehat{\boldsymbol{\Sigma}}\widehat{\mathbf{D}}\mathbf{x}$. We have that

$$\mathbf{x}^\top\widehat{\mathbf{D}}\widehat{\boldsymbol{\Sigma}}\widehat{\mathbf{D}}\mathbf{x} = \mathbf{x}^\top\boldsymbol{\Sigma}^{-1}\mathbf{x} - \mathbf{x}^\top\left(\widehat{\mathbf{D}}\widehat{\boldsymbol{\Sigma}}\widehat{\mathbf{D}} - \boldsymbol{\Sigma}^{-1}\right)\mathbf{x}$$
$$\geq \Lambda_{\max}^{-1}(\boldsymbol{\Sigma}) - \left|\mathbf{x}^\top\left(\widehat{\mathbf{D}}\widehat{\boldsymbol{\Sigma}}\widehat{\mathbf{D}} - \boldsymbol{\Sigma}^{-1}\right)\mathbf{x}\right| \qquad (8.22)$$

For the second term in (8.22), we have

$$\left|\mathbf{x}^\top\left(\widehat{\mathbf{D}}\widehat{\boldsymbol{\Sigma}}\widehat{\mathbf{D}} - \boldsymbol{\Sigma}^{-1}\right)\mathbf{x}\right| \leq \|\mathbf{x}\|_1 \cdot \max_{j\in[p+1]}\left|\mathbf{e}_j^\top\left(\widehat{\mathbf{D}}\widehat{\boldsymbol{\Sigma}}\widehat{\mathbf{D}} - \boldsymbol{\Sigma}^{-1}\right)\mathbf{x}\right|$$
$$\leq \|\mathbf{x}\|_1^2 \cdot \left\|\widehat{\mathbf{D}}\widehat{\boldsymbol{\Sigma}}\widehat{\mathbf{D}} - \boldsymbol{\Sigma}^{-1}\right\|_{\max}. \qquad (8.23)$$

Finally,

$$
\begin{aligned}
\left\|\widehat{\mathbf{D}}\widehat{\boldsymbol{\Sigma}}\widehat{\mathbf{D}} - \boldsymbol{\Sigma}^{-1}\right\|_{\max} &= \left\|\left(\widehat{\mathbf{D}} - \boldsymbol{\Sigma}^{-1}\right)\widehat{\boldsymbol{\Sigma}}\widehat{\mathbf{D}} + \boldsymbol{\Sigma}^{-1}\left(\widehat{\boldsymbol{\Sigma}}\widehat{\mathbf{D}} - \mathbf{I}\right)\right\|_{\max} \\
&\leq \max_i \left\|\mathbf{e}_i^\top\left(\widehat{\mathbf{D}} - \boldsymbol{\Sigma}^{-1}\right)\right\|_1 \cdot \max_{j,k}\left(\widehat{\boldsymbol{\Sigma}}\widehat{\mathbf{D}}\right)_{kj} + \left\|\boldsymbol{\Sigma}^{-1}\right\|_1 \cdot \left\|\widehat{\boldsymbol{\Sigma}}\widehat{\mathbf{D}} - \mathbf{I}\right\|_{\max} \quad (8.24) \\
&\leq \left(\max_i \left\|\mathbf{e}_i^\top\left(\widehat{\mathbf{D}} - \boldsymbol{\Sigma}^{-1}\right)\right\|_1 + \left\|\boldsymbol{\Sigma}^{-1}\right\|_1\right) \cdot \left\|\widehat{\boldsymbol{\Sigma}}\widehat{\mathbf{D}} - \mathbf{I}\right\|_{\max} + \max_i \left\|\mathbf{e}_i^\top\left(\widehat{\mathbf{D}} - \boldsymbol{\Sigma}^{-1}\right)\right\|_1 \\
&\leq C\gamma_n \bar{s} M,
\end{aligned}
$$

where the last statement follows as a consequence of Lemma 1 and Theorem 6 of Cai et al. (2011). Combining the last three displays, we have

$$\mathbf{x}^\top \widehat{\mathbf{D}}\widehat{\boldsymbol{\Sigma}}\widehat{\mathbf{D}}\mathbf{x} \geq \Lambda_{\max}^{-1}(\boldsymbol{\Sigma}) - C\|x\|_1^2 M \gamma_n \bar{s}.$$

Combining with (8.21), we have

$$\sup_t \left| \mathbb{P}\left( n^{-1/2} \sum_{i\in[n]} Z_i \leq t \right) - \Phi(t) \right| \leq C n^{-1/2} \|\mathbf{x}\|_1 L_n \Lambda_{\max}^{1/2}(\boldsymbol{\Sigma})$$

under the conditions of the lemma.

For the term $B$ in (8.20), similar to the proof of Theorem 3.1 and Theorem 4.2, we have that

$$|B| = \mathcal{O}_P\left( \frac{L_n s^{3/4} \log^{3/4}(p\vee n)}{n^{1/4}} \vee \frac{L_n s \log(p\vee n)}{n^{1/2}} \vee \frac{s\log^{3/4}(p\vee n)}{n^{1/4}} \cdot \|x\|_1 \cdot \Lambda_{\max}^{1/2}(\boldsymbol{\Sigma}) \right). \quad (8.25)$$

Using Lemma B.3 of Barber and Kolar (2015), we have

$$
\begin{aligned}
&\sup_{t\in\mathbb{R}} \left| \mathbb{P}\left( \frac{\sqrt{n}\mathbf{x}^\top(\check{\boldsymbol{\beta}}(\tau) - \boldsymbol{\beta}^*(\tau))}{\widehat{\zeta}(\tau)\sqrt{\tau(1-\tau)\mathbf{x}^\top \widehat{\mathbf{D}}\widehat{\boldsymbol{\Sigma}}\widehat{\mathbf{D}}\mathbf{x}}} \leq t \right) - \Phi(t) \right| \\
&\leq C\left( \|x\|_1 \cdot \Lambda_{\max}^{1/2}(\boldsymbol{\Sigma}) \cdot \left( n^{-1/2} L_n \vee n^{-1/4} s \log^{3/4}(p\vee n)\right) \vee \frac{L_n s^{3/4} \log^{3/4}(p\vee n)}{n^{1/4}} \vee \frac{L_n s \log(p\vee n)}{n^{1/2}} \right).
\end{aligned}
$$
(8.26)

Since the right hand side does not depend on $\boldsymbol{\beta}^*(\tau)$ the result follows.

$\square$

*Proof of Lemma 8.* First we consider the statements conditional on the covariates $\mathbf{X}_i$. Let $a_i = \mathbf{e}^\top \widehat{\mathbf{D}} \mathbf{X}_i \in \mathbb{R}$ be a univariate random variable. Let $\delta > 0$ be a small positive constant and let $C > 0$ be a constant independent of $n$ and $p$.

Consider an event of interest

$$\Omega = \left\{ \sup_{\alpha\in(0,1)} n^{-1} \sum_{i\in[n]} (\alpha a_i)^2 \leq C, \max_{i\in[n]} |a_i| \leq u_p, 1/u_p \leq 1, u_p^6 = o(n^{1-\delta}) \right\}. \quad (8.27)$$

By the symmetry of the $\widehat{D}$ we can write $a_i = \sum_{j=1}^{p} e_j \mathbf{X}_i^\top \widehat{d}_j$. We next observe that the third constraint of the hessian estimator (2.4) guarantees $|\mathbf{X}_i^\top \widehat{d}_j| \leq L_n$. Then, $\max_{1 \leq i \leq n} |a_i| \leq \|\mathbf{e}\|_1 L_n$, hence obtaining $u_p$ of the event $\Omega$ to be $u_n = \|\mathbf{e}\|_1 L_n$. Additionally Lemma 12 guarantees that $\sup_{\alpha \in (0,1)} n^{-1} \sum_{i \in [n]} (\alpha a_i)^2 \leq C$ happens with high probability for the choice of $a_i$ as the above.

Per these results, $\mathbb{P}(\Omega) \to 1$ as $n, p \to \infty$. Then, Lemma 14 of Belloni et al. (2011) with $v_i = 1$ and $Z_i = a_i$ (where $v_i$ and $Z_i$ are the notation therein) implies that there exists a process $B^*(\cdot) = B_n^*(\cdot)$ on $\mathcal{T} \subset (0,1)$ that, conditionally on $\mathbf{X}_i$ is a zero-mean Gaussian process with a.s. continuous sample paths and the covariance structure

$$(\tau \wedge \tau' - \tau \tau') \mathbf{e}^\top \widehat{\mathbf{D}} \widehat{\mathbf{\Sigma}} \widehat{\mathbf{D}} \mathbf{e}$$

such that

$$\sup_{\tau \in \mathcal{T}} \left| \frac{1}{\sqrt{n}} \sum_{i \in [n]} a_i \psi(u_i - F^{-1}(\tau)) - B^*(\tau) \right| \leq o(n^{-\epsilon'})$$

for some $\epsilon' > 0$. The process $B^*(\cdot)$ is a Brownian motion with respect to the filtration $\mathcal{F}_\tau^u = \sigma(\psi(u - F^{-1}(\tau)); \tau \in \mathcal{T})$.

Let $\mathbf{M} = \widehat{\mathbf{D}} \widehat{\mathbf{\Sigma}} \widehat{\mathbf{D}}$ and let $\mathbf{M}^{1/2}$ denote the square root of the matrix $\mathbf{M}$. Note that the covariance function of the process $B^*(\cdot)$ conditional on $\mathbf{X}_i$ is equal to that of the process $\mathbf{e}^\top \mathbf{M}^{1/2} B_p(\cdot)$ where $B_p(\cdot) = \{B_p(u) : u \in \mathcal{T}\}$ is a standard $p$-dimensional Brownian bridge, that is a vector consisting of $p$ independent scalar Brownian bridges. Since the sample path of Brownian bridge is continuous a.s., it follows that the process $\mathbf{e}^\top \mathbf{M}^{1/2} B_p(\cdot)$ is a copy of the process $B^*(\cdot)$, conditional on $\{X_i\}_{i=1}^n$.

Now we define the process $\mathbf{e}^\top \mathbf{\Sigma}^{-1/2} B_p(\cdot) = B(\cdot)$ on $\mathcal{T} \subset (0,1)$ as a zero-mean Gaussian process with a.s. continuous sample paths and the covariance function

$$(\tau \wedge \tau' - \tau \tau') \mathbf{e}^\top \mathbf{\Sigma}^{-1} \mathbf{e}$$

Let $\widehat{c} = \mathbf{e}^\top \widehat{\mathbf{D}} \widehat{\mathbf{\Sigma}} \widehat{\mathbf{D}} \mathbf{e}$ and $c = \mathbf{e}^\top \mathbf{\Sigma}^{-1} \mathbf{e}$. By the scaling property of Brownian motion we can see that $1/\sqrt{\widehat{c}} B^*(\cdot)$ is a Brownian motion with unitary variance, conditional on $\mathbf{X}_i$. Additionally $1/\sqrt{c} B(\cdot)$ is Brownian motion with unitary variance.

Combined with $|\widehat{c} - c| = o_P(1)$ and $\widehat{c} > 0$ (Lemma 12 and Equation (8.34) in the proof of Lemma 9), and Theorem 3.1 the above uniform approximation of the quantile process with Brownian bridge provides the asserted claim. $\square$

**Lemma 12.** *Let $\gamma_n \geq c_1 M \sqrt{\log(p \vee n)/n}$ for a constant $c_1 > 0$ independent of $n$ and $p$. For $a_i = \mathbf{x}^\top \widehat{\mathbf{D}} \mathbf{X}_i \in \mathbb{R}$, there exists a positive constant $C$ independent of $n$ and $p$ such that*

$$\mathbb{P}\left( \sup_{\alpha \in (0,1)} n^{-1} \sum_{i \in [n]} (\alpha a_i)^2 \leq C \right) \to 1$$

*Proof of Lemma 12.* First, we observe $\sup_{\alpha \in (0,1)} n^{-1} \sum_{i \in [n]} (\alpha a_i)^2 = \mathbf{e}^\top \widehat{\mathbf{D}} \widehat{\mathbf{\Sigma}} \widehat{\mathbf{D}} \mathbf{e}$.

Second, for $1 \leq i, j \leq p$ we observe that

$$\widehat{d}_i \widehat{\mathbf{\Sigma}} \widehat{d}_j^\top = \left( \widehat{d}_i \widehat{\mathbf{\Sigma}} - e_i \right) \widehat{d}_j^\top + e_i \widehat{d}_j^\top.$$

Additionally as $\|\widehat{d}_j\widehat{\boldsymbol{\Sigma}} - e_j\|_\infty \leq \gamma_n$, by Hölder's inequality and Lemma 2, we have

$$\widehat{d}_i\widehat{\boldsymbol{\Sigma}}\widehat{d}_j^\top = \left(\widehat{d}_i\widehat{\boldsymbol{\Sigma}} - e_i\right)\widehat{d}_j^\top + e_i\widehat{d}_j^\top \leq \left\|\widehat{d}_i\widehat{\boldsymbol{\Sigma}} - e_i\right\|_\infty \|\widehat{d}_j\|_1 + e_i\widehat{d}_j^\top \tag{8.28}$$

$$\leq \gamma_n\left(\|\widehat{d}_j - \boldsymbol{\Sigma}_j^{-1}\|_1 + \|\boldsymbol{\Sigma}_j^{-1}\|_1\right) + e_i\widehat{d}_j^\top \leq \gamma_n\|\boldsymbol{\Sigma}_j^{-1}\|_1 + e_i\widehat{d}_j^\top \tag{8.29}$$

where in the last line we have used that $\|\widehat{d}_j - \boldsymbol{\Sigma}_j^{-1}\|_1 = o_P(1)$ (see for example Theorem 6 of Cai et al. (2011) ). Lastly, we observe that $\sum_{m=1}^p |\Sigma_{im}^{-1}| = \|\boldsymbol{\Sigma}^{-1}\|_1 \leq M$ by Condition (**C**). Additionally, as a consequence of Theorem 6 of Cai et al. (2011) (see Eq (13)) provides a bound on the max matrix norm

$$\|\widehat{\mathbf{D}} - \boldsymbol{\Sigma}^{-1}\|_{\max} = \mathcal{O}_P(M\gamma_n)$$

as long as $\gamma_n \geq c_1 M\|\widehat{\boldsymbol{\Sigma}} - \boldsymbol{\Sigma}\|_{\max}$ for a constant $c_1 > 0$ independent of $n$ and $p$. As $\mathbf{X}_i$'s are sub-Gaussian random vectors, we easily obtain $\|\widehat{\boldsymbol{\Sigma}} - \boldsymbol{\Sigma}\|_{\max} = \mathcal{O}_P(\sigma_X\sqrt{\log p/n})$. Putting all together, we obtain

$$\|\widehat{\mathbf{D}}\widehat{\boldsymbol{\Sigma}}\widehat{\mathbf{D}} - \boldsymbol{\Sigma}^{-1}\|_{\max} = \mathcal{O}_P(M\gamma_n).$$

Thus,

$$\mathbf{e}^\top\widehat{\mathbf{D}}\widehat{\boldsymbol{\Sigma}}\widehat{\mathbf{D}}\mathbf{e} \leq \|\mathbf{e}\|_1^2\|\widehat{\mathbf{D}}\widehat{\boldsymbol{\Sigma}}\widehat{\mathbf{D}}\|_{\max} \leq \|\mathbf{e}\|_1^2(1 + M\gamma_n).$$

The proof is concluded by observing that $C \geq M\|\mathbf{e}\|_1^2$ by assumptions of the Lemma. □

*Proof of Lemma 9.* The proof strategy is similar to that of Lemma 8, in that we divide an argument into two stages – the first conditional on the covariates $\mathbf{X}_i$ and the second moving forward to unconditional distributions.

For the first stage we utilize Lemma 14 of Belloni et al. (2011) where we define an event $\Omega$

$$\Omega = \left\{\sup_{\alpha \in S^{d-1}} n^{-1}\sum_{i \in [n]}(\alpha^\top Z_i)^2 \leq C, \max_{i \in [n]}\|Z_i\|_2 \leq u_p, 1/u_p \leq 1, d^7 u_p^6 = o(n^{1-\delta})\right\} \tag{8.30}$$

with $\delta > 0$ and $S^d$ denoting a unit sphere in $d$ dimensions. With little abuse in notation, constant $C$ in the above display may be different from the one used in other proofs.

In the notation of Lemma 14 therein, we define $v_i = 1$ and $Z_i = \mathbf{w}\widehat{\mathbf{D}}\mathbf{X}_i \in \mathbb{R}^d$ where with little abuse in notation $\mathbf{w} = (\mathbf{V}\widehat{\mathbf{D}}\widehat{\boldsymbol{\Sigma}}\widehat{\mathbf{D}}\mathbf{V}^\top)^{-1/2}\mathbf{V} \in \mathbb{R}^d$. We proceed to discover the size of the constant $C$ and the sequence $u_p$, defined in (8.30).

First we observe that by construction of the vector $Z_i$, $\sup_{\alpha \in S^{d-1}} n^{-1}\sum_{i=1}^n (\alpha^\top Z_i)^2 \leq \lambda_{\max}(\mathbb{I}_d)$ almost surely.

Regarding the sequence $u_p$, we observe

$$\max_{i \in [n]}\|Z_i\|_2^2 = \max_{i \in [n]} \mathbf{X}_i^\top\widehat{\mathbf{D}}\mathbf{V}^\top\mathbf{A}^{-1}\mathbf{V}\widehat{\mathbf{D}}\mathbf{X}_i \in \mathbb{R}_+, \tag{8.31}$$

for a matrix $\mathbf{A} = \mathbf{V}\widehat{\mathbf{D}}\widehat{\boldsymbol{\Sigma}}\widehat{\mathbf{D}}\mathbf{V}^\top \in \mathbb{R}^{d \times d}$. Let $J = \max_j \|\widehat{d}_j\|_0$. Then, by Cauchy-Schwarz inequality we obtain

$$\|Z_i\|_2^2 \leq \|\mathbf{X}_{i,J}\|_2^2\|\widehat{\mathbf{D}}\|_2^2\|\mathbf{V}^\top\mathbf{V}\|_2\|\mathbf{A}^{-1}\|_2 \tag{8.32}$$

$$\leq d\sigma_X\Lambda_{\max}(\boldsymbol{\Sigma})\left(\|\widehat{\mathbf{D}} - \boldsymbol{\Sigma}^{-1}\|_2^2 + \|\boldsymbol{\Sigma}^{-1}\|_2^2\right)\|\mathbf{A}^{-1}\|_2 \tag{8.33}$$

where the last line follows from $\|\mathbf{V}^\top\mathbf{V}\|_2 = \mathcal{O}(d)$ and $\|\mathbf{X}_{i,J}\|_2^2 \leq \Lambda_{\max}(\mathbf{\Sigma})$. We are now left to bound all the elements of the right hand side of (8.33). For that end, firstly we observe that by Condition (**X**)

$$\|\widehat{\mathbf{D}} - \mathbf{\Sigma}^{-1}\|_2^2 = \mathcal{O}_P\left(\bar{s} M \gamma_n\right),$$

by the applying the same methods as in Theorem 6 of Cai et al. (2011) and Eq (14) therein (details are omitted due to space considerations).

Now, we proceed to bound $\|\mathbf{A}^{-1}\|_2$. By Weyl's inequality, we have $|\lambda_{\min}(\mathbf{A}) - \lambda_{\min}(\mathbb{E}[\mathbf{A}])| \leq \|\mathbf{A} - \mathbb{E}[\mathbf{A}]\|_2$. Hence, it suffices to bound the right hand side. As $\mathbb{E}[\mathbf{A}] = \mathbf{V}\mathbf{\Sigma}^{-1}\mathbf{V}^\top$ we observe

$$\lambda_{\min}(\mathbf{A}) \geq \lambda_{\min}(\mathbf{V}\mathbf{\Sigma}^{-1}\mathbf{V}^\top) - \|\mathbf{V}\widehat{\mathbf{D}}\widehat{\mathbf{\Sigma}}\widehat{\mathbf{D}}\mathbf{V}^\top - \mathbf{V}\mathbf{\Sigma}^{-1}\mathbf{V}^\top\|_2.$$

Lastly, we observe $\lambda_{\min}(\mathbf{V}\mathbf{\Sigma}^{-1}\mathbf{V}^\top) \geq \Lambda_{\min}(\mathbf{\Sigma}^{-1}) = 1/\Lambda_{\max}(\mathbf{\Sigma})$ and

$$\|\mathbf{V}\widehat{\mathbf{D}}\widehat{\mathbf{\Sigma}}\widehat{\mathbf{D}}\mathbf{V}^\top - \mathbf{V}\mathbf{\Sigma}^{-1}\mathbf{V}^\top\|_2 \leq d\|\widehat{\mathbf{D}}\widehat{\mathbf{\Sigma}}\widehat{\mathbf{D}} - \mathbf{\Sigma}^{-1}\|_2 = \mathcal{O}_P(d\bar{s}^2 M^2 \gamma_n) \tag{8.34}$$

where we utilized Lemma 13. Finally, with $\gamma_n \geq CM\sqrt{\log p/n}$ for a constant $C > 0$, we obtain

$$\max_{i \in [n]} \|Z_i\|_2^2 \leq \sigma_X \Lambda_{\max}^3(\mathbf{\Sigma}) d^2 \bar{s}^2 M^2 \log^{3/2}(p \vee n)/\sqrt{n}. \tag{8.35}$$

Per these results, $\mathbb{P}(\Omega) \to 1$ as $n \to \infty$. Hence, the above implies that there exists a process $B^*(\cdot) = B_n^*(\cdot)$ on $\mathcal{T} \subset (0,1)$ that, conditionally on $\mathbf{X}_i$ is zero-mean Gaussian process with a.s. continuous sample paths and the covariance function

$$(\tau \wedge \tau' - \tau\tau')\mathbb{I}_d$$

such that

$$\sup_{\tau \in \mathcal{T}} \left\| \frac{1}{\sqrt{n}} \sum_{i \in [n]} Z_i \psi(u_i - F^{-1}(\tau)) - B_d^*(u) \right\|_2 \leq o(n^{-\epsilon'})$$

for some $\epsilon' > 0$. The process $B_d^*(\cdot)$ is a $d$-dimensional Brownian bridge with respect to the filtration $\mathcal{F}_\tau^u = \sigma(\psi(u - F^{-1}(\tau)); \tau \in \mathcal{T})$.

$\square$

**Lemma 13.** *For $\widehat{\mathbf{D}}$ and a tuning parameter $\gamma_n$ as defined in (2.4) and chosen as $\gamma_n \geq c_1 M\sqrt{\log(p \vee n)/n}$ for a constant $c_1 > 0$ independent of $n$ and $p$,*

$$\|\widehat{\mathbf{D}}\widehat{\mathbf{\Sigma}}\widehat{\mathbf{D}} - \mathbf{\Sigma}^{-1}\|_1 = \mathcal{O}_P\left(\bar{s} M^2 \sqrt{\log p/n} \bigvee \bar{s}^2 M^2 \Lambda_{\max}(\mathbf{\Sigma})\gamma_n\right).$$

*Proof of Lemma 13.* With some abuse in notation let us define a matrix $\mathbf{\Xi} = \widehat{\mathbf{D}}\widehat{\mathbf{\Sigma}}\widehat{\mathbf{D}}$, where it is understood that $\mathbf{\Xi}$ defined above is only used in this proof.

Let $u_n = \|\widehat{\mathbf{D}} - \mathbf{\Sigma}^{-1}\|_1$. Following the steps of the proof of Theorem 6 of Cai et al. (2011) and Eq (14) therein, it is easy to obtain that $u_n = \gamma_n M \bar{s}$ and that for a certain choice of $\gamma_n$ it is a sequence of numbers converging to zero. Here $\bar{s} = \max_j \|\mathbf{\Sigma}_j^{-1}\|_0$ denote the maximal row- sparsity of the precision matrix throughout this proof. Observe that by Lemma 2, we have

$$\|\widehat{\mathbf{d}}_j\|_1 \leq \|\mathbf{\Sigma}_j^{-1}\|_1, \tag{8.36}$$

where $\mathbf{d}_j$ is the $j$-th row of $\mathbf{D}$. We proceed to show that there exists a sequence of positive numbers $t_n$ converging to zero, such that $\|\boldsymbol{\xi}_j - \boldsymbol{\Sigma}_j^{-1}\|_1 = \mathcal{O}_P(t_n)$, where $\boldsymbol{\xi}_j$ is the $j$-th row of $\boldsymbol{\Xi}$. For that end we denote with $\widehat{\boldsymbol{\sigma}}_l, \boldsymbol{\sigma}_l$, the $l$-th column of the covariance matrix $\widehat{\boldsymbol{\Sigma}}, \boldsymbol{\Sigma}$ and observe

$$\|\boldsymbol{\xi}_j - \boldsymbol{\Sigma}_j^{-1}\|_1 = \sum_{l,k\in[p+1]} \left|\widehat{\mathbf{d}}_j\widehat{\boldsymbol{\sigma}}_l\widehat{\mathbf{d}}_{lk} - \boldsymbol{\Sigma}_j^{-1}\boldsymbol{\sigma}_l\boldsymbol{\Sigma}_{lk}^{-1}\right|$$

$$\stackrel{(i)}{\leq} \sum_{l,k\in[p+1]} \left|\widehat{\mathbf{d}}_j\widehat{\boldsymbol{\sigma}}_l\left(\widehat{\mathbf{d}}_{lk} - \boldsymbol{\Sigma}_{lk}^{-1}\right)\right| + \sum_{l,k\in[p+1]} \left|\widehat{\mathbf{d}}_j\widehat{\boldsymbol{\sigma}}_l - \boldsymbol{\Sigma}_j^{-1}\boldsymbol{\sigma}_l\right| \left|\boldsymbol{\Sigma}_{lk}^{-1}\right|$$

$$:= T_1 + T_2, \tag{8.37}$$

where (i) follows from the triangular inequality. We proceed to bound each of the terms $T_1$ and $T_2$ independently. By the basic inequality of the Dantzig-type estimators in that $\|\widehat{\mathbf{d}}_j - \boldsymbol{\Sigma}_j^{-1}\|_1 \leq 2\|\widehat{\mathbf{d}}_{j,\bar{S}} - \boldsymbol{\Sigma}_{j,\bar{S}}^{-1}\|_1$ where $\bar{S}$ denotes the set of true non-zero indices

$$T_1 \leq 2\left(\max_{l\in[p+1]} \left|\widehat{\mathbf{d}}_j\widehat{\boldsymbol{\sigma}}_l\right|\right) \sum_{l\in[p+1]} \sum_{k\in\bar{S}} \left|\widehat{\mathbf{d}}_{lk} - \boldsymbol{\Sigma}_{lk}^{-1}\right|$$

$$\stackrel{(ii)}{\leq} 2\bar{s}\|\widehat{\mathbf{d}}_j\|_1 \max_{l\in[p+1]} \|\widehat{\boldsymbol{\sigma}}_l\|_\infty \max_{k\in[p+1]} \sum_{l\in[p+1]} \left|\widehat{\mathbf{d}}_{lk} - \boldsymbol{\Sigma}_{lk}^{-1}\right|$$

$$\stackrel{(iii)}{\leq} 4\bar{s}M\|\boldsymbol{\Sigma}\|_\infty\|\widehat{\mathbf{D}} - \boldsymbol{\Sigma}^{-1}\|_1 \leq 4\bar{s}^2 M^2 \Lambda_{\max}(\boldsymbol{\Sigma})\gamma_n, \tag{8.38}$$

(ii) follows from Hölder's inequality; (iii) follows from the $\|\widehat{\boldsymbol{\Sigma}} - \boldsymbol{\Sigma}\|_{\max} = \mathcal{O}_P(\sqrt{\log p/n}) = o_P(1)$, the equation (8.36) and the symmetric structure of the estimator $\widehat{\mathbf{D}}$; and the last inequality follows from the definition of the constant $M$ and the sequence $u_n$.

Regarding the term $T_2$ we see by the definition of the precision matrix $\boldsymbol{\Sigma}^{-1}$ and its sparsity we see

$$T_2 = \sum_{k\in[p+1]} \sum_{l\in\bar{S}} \left|\widehat{\mathbf{d}}_j\widehat{\boldsymbol{\sigma}}_l - \boldsymbol{\Sigma}_j^{-1}\boldsymbol{\sigma}_l\right| \left|\boldsymbol{\Sigma}_{kl}^{-1}\right|$$

$$\stackrel{(iv)}{\leq} \max_{l\in[p+1]} \left(\sum_{k\in[p+1]} \left|\boldsymbol{\Sigma}_{kl}^{-1}\right|\right) \sum_{l\in\bar{S}} \left(\left|\widehat{\mathbf{d}}_j\widehat{\boldsymbol{\sigma}}_l - \widehat{\mathbf{d}}_j\boldsymbol{\sigma}_l\right| + \left|\widehat{\mathbf{d}}_j\boldsymbol{\sigma}_l - \boldsymbol{\Sigma}_j^{-1}\boldsymbol{\sigma}_l\right|\right)$$

$$\stackrel{(v)}{\leq} M \sum_{l\in\bar{S}} \left(\|\widehat{\mathbf{d}}_j\|_1\|\widehat{\boldsymbol{\sigma}}_l - \boldsymbol{\sigma}_l\|_\infty + \|\widehat{\mathbf{d}}_j - \boldsymbol{\Sigma}_j^{-1}\|_1\|\boldsymbol{\sigma}_l\|_\infty\right)$$

$$\stackrel{(vi)}{\leq} M^2\bar{s}\sqrt{\log p/n} + M\bar{s}\|\widehat{\mathbf{D}} - \boldsymbol{\Sigma}^{-1}\|_1\|\boldsymbol{\Sigma}\|_{\max}$$

$$\stackrel{(vii)}{\leq} M^2\bar{s}\sqrt{\log p/n} + M^2\bar{s}^2\Lambda_{\max}(\boldsymbol{\Sigma})\gamma_n, \tag{8.39}$$

(iv) follows from triangular inequality; (v) follows from the Hoölder's inequality and the definition of $M$; (vi) follows from (8.36), the definition of M and from $\|\widehat{\boldsymbol{\Sigma}} - \boldsymbol{\Sigma}\|_{\max} = \mathcal{O}_P(\sqrt{\log p/n})$; (vii) the definition of $u_n$ and the symmetric structure of the estimator $\widehat{\mathbf{D}}$.

□

*Proof of Lemma 10.* Apply Lemma 9 together with Theorem 4.2 and repeat the same steps as in the proof of Lemma 7. □